\documentclass{article}
\usepackage[final]{nips_2019}
\usepackage{times}

\usepackage
[acronym,smallcaps,nowarn,section,nogroupskip,nonumberlist]{glossaries}
\glsdisablehyper{}

\newacronym{VI}{vi}{variational inference}
\newacronym{KL}{kl}{Kullback-Leibler}
\newacronym{ELBO}{elbo}{\emph{evidence lower bound}}
\newacronym{MCMC}{mcmc}{Markov chain Monte Carlo}

\newacronym{BNP}{bnp}{Bayesian Nonparametric}
\newacronym{PPL}{ppl}{Probabilistic Programming Language}
\newacronym{PPLs}{ppls}{Probabilistic Programming Languages}
\newacronym{PL}{pl}{Programming Language}

\newacronym{SGMCMC}{sgmcmc}{Stochastic Gradient Markov chain Monte Carlo}
\newacronym{MH}{mh}{Metropolis-Hastings}
\newacronym{IS}{is}{Importance Sampling}
\newacronym{SIS}{sis}{Sequential Importance Sampling}
\newacronym{SMC}{smc}{Sequential Monte Carlo}
\newacronym{CSMC}{csmc}{Conditional Sequential Monte Carlo}
\newacronym{PMCMC}{PMCMC}{Particle Markov chain Monte Carlo}
\newacronym{PG}{pg}{Particle Gibbs}
\newacronym{PMMH}{pmmh}{Particle Marginal Metropolis-Hastings}
\newacronym{IPMCMC}{IPMCMC}{interacting particle Markov chain Monte Carlo}
\newacronym{PGAS}{pgas}{Particle Gibbs with Ancestor Sampling}
\newacronym{AS}{as}{Ancestor Sampling}

\newacronym{HMC}{hmc}{Hamiltonian Monte Carlo}
\newacronym{HMCDA}{hmcda}{Hamiltonian Monte Carlo with Dual Averaging}
\newacronym{NUTS}{nuts}{No-U-Turn Sampler}
\newacronym{SGLD}{sgld}{Stochastic Gradient Langevin Dynamics}
\newacronym{SGHMC}{sghmc}{Stochastic Gradient Hamiltonian Monte Carlo}

\newacronym{CPS}{cps}{Continuation-Passing Style}
\newacronym{AD}{ad}{Automatic Differentiation}
\newacronym{TCO}{tco}{Tail Call Optimization}

\newacronym{SSM}{ssm}{State-Space Model}
\newacronym{ESS}{ess}{Effective Sample Size}

\newacronym{CRM}{crm}{Completely Random Measure}
\newacronym{NRM}{nrm}{Normalised Random Measure}
\newacronym{PK}{pk}{Poisson Kingman}
\newacronym{PKP}{pkp}{Poisson Kingman Process}
\newacronym{RPM}{rpm}{Random Probability Measure}
\newacronym{DP}{dp}{Dirichlet Process}
\newacronym{CRP}{crp}{Chinese Restaurant Process}
\newacronym{PY}{py}{Pitman-Yor process}
\newacronym{NGGP}{nggp}{Normalised Generalised Gamma Process}
\newacronym{GGP}{ggp}{Generalised Gamma Process}
\newacronym{IG}{ig}{inverse-Gaussian}
\newacronym{NS}{ns}{Normalised $\sigma$-Stable process}
\newacronym{pgc}{PG}{Poisson-Gamma class}
\newacronym{GT}{gt}{Gamma-tilted process}

\newacronym{IBP}{ibp}{Indian Buffet Process}
\newacronym{BP}{bp}{Beta Process}
\newacronym{HMM}{hmm}{Hidden Markov Model}
\newacronym{HDP}{hdp}{Hierarchical Dirichlet Process}

\newacronym{GP}{gp}{Gaussian Process}
\newacronym{SBS}{SBS}{Size-Biased Sampling}

\newcommand{\g}{\,|\,}

\DeclareMathOperator*{\argmin}{arg\,min}

\usepackage[utf8]{inputenc} 
\usepackage[T1]{fontenc}    
\usepackage{hyperref}       
\usepackage{url}            
\usepackage{booktabs}       
\usepackage{amsfonts}       
\usepackage{nicefrac}       
\usepackage{microtype}      
\usepackage{todonotes}      
\usepackage{bm}             
\usepackage{appendix}       
\usepackage{amssymb}        
\usepackage{amsmath}        
\usepackage{amsthm}
\usepackage{mathtools}
\usepackage{esvect}         
\usepackage{bbold}          %
\usepackage{tkz-euclide}    
\usepackage[noend]{algpseudocode}
\usepackage{algorithm}  
\usepackage{caption, subcaption}
\usetkzobj{all}
\usepackage{wrapfig}
\usepackage{graphicx}
\usepackage{multirow}
\usepackage{sectsty}
\usepackage{arydshln}
\usepackage{booktabs}
\usepackage{titlesec}
\usepackage{enumitem}

\def\x{\bm{x}}
\def\z{\bm{z}}
\def\y{\bm{y}}

\def\erf{\text{erf}}

\def\u{\bm{u}}

\def\v{\bm{v}}
\def\a{\bm{a}}
\def\p{\bm{p}}
\def\0{\bm{0}}
\def\M{\mathcal{M}}

\def\X{\mathcal{X}}
\def\Z{\mathcal{Z}}
\def\T{\mathcal{T}}
\def\R{\mathbb{R}}
\def\B{\mathbb{B}}

\def\g{\mathfrak{g}}
\def\1{\mathbb{1}}

\DeclareRobustCommand{\pvae}[1]{$\mathcal{P}^{#1}$-VAE}
\DeclareRobustCommand{\nvae}{$\mathcal{N}$-VAE}
\DeclarePairedDelimiter{\ceil}{\lceil}{\rceil}

\titlespacing{\section}{0pt}{.5\baselineskip}{0.3em}
\titlespacing{\subsection}{0pt}{.5\baselineskip}{0.3em}
\titlespacing{\paragraph}{0pt}{.3\baselineskip}{1em}

\glsdisablehyper{}
\newacronym{GCN}{gcn}{graph convolutional network}
\newacronym{VGAE}{vgae}{variational graph auto-encoder}
\newacronym{VAE}{vae}{variational auto-encoder}
\newacronym{AE}{ae}{auto-encoder}
\newacronym{SGD}{sgd}{stochastic gradient descent}
\newacronym[sort=beta]{BVAE}{\(\beta\)-vae}{}
\newacronym{TC}{tc}{total correlation}
\newacronym{MMD}{mmd}{maximum mean discrepancy}
\newacronym{GAN}{gan}{generative adversarial network}
\newacronym{AAE}{aae}{adversarial auto-encoder}
\newacronym{CCM}{ccm}{constant curvature manifold}
\newacronym{IWAE}{iwae}{importance weighted auto-encoder}
\newacronym{MC}{mc}{Monte Carlo}
\newacronym{ARS}{ars}{adaptive rejection sampling}
\newacronym{PCA}{pca}{principal component analysis}
\newacronym{GPLVM}{gplvm}{Gaussian process latent variable model}
\newacronym{SVD}{svd}{singular-value decomposition}
\newacronym{MLP}{mlp}{multilayer perceptron}

\graphicspath{{images/}}

\title{Continuous Hierarchical Representations with \\ Poincar\'e Variational Auto-Encoders}

\author{
  Emile Mathieu$^{\dagger}$ \\ %
  \texttt{emile.mathieu@stats.ox.ac.uk} \\
  \And
  Charline Le Lan$^{\dagger}$ \\
  \texttt{charline.lelan@stats.ox.ac.uk} \\
  \AND
  Chris J. Maddison$^{\dagger*}$ \\
  \texttt{cmaddis@stats.ox.ac.uk} \\
  \And
  Ryota Tomioka$^{\ddagger}$ \\
  \texttt{ryoto@microsoft.com} \\
  \And
  Yee Whye Teh$^{\dagger*}$ \\
  \texttt{y.w.teh@stats.ox.ac.uk} \\
  \\
  $\dagger$ Department of Statistics, University of Oxford, United Kingdom\\
  $*$ DeepMind, London, United Kingdom\\
  $\ddagger$ Microsoft Research, Cambridge, United Kingdom\\
}

\begin{document}

\maketitle

\begin{abstract}
\vspace{-.3em}
The \gls{VAE} is a popular method for learning a generative model and embeddings of the data. 
Many real datasets are hierarchically structured.
However, traditional \glspl{VAE} map data in a Euclidean latent space which cannot efficiently embed tree-like structures.
Hyperbolic spaces with negative curvature can.
We therefore endow \glspl{VAE} with a Poincar\'e ball model of hyperbolic geometry as a latent space and 
rigorously derive the necessary methods to work with two main Gaussian generalisations on that space.
We empirically show better generalisation to unseen data than the Euclidean counterpart, and can qualitatively and quantitatively better recover hierarchical structures.
\end{abstract}


\section{Introduction}
\begin{wrapfigure}{r}{0.45\textwidth}
\begin{center}
\vspace{-2.em}
  \includegraphics[width=0.44\textwidth]{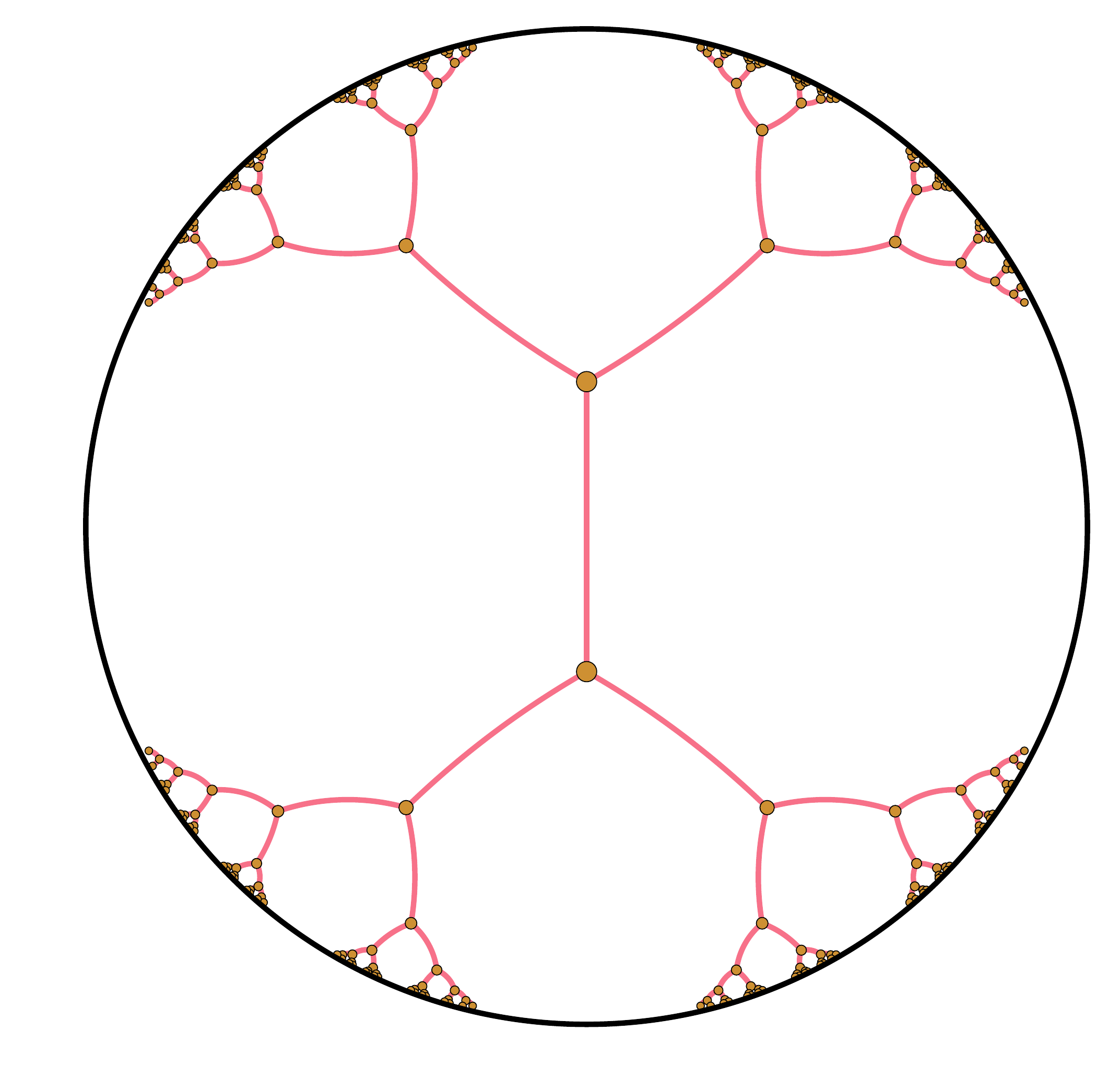}
  \caption{A regular tree isometrically embedded in the Poincar\'e disc. Red curves are same length \emph{geodesics}, i.e. "straight lines".}
  \label{fig:tree_embedding}
\end{center}
\vspace{-2.5em}
\end{wrapfigure}
Learning useful representations from unlabelled raw sensory observations, which are often high-dimensional, is a problem of significant importance in machine learning.
\Acrfullpl{VAE} \citep{KingmaW13,Rezende:2014vm} are a popular approach to this: they are probabilistic generative models composed of an \emph{encoder} stochastically embedding observations in a low dimensional latent space $\Z$, and a \emph{decoder} generating observations $\x \in \X$ from encodings $\z \in \Z$.
After training, the encodings constitute a low-dimensional representation of the original raw observations, which can be used as features for a downstream task \citep[e.g.][]{DBLP:conf/cvpr/HuangL06,coates2011analysis} or be interpretable for their own sake.
\Glspl{VAE} are therefore of interest for representation learning \citep{Bengio2013}, a field which aims to learn \emph{good representations}, e.g.\ interpretable representations, ones yielding better generalisation, or ones useful for downstream tasks. 

It can be argued that in many domains data should be represented hierarchically. 
For example, in cognitive science, it is widely accepted that human beings use a hierarchy to organise object categories \citep[e.g.][]{DBLP:conf/nips/RoyKMT06,collins69,keil1979semantic}.
In biology, the theory of evolution \citep{darwin1859} implies that features of living organisms are related in a hierarchical manner given by the evolutionary tree.
Explicitly incorporating hierarchical structure in probabilistic models has unsurprisingly been a long-running research topic \citep[e.g.][]{Duda:2000:PC:954544,Heller:2005:BHC:1102351.1102389}.

Earlier work in this direction tended to use trees as data structures to represent hierarchies. 
Recently, hyperbolic spaces have been proposed as an alternative continuous approach to learn hierarchical representations from textual and graph-structured data \citep{Nickel:2017wz,Tifrea:2018ty}.
Hyperbolic spaces can be thought of as continuous versions of trees, and vice versa, as illustrated in Figure \ref{fig:tree_embedding}.
Trees can be embedded with arbitrarily low error into the Poincar\'e disc model of hyperbolic geometry \citep{10.1007/978-3-642-25878-7_34}.
The exponential growth of the Poincar\'e surface area with respect to its radius is analogous to the exponential growth of the number of leaves in a tree with respect to its depth.
Further, these spaces are smooth, enabling the use of deep learning approaches which rely on differentiability.

We show that replacing \glspl{VAE} latent space components, which traditionally assume a Euclidean metric over the latent space, by their hyperbolic generalisation helps to represent and discover hierarchies.
Our goals are twofold: 
(a) learn a latent representation that is interpretable in terms of hierarchical relationships among the observations, 
(b) learn a more efficient representation which generalises better to unseen data that is hierarchically structured.
Our main contributions are as follows:
\begin{enumerate}[leftmargin=2em]
	\item We propose efficient and reparametrisable sampling schemes, and calculate the probability density functions, for two canonical Gaussian generalisations defined on the Poincar\'e ball, namely the maximum-entropy and wrapped normal distributions. These are the ingredients required to train our \glspl{VAE}.
	\item We introduce a decoder architecture that explicitly takes into account the hyperbolic geometry, which we empirically show to be crucial.
    \item We empirically demonstrate that endowing a \gls{VAE} with a Poincar\'e ball latent space can be beneficial in terms of model generalisation and can yield more interpretable representations.
\end{enumerate}
Our work fits well with a surge of interest in combining hyperbolic geometry and \glspl{VAE}.
Of these, it relates most strongly to the concurrent works of \citet{Ovinnikov:2018ue,Grattarola:2018ue,Nagano}.
In contrast to these approaches, we introduce a decoder that takes into account the geometry of the hyperbolic latent space.
Along with the \emph{wrapped normal} generalisation used in the latter two articles, we give a thorough treatment of the \emph{maximum entropy normal} generalisation and a rigorous analysis of the difference between the two.
Additionally, we train our model by maximising a lower bound on the marginal likelihood, as opposed to \citet{Ovinnikov:2018ue,Grattarola:2018ue} which consider a Wasserstein and an adversarial auto-encoder setting, respectively.
We discuss these works in more detail in Section \ref{sec:related_work}.


\section{The Poincar\'e Ball model of hyperbolic geometry}
\label{sec:geometry}
\subsection{Review of Riemannian geometry}
Throughout the paper we denote the Euclidean norm and inner product by $\left\|\cdot\right\|$ and $\langle \cdot, \cdot \rangle$ respectively.
A real, smooth \emph{manifold} $\M$ is a set of points $\z$, which is "locally similar" to a linear space.
For every point $\z$ of the manifold $\M$ is attached a real vector space of the same dimensionality as $\M$ called the \emph{tangent space} $\T_{\z}\M$.
Intuitively, it contains all the possible directions in which one can tangentially pass through $\z$.
For each point $\z$ of the manifold, the \emph{metric tensor} $\g(\z)$ defines an inner product on the associated tangent space
: $\g(\z) = \langle\cdot,\cdot \rangle_{\z}: \T_{\z}\M \times \T_{\z}\M \rightarrow \R$.
The \emph{matrix representation of the Riemannian metric} $G(\z)$, is defined such that $\forall \bm{u},\bm{v} \in \T_{\z}\M\times\T_{\z}\M, \ \langle\bm{u},\bm{v}\rangle_{\z} = \g(\z)(\bm{u},\bm{v}) = \bm{u}^T G(\z) \bm{v}$.
A \emph{Riemannian manifold} is then defined as a tuple $(\M, \g)$ \citep{Petersen}.
The metric tensor gives a \emph{local} notion of angle, length of curves, surface area and volume, from which \emph{global} quantities can be derived by integrating local contributions.
A norm is induced by the inner product on $\T_{\z}\M$: $\left\| \cdot \right\|_{\z}=\sqrt{\langle\cdot,\cdot\rangle_{\z}}$. 
An infinitesimal volume element is induced on each tangent space $\T_{\z}\M$, and thus a measure $d\M(\z) = \sqrt{|G(\z)|}d\z$ on the manifold, with $d\z$ being the Lebesgue measure.
The length of a curve $\gamma: t \mapsto \gamma(t) \in \M$ is given by 
$L(\gamma)=\int_{0}^{1}{\|\gamma'(t)\|_{\gamma(t)}^{1/2} dt}$.
The concept of straight lines can then be generalised to \emph{geodesics}, which are constant speed curves giving the shortest path between pairs of points $\z, \y$ of the manifold: $\gamma^* = \argmin L(\gamma)$ with $\gamma(0)=\z$, $\gamma(1)=\y$ and $\left\|\gamma'(t)\right\|_{\gamma(t)}=1$.
A \emph{global} distance is thus induced on $\M$ given by $d_{\M}(\z, \y) = \inf L(\gamma)$.
Endowing $\M$ with that distance consequently defines a metric space $(\M, d_\M)$.
The concept of moving along a "straight" curve with constant velocity is given by the \emph{exponential map}.
In particular, there is a unique unit speed \emph{geodesic} $\gamma$ satisfying $\gamma(0) = \z$ with initial tangent vector $\gamma'(0) = \v$.
The corresponding exponential map is then defined by $\exp_{\z}(\bm{v}) = \gamma(1)$, as illustrated on Figure \ref{fig:geodesics}.
The \emph{logarithm map} is the inverse $\log_{\z} = \exp_{\z}^{-1}: \M \rightarrow \T_{\z}\M$.
For geodesically complete manifolds, such as the Poincar\'e ball, $\exp_{\z}$ is well-defined on the full tangent space $\T_{\z}\M$ for all $z \in \M$.
\subsection{The Poincar\'e ball model of hyperbolic geometry}
\begin{wrapfigure}{r}{0.33\textwidth}
\vspace{-5.em}
\begin{center}
  \includegraphics[width=0.30\textwidth]{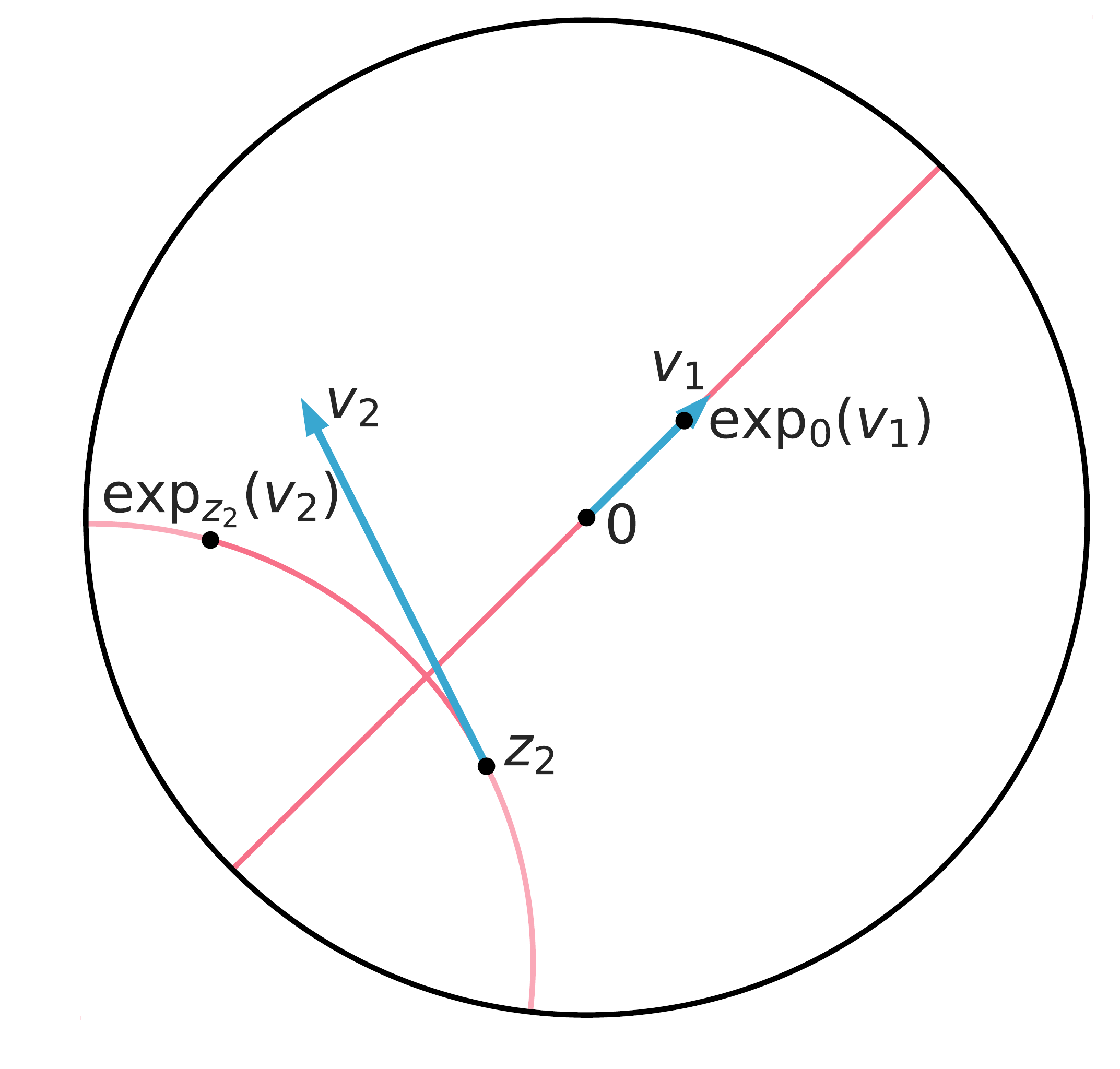}
  \caption{Geodesics and exponential maps in the Poincar\'e disc.}
  \label{fig:geodesics}
\end{center}
\vspace{-2.em}
\end{wrapfigure}
A $d$-dimensional hyperbolic space, denoted $\mathbb{H}^d$, is a complete, simply connected, $d$-dimensional Riemannian manifold with constant negative curvature $c$.
In contrast with the Euclidean space $\R^d$, $\mathbb{H}^d$ can be constructed using various isomorphic models (none of which is prevalent), including the hyperboloid model, the Beltrami-Klein model, the Poincar\'e half-plane model and the Poincar\'e ball $\mathcal{B}_c^d$ \citep{beltrami1868teoria}.
The Poincar\'e ball model is formally defined as the Riemannian manifold $\B_c^d = (\mathcal{B}_c^d, \g_p^c)$, where $\mathcal{B}_c^d$ is the open ball of radius $1/\sqrt{c}$, and $\g_p^c$ its \emph{metric tensor}, which along with its induced distance are given by
\begin{align*}
\g_p^c(\z) = ({\lambda_{\z}^c)}^2 ~\g_e(\z), \ \
d^c_p(\z,\y) = \frac{1}{\sqrt{c}} \cosh^{-1} \left( 1 + 2c \frac{||\z-\y||^2}{(1-c\left\|\z\right\|^2)(1-c\left\|\y\right\|^2)} \right),
\end{align*}
where $\lambda_{\z}^c = \frac{2}{1 - c\left\|\z\right\|^2} $ and $\g_e$ denotes the Euclidean metric tensor, i.e. the usual dot product.

The \emph{Möbius addition} \citep{doi:10.2200/S00175ED1V01Y200901MAS004} of $\z$ and $\y$ in $\B^d_c$ is defined as
\begin{align*}
\z \oplus_c \y = \frac{(1 + 2c \left\langle\z,\y\right\rangle + c \|\y\|^2)\z + (1 - c\|\z\|^2)\y}{1 + 2c \left\langle\z,\y\right\rangle + c^2 \|\z\|^2 \|\y\|^2}.
\end{align*}
One recovers the Euclidean addition of two vectors in $\R^d$ as $c \rightarrow 0$.
Building on that framework, \cite{Ganea:2018wy} derived closed-form formulations for the \emph{exponential map} (illustrated in Figure \ref{fig:geodesics})
\begin{align*}
\exp^c_{\z}(\v) = \z \oplus_c \left(\tanh\left( \sqrt{c} \frac{\lambda^c_{\z}\|\v\|}{2} \right)\frac{\v}{\sqrt{c}\|\v\|} \right)
\end{align*}
and its inverse, the \emph{logarithm map}
\begin{align*}
\log^c_{\z}(\y) = \frac{2}{\sqrt{c}\lambda^c_{\z}} \tanh^{-1} \left(\sqrt{c} \|-\z \oplus_c \y \| \right) \frac{-\z \oplus_c \y}{\|-\z \oplus_c \y \|}.
\end{align*}

%
\vspace{-1.em}
\section{The Poincar\'e VAE}
\label{sec:pvae}
We consider the problem of mapping an empirical distribution of observations to a lower dimensional Poincar\'e ball $\B_c^d$, as well as learning a map from this latent space $\Z=\B_c^d$ to the observation space $\X$.
Building on the \gls{VAE} framework, this \emph{Poincar\'e}-\gls{VAE} model, or \pvae{c} for short, differs by the choice of prior and posterior distributions being defined on $\B_c^d$, and by the encoder $g_{\bm{\phi}}$ and decoder $f_{\bm{\theta}}$ maps which take into account the latent space geometry.
Their parameters $\{\bm{\theta}, \bm{\phi}\}$ are learned by maximising the \gls{ELBO}.
Our model can be seen as a generalisation of a classical Euclidean \gls{VAE} \citep{KingmaW13,Rezende:2014vm} that we denote by \nvae{}, i.e. \pvae{c} $\xrightarrow [ c \rightarrow 0 ]{  } \mathcal{N}$-VAE.

\subsection{Prior and variational posterior distributions}
In order to parametrise distributions on the Poincar\'e ball, we consider two canonical generalisations of normal distributions on that space.
A more detailed review of Gaussian generalisations on manifolds can be found in Appendix \ref{sec:measures_riem}.
\paragraph{Riemannian normal}
One generalisation is the distribution maximising entropy given an expectation and variance \citep{DBLP:journals/entropy/SaidBB14,Pennec2006,Hauberg}, often called the \emph{Riemannian normal} distribution, which has a density w.r.t. the metric induced measure $d\M$ given by
\begin{align} \label{eq:core_maxent_normal_hyp}
\mathcal{N}^{\text{R}}_{\B_c^d}(\z|\bm{\mu}, \sigma^2) 
=\frac{d\nu^{\text{R}}(\z|\bm{\mu}, \sigma^2)}{d\M(\z)} 
=  \frac{1}{Z^{\text{R}}} \exp\left(- \frac{d_p^c(\bm{\mu}, \z)^2} {2 \sigma^2}\right),
\end{align}
where $\sigma > 0$ is a dispersion parameter, $\bm{\mu} \in \B_c^d$ is the Fr\'echet mean 
, and $Z^{\text{R}}$ is the normalising constant derived in Appendix \ref{sec:normconst}. 
\begin{wrapfigure}{r}{0.4\textwidth}
\vspace{-1em}
\centering
\begin{subfigure}{0.17\textwidth}
\includegraphics[width=\linewidth]{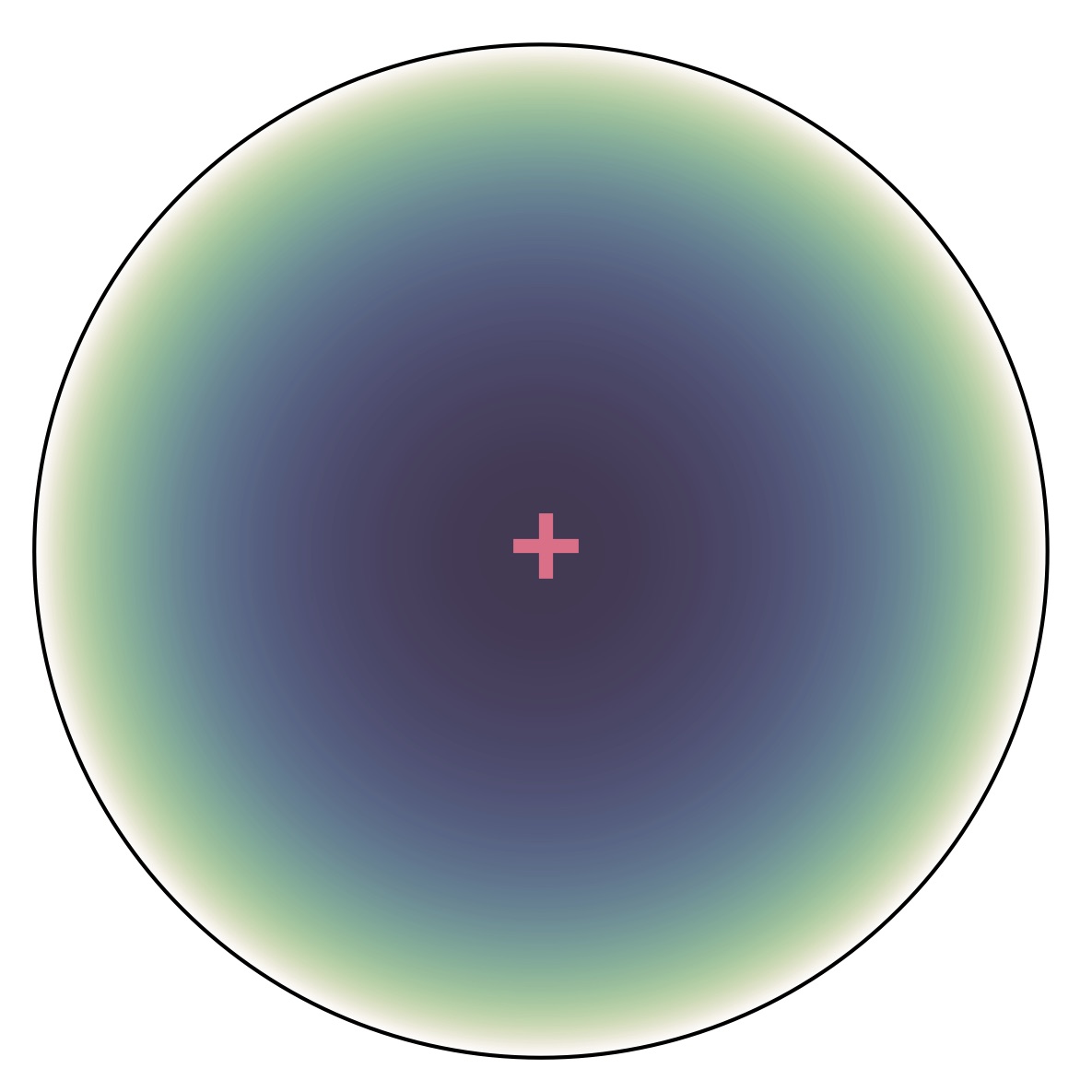}
\put(-60,70){Riemannian}
\put(-80,17){\rotatebox{90}{$\sqrt{c}\|\bm{\mu}\|_2=0$}}
\end{subfigure}
\begin{subfigure}{0.17\textwidth}
\includegraphics[width=\linewidth]{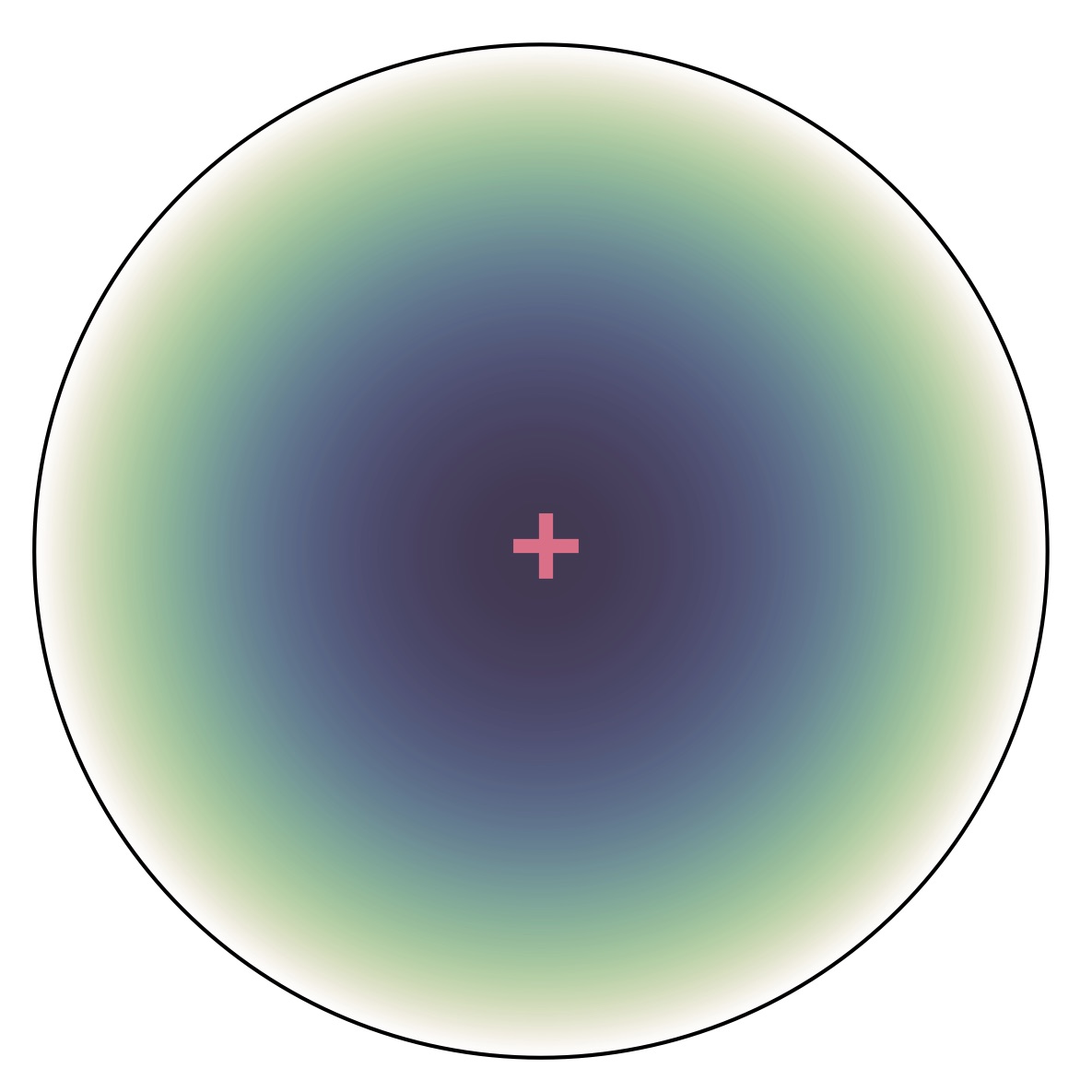}
\put(-50,70){Wrapped}
\end{subfigure}
\begin{subfigure}{0.17\textwidth}
\includegraphics[width=\linewidth]{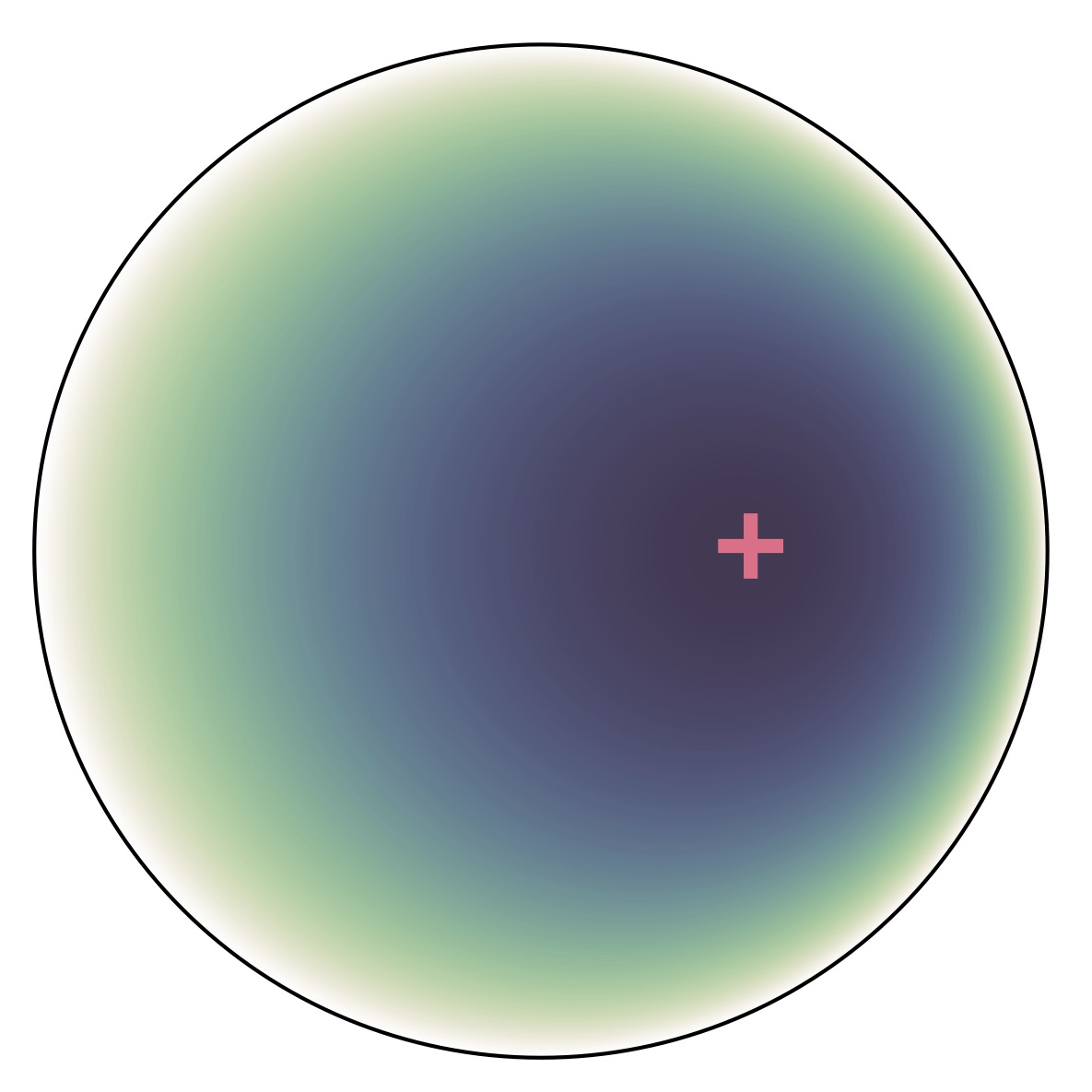}
\put(-80,13){\rotatebox{90}{$\sqrt{c}\|\bm{\mu}\|_2=0.4$}}
\end{subfigure}
\begin{subfigure}{0.17\textwidth}
\includegraphics[width=\linewidth]{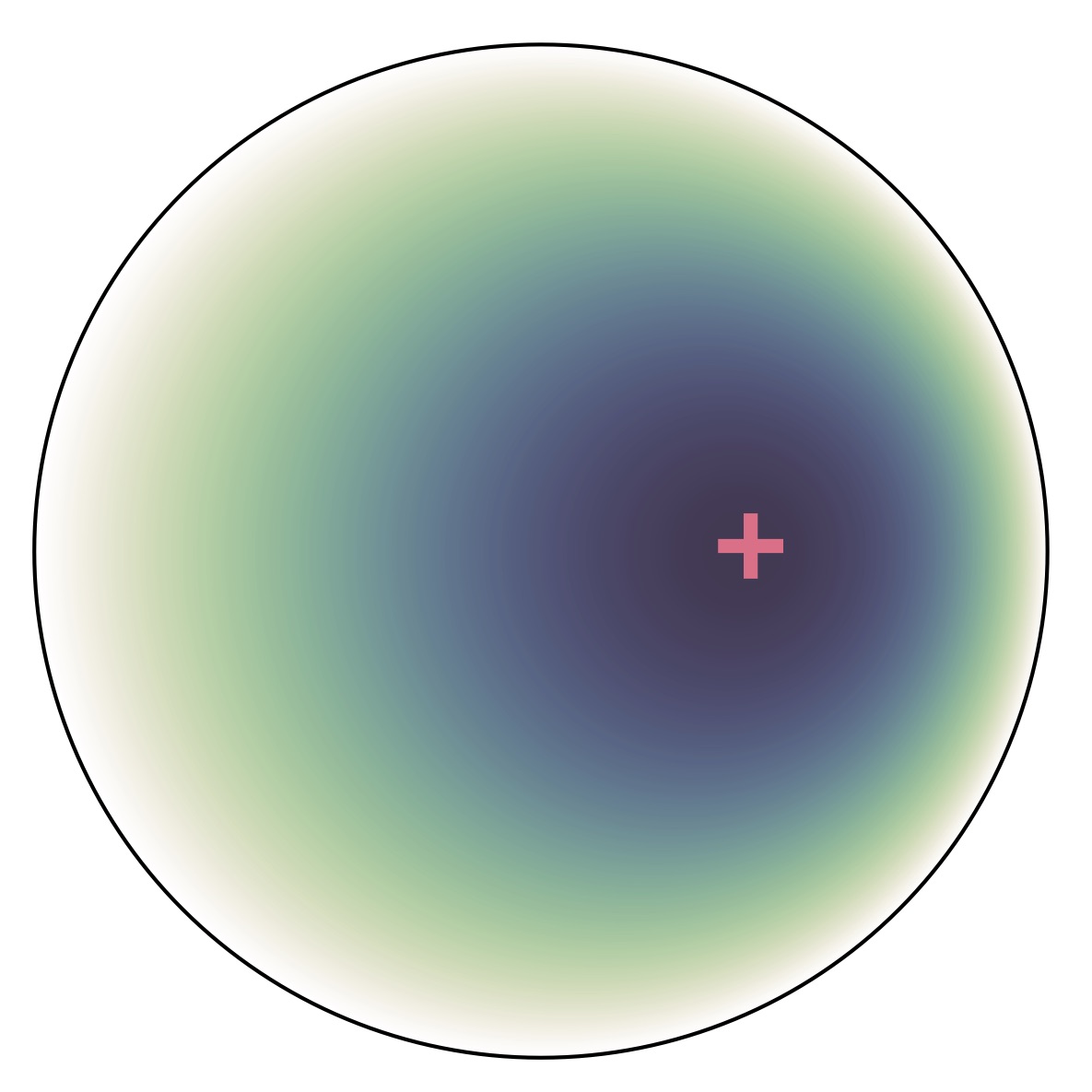}
\end{subfigure}
\begin{subfigure}{0.17\textwidth}
\includegraphics[width=\linewidth]{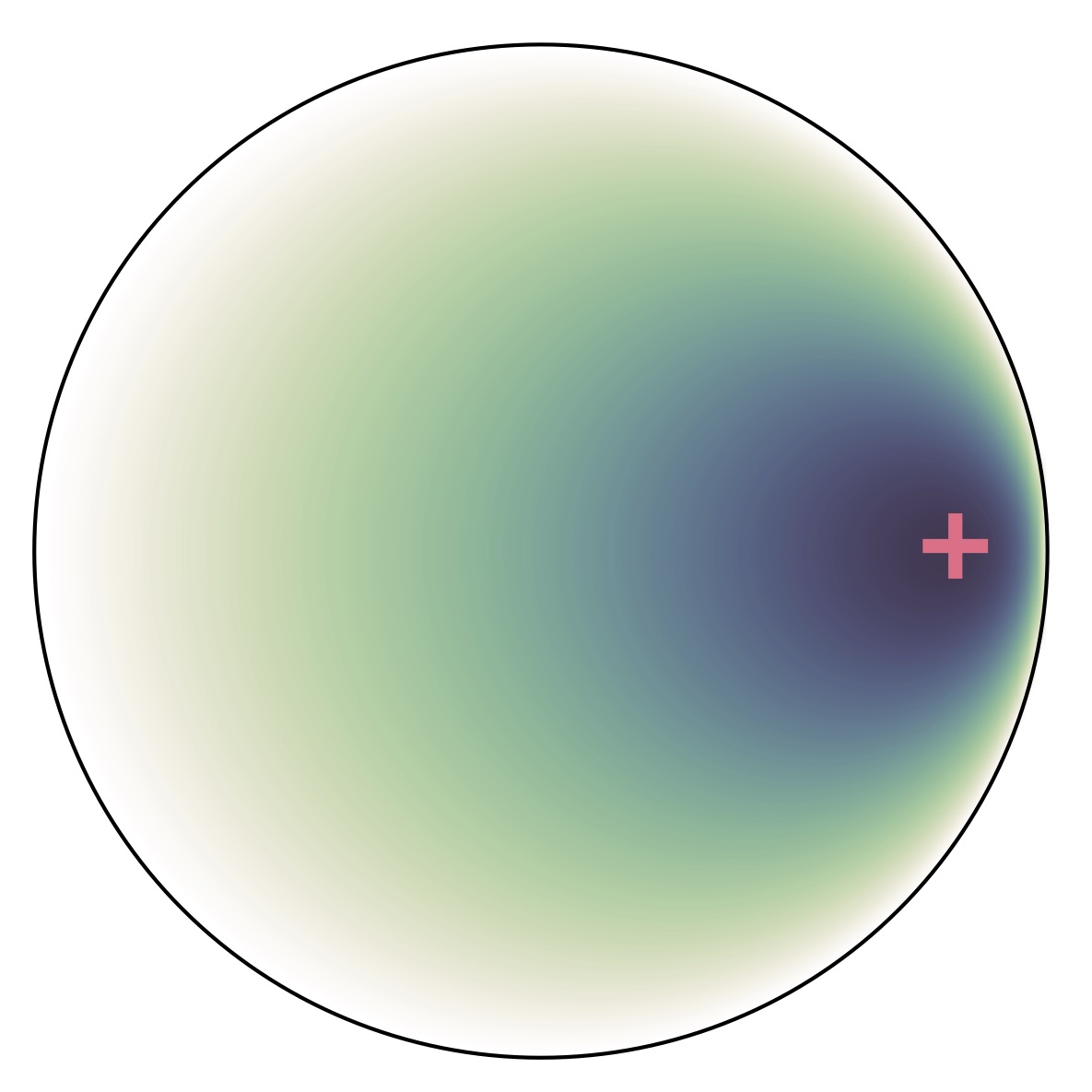}
\put(-80,11){\rotatebox{90}{$\sqrt{c}\|\bm{\mu}\|_2=0.8$}}
\end{subfigure}
\begin{subfigure}{0.17\textwidth}
\includegraphics[width=\linewidth]{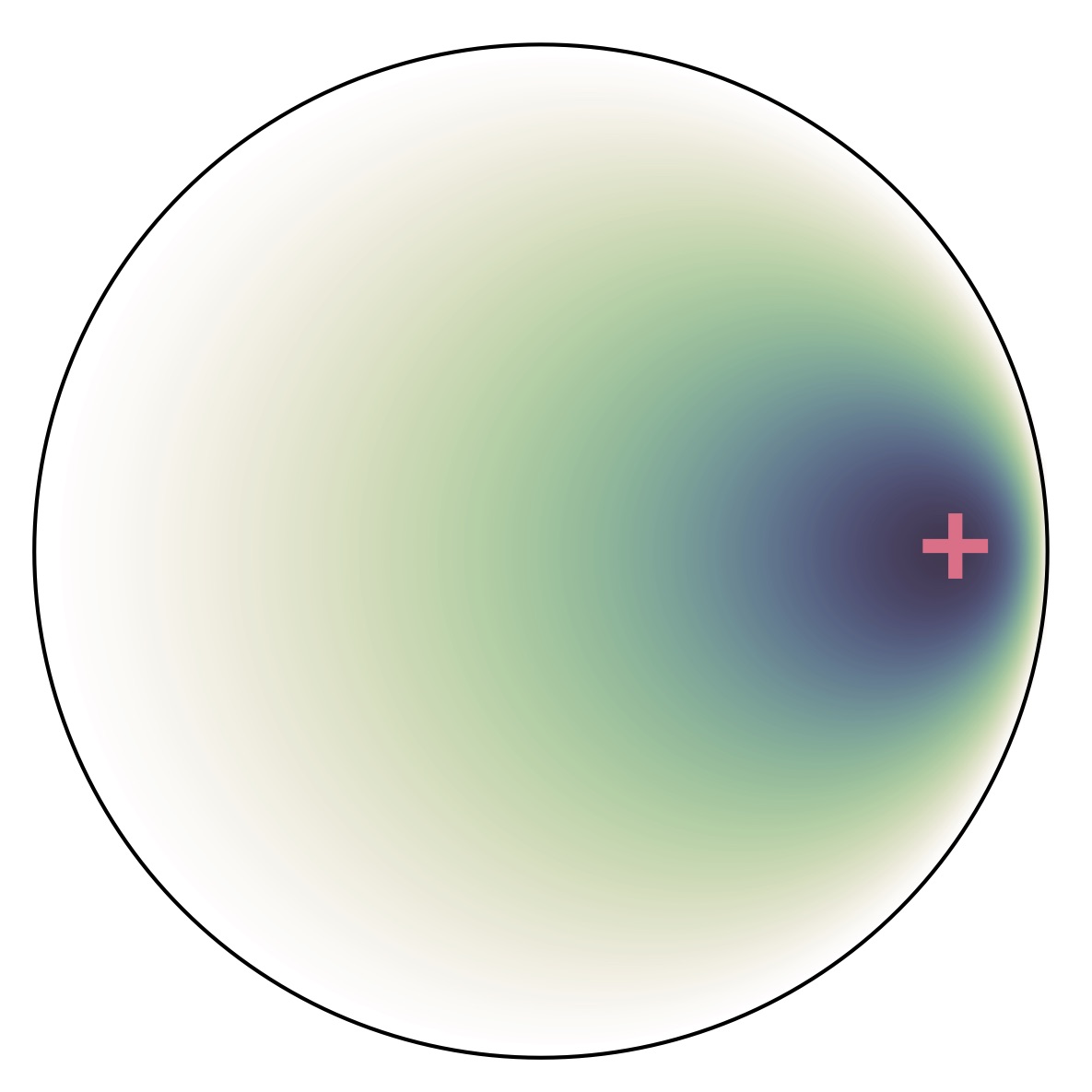}
\end{subfigure}
  \caption{
  Hyperbolic normal probability density for different Fr\'echet mean, same standard deviation and $c=10$.
  The \emph{Riemannian} hyperbolic radius has a slightly larger mode.
  }
  \label{fig:hypp_norm_samples}
  \vspace{-4.em}
\end{wrapfigure}
%
\paragraph{Wrapped normal}
An alternative is to consider the pushforward measure obtained by mapping a normal distribution 
along the \emph{exponential map} $\exp_{\bm{\mu}}$.
That probability measure is often referred to as the \emph{wrapped normal} distribution, and has been used in auto-encoder frameworks with other manifolds \citep{Grattarola:2018ue,Nagano,2018arXiv180704689F}.
Samples $\z \in \B_c^d$ are obtained as
$\z=\exp^c_{\bm{\mu}} \left( {\bm{\v}}/{\lambda^c_{\bm{\mu}}} \right)$ with $\bm{\v} \sim \mathcal{N}(\cdot|\bm{0}, \Sigma)$
%
and its density is given by (details given in Appendix \ref{sec:wrapped_normal})
\begin{align}
\label{eq:core_wrapped_normal_hyp}
&\mathcal{N}^{\text{W}}_{\B_c^d}(\z|\bm{\mu}, \Sigma)   =
 \frac{d\nu^{\text{W}}(\z|\bm{\mu}, \Sigma)}{d\M(\z)} \\ \nonumber
&\hspace{0em}= \mathcal{N} \left(\lambda^c_{\bm{\mu}} \log_{\bm{\mu}}(\z) \middle|\bm{0}, \Sigma \right)
 \left(   \frac{\sqrt{c} ~d_p^c(\bm{\mu}, \z)}{\sinh(\sqrt{c} ~d_p^c(\bm{\mu}, \z))}  \right)^{d-1}.
\end{align}
The (usual) normal distribution is recovered for both generalisations as $c \rightarrow 0$.
We discuss the benefits and drawbacks of those two distributions in Appendix \ref{sec:measures_riem}.
We refer to both as \emph{hyperbolic normal} distributions with pdf $\mathcal{N}_{\B_c^d}(\z|\bm{\mu}, \sigma^2)$.
Figure \ref{fig:hypp_norm_samples} shows several probability densities for both distributions.

%
The prior distribution defined on $\Z$ is chosen to be a hyperbolic normal distribution with mean zero, $p(\z) = \mathcal{N}_{\B^d_c}(\cdot|\bm{0}, \sigma_0^2)$, and the variational family is chosen to be parametrised as $\mathcal{Q}=\{ \mathcal{N}_{\B_c^d}(\cdot|\bm{\mu}, \sigma^2) ~|~ \bm{\mu}\in \B_c^d, \sigma \in \R^+_* \}$.
\subsection{Encoder and decoder}
\label{sec:arch}
\begin{wrapfigure}{r}{0.53\textwidth}
\vspace{-2em}
\begin{center}
  \includegraphics[width=0.26\textwidth, trim={0em 0 .5em 0}, clip]{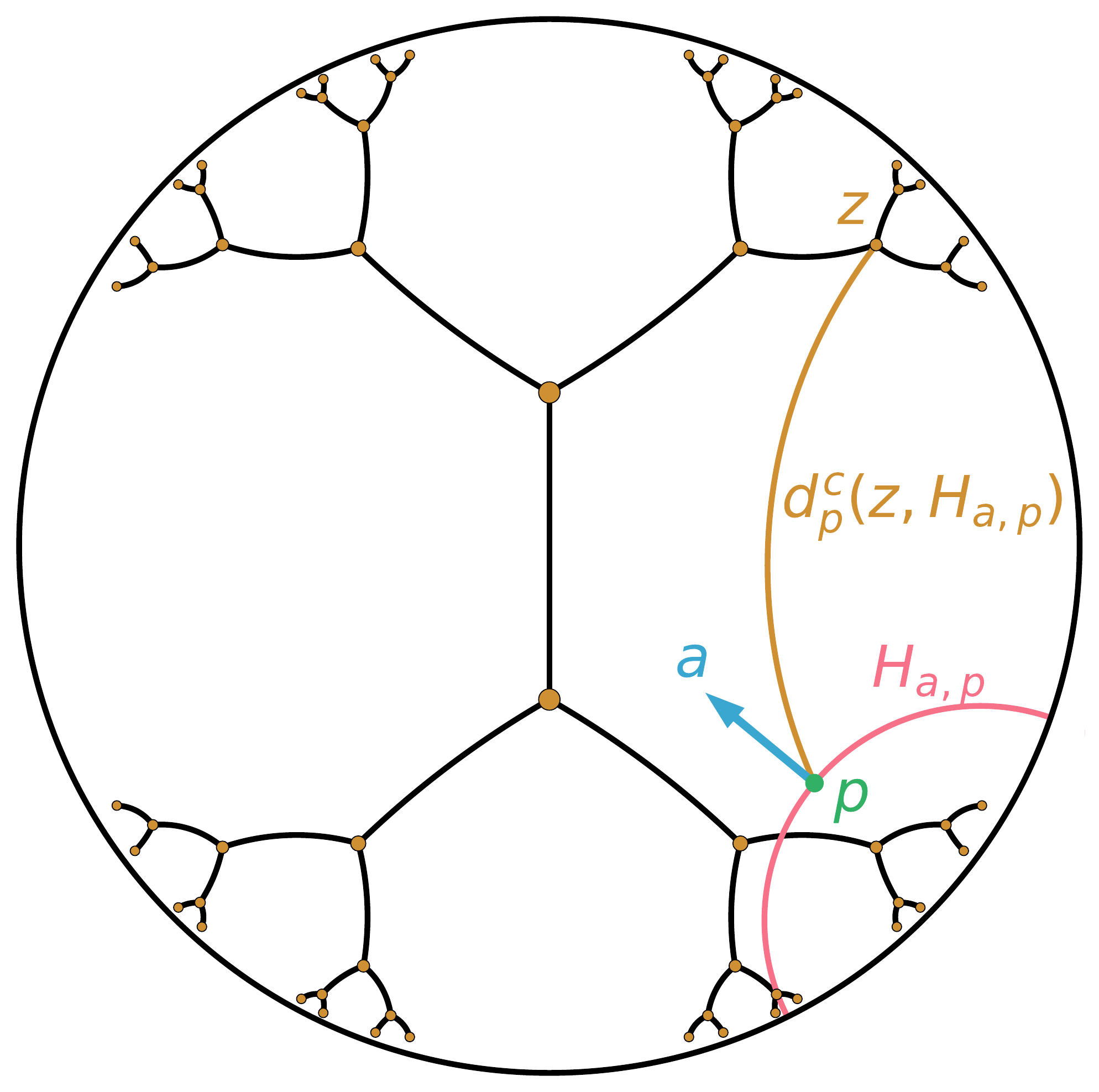}
  \includegraphics[width=0.26\textwidth, trim={16em 12em 6em 5em}, clip]{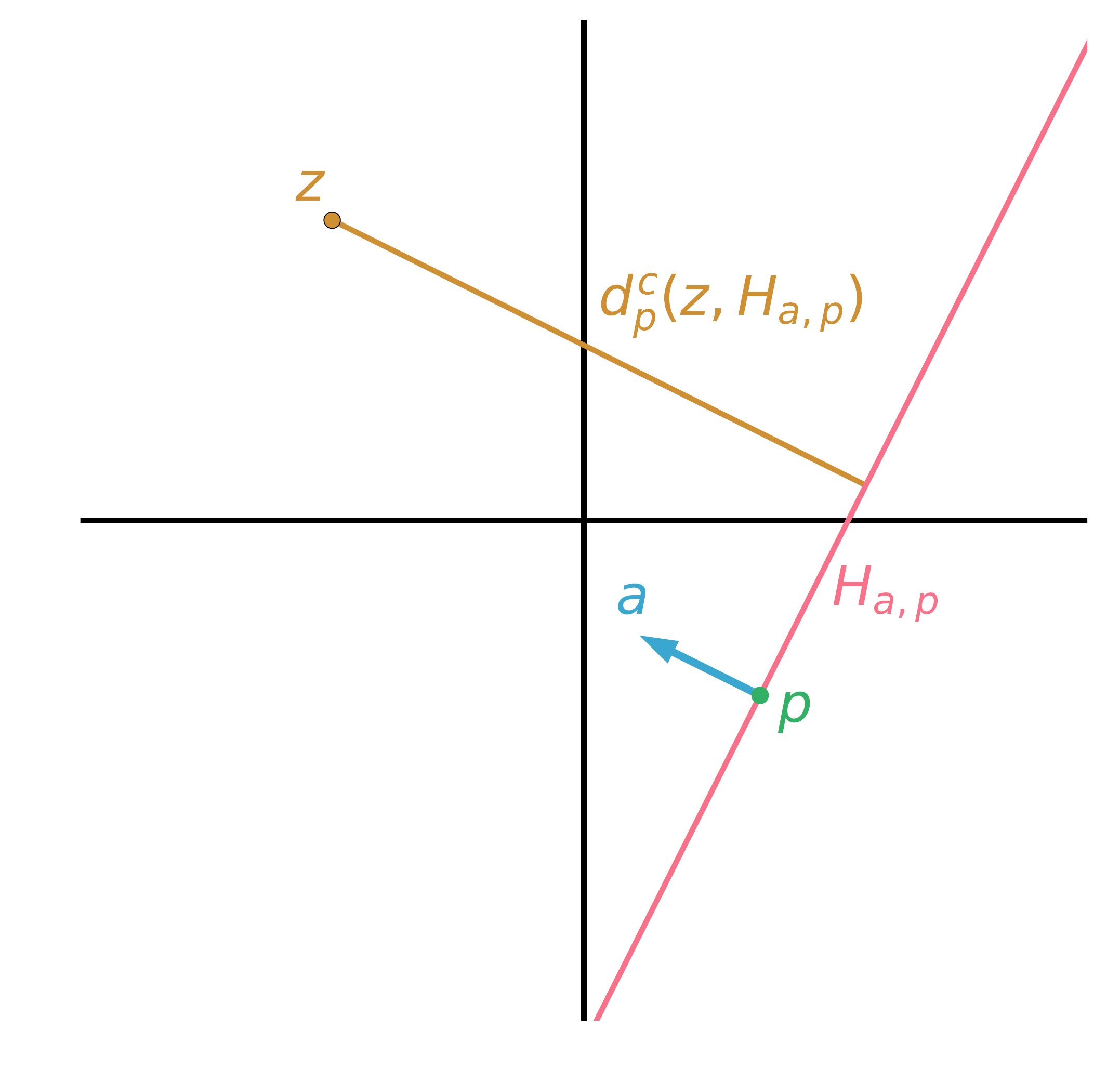}
  \caption{
  Illustration of an orthogonal projection on a hyperplane in a Poincar\'e disc (Left) and an Euclidean plane (Right).
  }

  \label{fig:hyperplane}
\end{center}
\vspace{0.em}
\end{wrapfigure}
%
%
We make use of two neural networks, 
a \emph{decoder} $f_{\bm{\theta}}$ and an \emph{encoder} $g_{\bm{\phi}}$, to parametrise the likelihood 
$p(\cdot|f_{\bm{\theta}}(\z))$ 
and the variational posterior 
$q(\cdot|g_{\bm{\phi}}(\x))$ respectively.
The input of $f_{\bm{\theta}}$ and the output of $g_{\bm{\phi}}$ need to respect the hyperbolic geometry of $\Z$.
In the following we describe appropriate choices for the first layer of the decoder and the last layer of the encoder.
\paragraph{Decoder}
In the Euclidean case, an affine transformation can be written in the form $f_{\a, \p}(\z) = \left\langle\a, \z-\p \right\rangle$, with orientation and offset parameters $\a,\p \in \R^d$.
This can be rewritten in the form
\begin{align*}
&f_{\a, \p}(\z) = \text{sign}(\left\langle\a, \z-\p \right\rangle) \left\|\a\right\| d_E(\z,H^c_{\a,\p})
\end{align*}
where $H_{\a,\p} = \{\z \in \R^p~|~ \left\langle\a,\z-\p \right\rangle = 0 \} = \p + \{ \a \}^{\perp}$ is the decision hyperplane. 
The third term is the distance between $\z$ and the decision hyperplane $H^c_{\a,\p}$ and
the first term refers to the side of $H^c_{\a,\p}$ where $\z$ lies. 
\cite{Ganea:2018wy} analogously introduced an operator $f^c_{\a, \p}: \B_c^d \rightarrow \R^p$ on the Poincar\'e ball,
\begin{align*}
f^c_{\a, \p}(\z) = \text{sign}(\left\langle\a, \log^c_{\p}(\z) \right\rangle_{\p}) \left\|\a\right\|_{\p} d^c_p(\z,H^c_{\a,\p})
\end{align*} 
with $H^c_{\a,\p} = \{\z \in \B_c^d~|~ \left\langle\a,\log^c_{\p}(\z) \right\rangle = 0 \} =\exp^c_{\p}(\{ \a \}^{\perp})$. 
A closed-formed expression for the distance $d^c_p(\z,H^c_{\a,\p})$ was also derived,
$d^c_p(\z,H^c_{\a,\p}) = \frac{1}{\sqrt{c}} \sinh^{-1} \left( \frac{2 \sqrt{c} |\left\langle -\p \oplus_c \z, a \right\rangle |}{(1 - c \| -\p \oplus_c \z\|^2)\|\a\|} \right)$.
The hyperplane decision boundary $H^c_{\a,\p}$ is called \emph{gyroplane} and is a semi-hypersphere orthogonal to the Poincar\'e ball's boundary as illustrated on Figure \ref{fig:hyperplane}.
The decoder's first layer, called \emph{gyroplane} layer, is chosen to be a concatenation of such operators, which are then composed with a standard feed-forward neural network.
\paragraph{Encoder}
The encoder $g_{\bm{\phi}}$ outputs a Fr\'echet mean $\bm{\mu} \in \B_c^d$ and a distortion $\sigma \in \R^+_*$ which parametrise the hyperbolic variational posterior.
The Fr\'echet mean $\bm{\mu}$ is obtained as the image of the exponential map $\exp^c_{\bm{0}}$, and the distortion $\sigma$ through a \emph{softplus} function.
\subsection{Training}
We follow a standard variational approach by deriving a lower bound on the marginal likelihood.
The \gls{ELBO} is optimised via an unbiased \gls{MC} estimator thanks to the reparametrisable sampling schemes that we introduce for both hyperbolic normal distributions.
%
\paragraph{Objective}
The \acrfull{ELBO} can readily be extended to Riemannian latent spaces by applying Jensen's inequality w.r.t. $d\M$ (see Appendix \ref{sec:elbo})
\begin{align}
 \log p(\x) \ge \mathcal{L}_{\M}(\x; \theta, \phi) 
 \triangleq \int_{\M} \ln \left(\frac{p_\theta(\x|\z) p(\z)} {q_\phi(\z|\x)}\right) \ q_\phi(\z|\x) \ d\M(\z). \nonumber 
\end{align}
Densities have been introduced earlier in Equations \ref{eq:core_maxent_normal_hyp} and \ref{eq:core_wrapped_normal_hyp}.
%
\paragraph{Reparametrisation}
\begin{wrapfigure}{R}{0.47\textwidth}
    \vspace{-2em}
    \begin{minipage}{0.47\textwidth}
      \begin{algorithm}[H]
        \caption{Hyperbolic normal sampling scheme
    \label{alg:sampling}}  
        \begin{algorithmic}
          \Require{$\bm{\mu}$, $\sigma^2$, dimension $d$, curvature $c$}
          \If{Wrapped normal}{ $\v \sim \mathcal{N}(\bm{0}_d, \sigma^2)$}
          \ElsIf{Riemannian normal}
            \State Let $g$ be a piecewise exponential proposal
            \While{sample $r$ not accepted}
            \State Propose $r \sim g(\cdot)$, $u \sim \mathcal{U}([0, 1])$
             \If{$u < \frac{\rho^{\text{R}}(r)}{g(r)}$} { Accept sample $r$}
             \EndIf
            \EndWhile
            \State Sample direction $\bm{\alpha} \sim \mathcal{U}(\mathbb{S}^{d-1})$
            \State $\v \gets r \bm{\alpha}$
          \EndIf
          \State Return $\z = \exp^c_{\bm{\mu}}\left( {\v}/{\lambda^c_{\bm{\mu}}}\right)$
        \end{algorithmic}
      \end{algorithm}
    \end{minipage}
    \vspace{-1.em}
  \end{wrapfigure}
%
%
In the Euclidean setting, by working in polar coordinates, an isotropic normal distribution centred at $\bm{\mu}$ can be described by a directional vector $\bm{\alpha}$ uniformly distributed on the hypersphere and a univariate radius $r=d_E(\bm{\mu}, \z)$ following a ${\chi}$-distribution.
In the Poincar\'e ball we can rely on a similar representation, through a \emph{hyperbolic polar} change of coordinates, given by
\begin{align}
\z = \exp^c_{\bm{\mu}} \left( G(\bm{\mu})^{-\frac{1}{2}} ~\v \right) = \exp^c_{\bm{\mu}} \left( \frac{r}{\lambda^c_{\bm{\mu}}} \bm{\alpha} \right)
\label{eq:main_reparameterisaition}
\end{align}
with $\v = r\bm{\alpha}$ and $r=d_p^c(\bm{\mu}, \z)$.
The direction $\bm{\alpha}$ is still uniformly distributed on the hypersphere and for the \emph{wrapped normal}, the radius $r$ is still ${\chi}$-distributed,
while for the \emph{Riemannian normal} its density $\rho^{\text{R}}(r)$ is given by (derived in Appendix \ref{sec:riemannian_reparametrisation})
\begin{align*}
\rho^{\text{W}}(r) \propto {\1_{\R_+}(r)} ~e^{- \frac{r^2}{2\sigma^2}} r^{d-1},
\quad \rho^{\text{R}}(r) \propto {\1_{\R_+}(r)} e^{- \frac{r^2}{2\sigma^2}} \left(\frac{\sinh(\sqrt{c}r)}{\sqrt{c}} \right)^{d-1}.
\end{align*}
The latter density $\rho^{\text{R}}(r)$ can efficiently be sampled via rejection sampling with a piecewise exponential distribution proposal. This makes use of its log-concavity.
The \emph{Riemannian normal} sampling scheme is not directly affected by dimensionality since the radius is a \emph{one-dimensional} random variable. 
Full sampling schemes are described in Algorithm \ref{alg:sampling}, and in Appendices \ref{sec:riemannian_reparametrisation} and \ref{sec:sampling}.
\paragraph{Gradients}
Gradients $\nabla_{\bm{\mu}} \z$ can straightforwardly be computed thanks
to the exponential map reparametrisation (Eq \ref{eq:main_reparameterisaition}), and
gradients w.r.t. the dispersion $\nabla_{\sigma} \z$ are readily available for the \emph{wrapped normal}.
For the \emph{Riemannian normal}, we additionally rely on an implicit reparametrisation \citep{Figurnov:2018vr} of $\rho^{\text{R}}$ via its cdf $F^{\text{R}}(r;\sigma)$.

\paragraph{Optimisation}
Parameters of the model living in the Poincar\'e ball are parametrised via the exponential mapping: $\bm{\phi}_i = \exp^c_{\bm{0}}(\bm{\phi}^{0}_i)$ with $\bm{\phi}^{0}_i \in \R^m$, so we can make use of usual optimisation schemes.
Alternatively, one could directly optimise such manifold parameters with manifold gradient descent schemes \citep{Bonnabel:2011bl}.


\section{Related work}
\label{sec:related_work}

\paragraph{Hierarchical models}
The \acrlong{BNP} literature has a rich history of explicitly modelling the hierarchical structure of data \citep{NIPS2007_3266,Heller:2005:BHC:1102351.1102389,NIPS2003_2466,NIPS2010_4108,Larsen2001ProbabilisticHC,Salakhutdinov2011OneShotLW}.
The discrete nature of trees used in such models makes learning difficult, whereas performing optimisation in a continuous hyperbolic space is an attractive alternative.
Such an approach has been empirically and theoretically shown to be useful for graphs and word embeddings \citep{Nickel:2017wz,Nickel:2018vw,Chamberlain:2017wm,DeSa:2018we,Tifrea:2018ty}.
\paragraph{Distributions on manifold}
Probability measures defined on manifolds are of interest to model uncertainty of data living (either intrinsically or assumed to) on such spaces, e.g.  directional statistics \citep{directional_stats,mardia2009directional}.
\cite{Pennec2006} introduced a maximum entropy generalisation of the normal distribution, often referred to as \emph{Riemannian normal}, which has been used for maximum likelihood estimation in the Poincar\'e half-plane \citep{DBLP:journals/entropy/SaidBB14}
and on the hypersphere \citep{Hauberg}.
Another class of manifold probability measures are \emph{wrapped} distributions, i.e. push-forward of distributions defined on a tangent space, often along the exponential map.
They have recently been used in auto-encoder frameworks on the \emph{hyperboloid} model (of hyperbolic geometry) \citep{Grattarola:2018ue,Nagano} and on Lie groups \citep{2018arXiv180704689F}.
\cite{Rey:2019ty,Li:2019wa} proposed to parametrise a variational family through a Brownian motion on manifolds such as spheres, tori, projective spaces and $SO(3)$.
\paragraph{\Glspl{VAE} with Riemannian latent manifold}
\Glspl{VAE} with non Euclidean latent space have been recently introduced, such as \cite{Davidson:2018vg} making use of hyperspherical geometry and \cite{2018arXiv180704689F} endowing the latent space with a $\text{SO}(3)$ group structure.
Concurrent work considers endowing \glspl{AE} with a hyperbolic latent space. 
\cite{Grattarola:2018ue} introduces a \gls{CCM} (i.e. hyperspherical, Euclidean and hyperboloid) latent space within an adversarial auto-encoder framework.
However, the encoder and decoder are not designed to explicitly take into account the latent space geometry. 
\cite{Ovinnikov:2018ue} recently proposed to endow a \gls{VAE} latent space with a Poincar\'e ball model.
They choose a Wasserstein Auto-Encoder framework \citep{Tolstikhin:2017wy} because they could not derive a closed-form solution of the \gls{ELBO}'s entropy term.
We instead rely on a \gls{MC} estimate of the \gls{ELBO} by introducing a novel reparametrisation of the Riemannian normal.
They discuss the \emph{Riemannian} normal distribution, yet they make a number of heuristic approximations for sampling and reparametrisation.
Also, \cite{Nagano} propose using a \emph{wrapped} normal distribution to model uncertainty on the \emph{hyperboloid} model of hyperbolic space.
They derive its density and a reparametrisable sampling scheme, allowing such a distribution to be used in a variational learning framework.
They apply this \emph{wrapped} normal distribution to stochastically embed graphs and to parametrise the variational family in \glspl{VAE}.
\cite{Ovinnikov:2018ue} and \cite{Nagano} rely on a standard feed-forward decoder architecture, which does not take into account the hyperbolic geometry.
%


\section{Experiments}
\label{sec:experiments}
We implemented our model and ran our experiments within the automatic differentiation framework PyTorch \citep{paszke2017automatic}.
We open-source our code for reproducibility and to benefit the community \footnote{\url{https://github.com/emilemathieu/pvae}}.
Experimental details are fully described in Appendix \ref{sec:exp_details}.
\subsection{Branching diffusion process}
We assess our modelling assumption on data generated from a branching diffusion process which explicitly incorporate hierarchical structure.
Nodes $\y_i \in \R^n$ are normally distributed with mean given by their parent and with unit variance.
Models are trained on a noisy vector representations $(\x_1, \dots, \x_N)$, hence do not have access to the true hierarchical representation.
\begin{figure}[h]
\centering
\begin{subfigure}{0.25\textwidth}
  \includegraphics[width=\textwidth, trim={1.64cm 1.00cm 0.60cm 0.54cm}, clip]{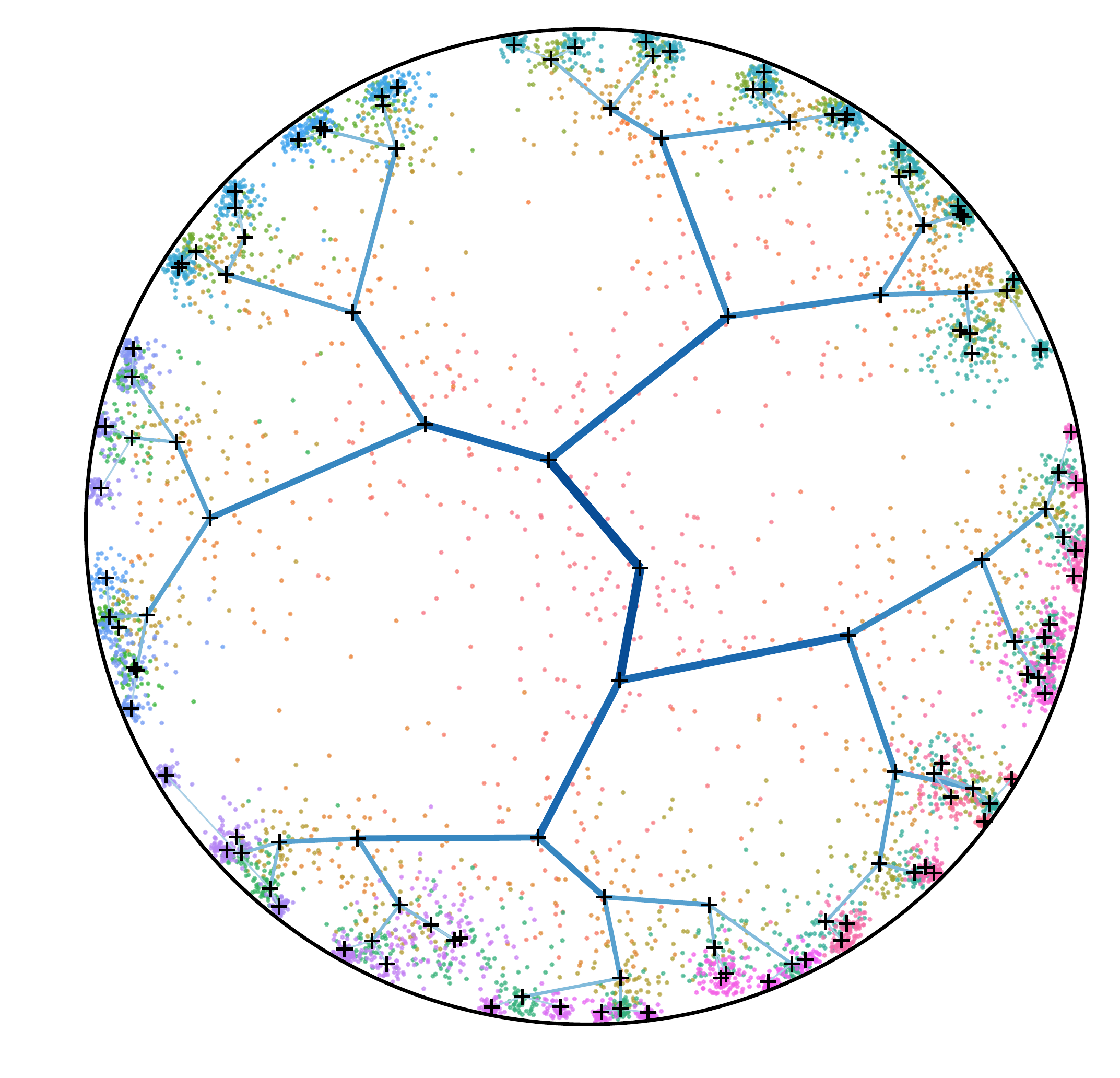}
\end{subfigure}\hfil
\begin{subfigure}{0.25\textwidth}
  \includegraphics[width=\textwidth, trim={2em 5em 5em 2em}, clip]{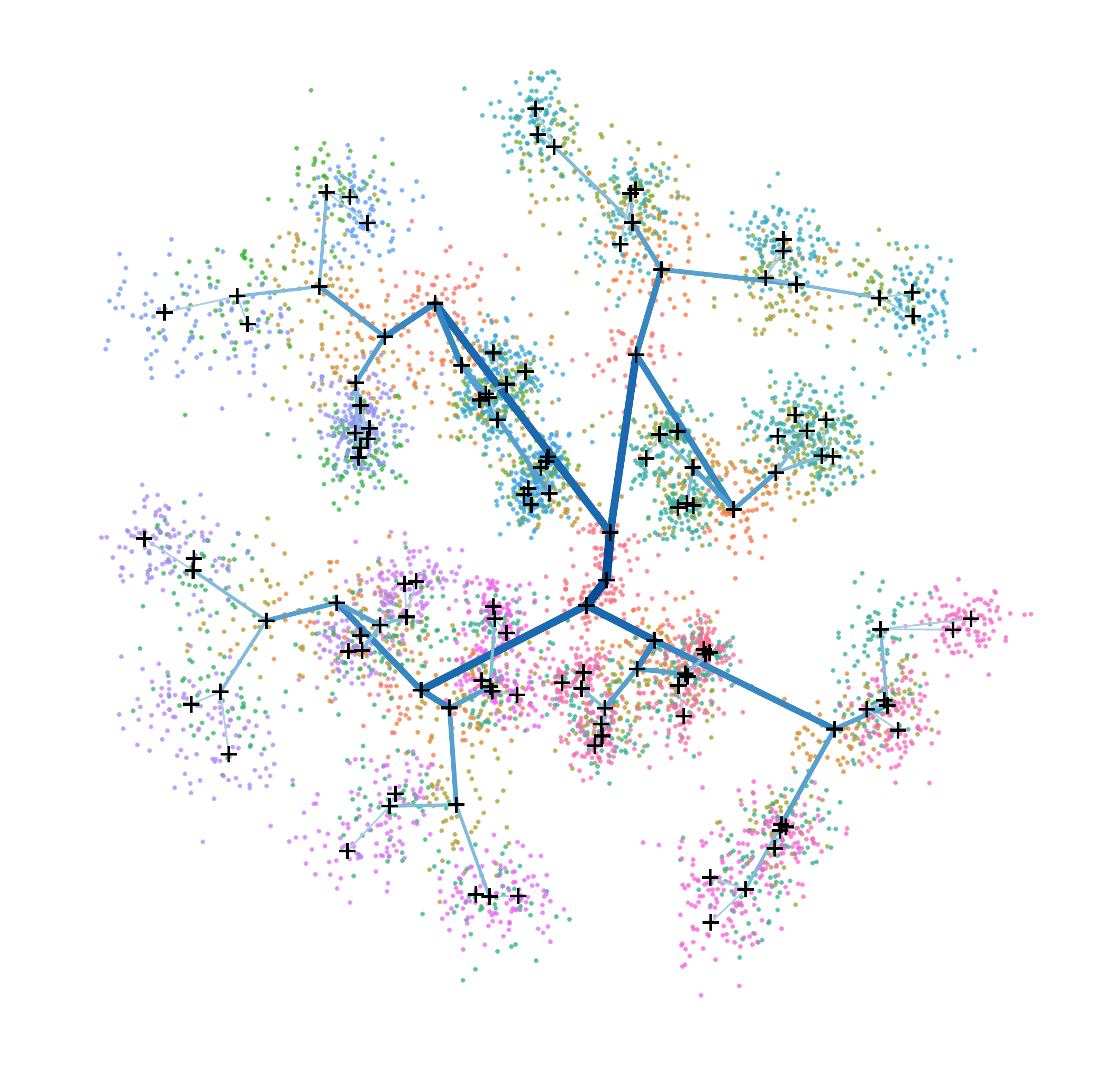}
\end{subfigure}\hfil
\begin{subfigure}{0.25\textwidth}
\includegraphics[width=\textwidth, trim={2em 2em 2em 2em}, clip]{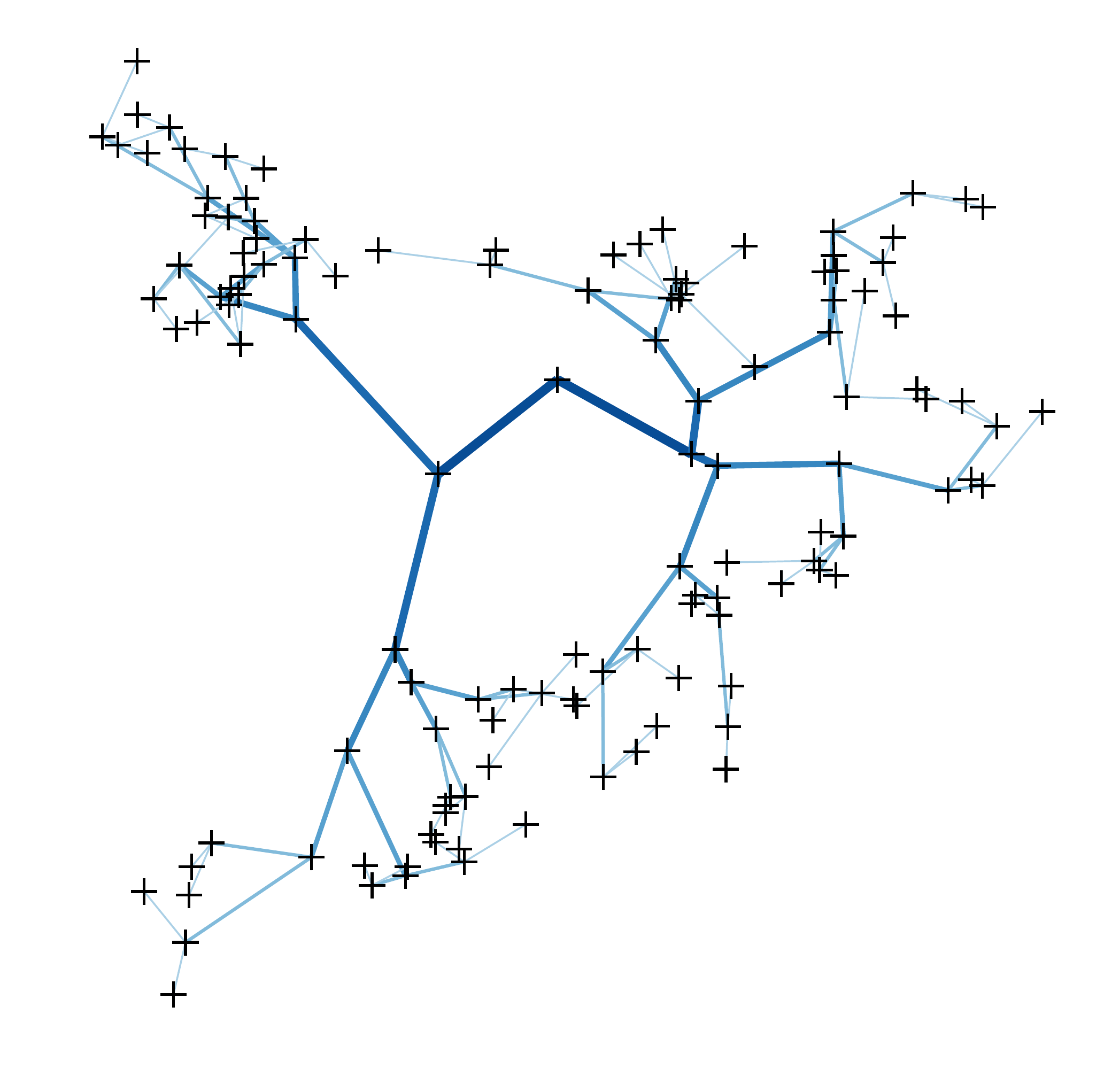}
\end{subfigure}\hfil
\begin{subfigure}{0.25\textwidth}
 \includegraphics[width=\textwidth, trim={2em 2em 2em 2em}, clip]{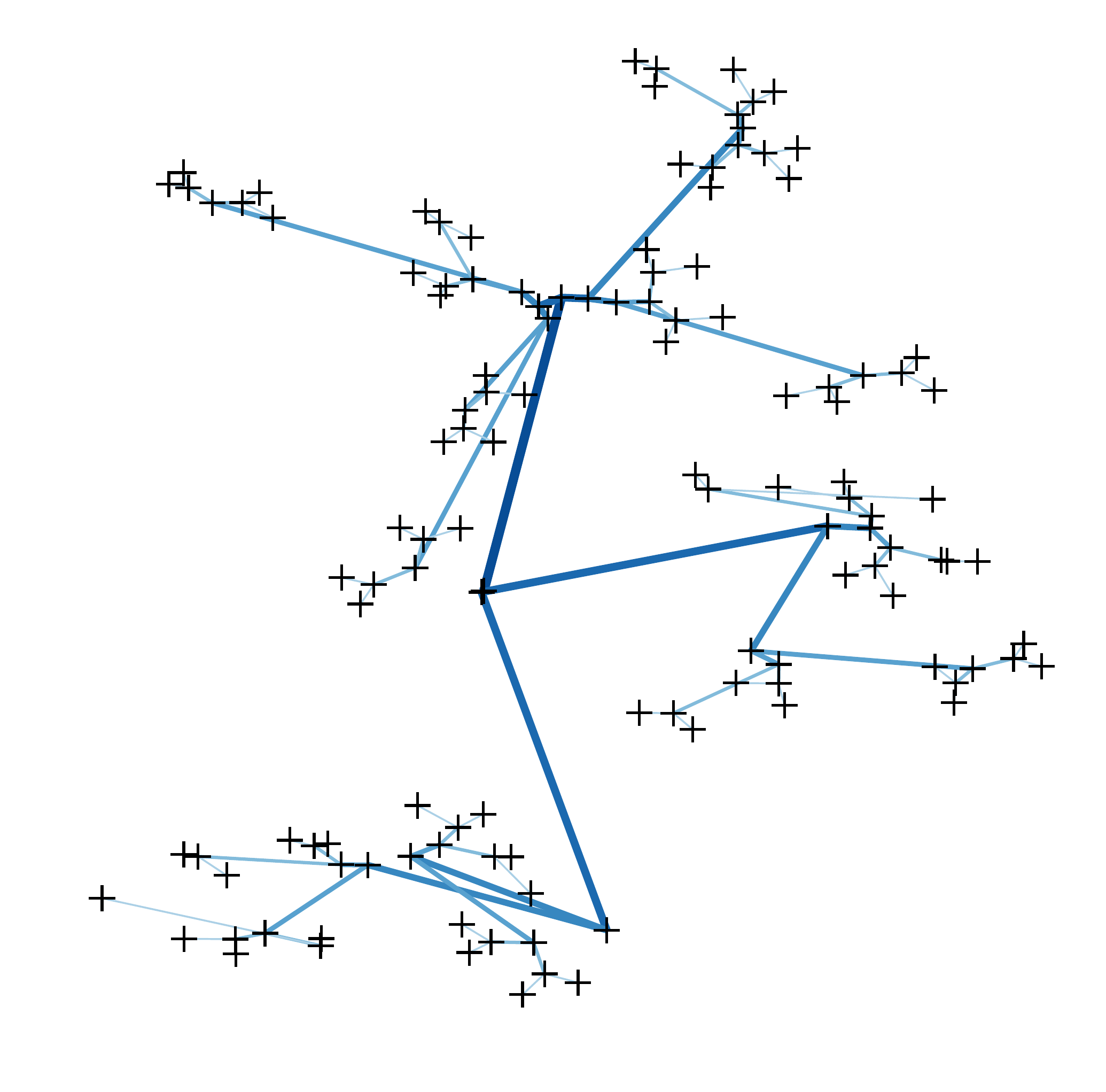}
 \end{subfigure}
  \caption{
  Latent representations learned by -- \pvae{1} (Leftmost), \nvae{} (Center-Left), \acrshort{PCA} (Center-Right) and \acrshort{GPLVM} (Rightmost) trained on synthetic dataset.
  Embeddings are represented by black crosses, and colour dots are posterior samples.
  Blue lines represent true hierarchy.
  }
  \label{fig:synthetic_posterior}
\vspace{-1em}
\end{figure}
We train several \pvae{c}s with increasing curvatures, along with a vanilla \nvae{} as a baseline.
Table \ref{table:curvatures} shows that the \pvae{c} outperforms its Euclidean counterpart in terms of test marginal likelihood.
As expected, we observe that the performance of the \nvae{} is recovered as the curvature $c$ tends to zero.
Also, we notice that increasing the prior distribution distortion $\sigma_0$ helps embeddings lie closer to the border, and as a consequence improved generalisation performance.
Figure \ref{fig:synthetic_posterior} represents latent embeddings for \pvae{1} and \nvae{}, along with two embedding baselines: \gls{PCA} and a \gls{GPLVM}.
A hierarchical structure is somewhat learned by all models, yet \pvae{c}'s latent representation is the least distorted.
\begin{center}
\vspace{-1em}
\begin{table}[h]
\caption{
Negative test marginal likelihood estimates  $\mathcal{L}_{\acrshort{IWAE}}$ \citep{Burda:2015ti} (computed with $5000$ samples) on the synthetic dataset.
95\% confidence intervals are computed over 20 trainings. 
}
\label{table:curvatures}
\centering
\begin{tabular}{@{}llllllll@{}}
\toprule
                 & & \multicolumn{5}{l}{Models} \\ \cmidrule(l){3-8} 
                 & $\sigma_0$ &   \nvae{}    & \pvae{0.1}     & \pvae{0.3}  & \pvae{0.8} &\pvae{1.0} & \pvae{1.2}  \\ \midrule
 $\mathcal{L}_{\acrshort{IWAE}}$& $1$  & $57.1_{\pm0.2}$ & $57.1_{\pm0.2}$ & $57.2_{\pm0.2}$ & $56.9_{\pm0.2}$ & $56.7_{\pm0.2}$ & $56.6_{\pm0.2}$ \\ \midrule
 $\mathcal{L}_{\acrshort{IWAE}}$& $1.7$ &$57.0_{\pm0.2} $ & $56.8_{\pm0.2}$  & $56.6_{\pm0.2}$ & $55.9_{\pm0.2}$ & $55.7_{\pm0.2}$  & $\bm{55.6}_{\pm0.2}$   \\
 \bottomrule
\end{tabular}
\end{table}
\vspace{-1.5em}
\end{center}
\subsection{Mnist digits}
The MNIST \citep{lecun-mnisthandwrittendigit-2010} dataset has been used in the literature for hierarchical modelling \citep{Salakhutdinov2011OneShotLW,Saha:2018tq}.
One can view the natural clustering in MNIST images as a hierarchy with each of the 10 classes being internal nodes of the hierarchy.
We empirically assess whether our model can take advantage of such simple underlying hierarchical structure, first by measuring its generalisation capacity via the test marginal log-likelihood.
Table \ref{table:mnist} shows that our model outperforms its Euclidean counterpart, especially for low latent dimension.
This can be interpreted through an information bottleneck perspective; as the latent dimensionality increases, the pressure on the embeddings quality decreases, hence the gain from the hyperbolic geometry is reduced (as observed by \cite{Nickel:2017wz}).
Also, by using the \emph{Riemannian normal} distribution, we achieve slightly better results than with the \emph{wrapped normal}.
\begin{table}[h!]
\vspace{-1em}
\caption{Negative test marginal likelihood estimates computed with 5000 samples.
 95\% confidence intervals are computed over 10 runs.
 * indicates numerically unstable settings.
}
\label{table:mnist}
\centering
\begin{tabular}{llllll}
\toprule
                                      & & \multicolumn{4}{l}{Dimensionality} \\ \cmidrule{3-6} 
                                      & c & 2      & 5      & 10      & 20     \\ \cmidrule{1-6} 
\bf{\nvae{}}                               &(0) &   $144.5_{\pm0.4}$     &    $114.7_{\pm0.1}$    &    $100.2_{\pm0.1}$     &    $97.6_{\pm0.1}$    \\ \cmidrule{1-6}  
\multirow{3}{*}{\bf{\pvae{} (Wrapped)}}& $0.1$ &   $143.9_{\pm0.5}$     &    $115.5_{\pm0.3}$    &    $100.2_{\pm0.1}$     &    $97.2_{\pm0.1}$   \\   
                                  &$0.2$ &   $144.2_{\pm0.5} $     &    $115.3_{\pm0.3}$    &    $100.0_{\pm0.1}$     &    $97.1_{\pm0.1}$   \\   
                                  & $0.7$ &   $ 143.8_{\pm0.6}$     &    $115.1_{\pm0.3}$    &    $100.2_{\pm0.1}$     &    $97.5_{\pm0.1}$   \\  
                                      & $1.4$  &   $144.0_{\pm0.6}$     &    $114.7_{\pm0.1}$    &    $100.7_{\pm0.1}$     &    $98.0_{\pm0.1}$    \\ \cmidrule{1-6} 

\multirow{3}{*}{\bf{\pvae{} (Riemannian)}}& $0.1$ &   $143.7_{\pm0.6}$     &    $115.2_{\pm0.2}$    &    $99.9_{\pm0.1}$  &   $\bm{97.0}_{\pm0.1}$\\  
                                    & $0.2$ &   $143.8_{\pm0.4}$     &    $114.7_{\pm0.3}$    &    $\bm{99.7}_{\pm0.1}$     &    $97.4_{\pm0.1}$\\ 
                                      & $0.7$ &   $143.1_{\pm0.4}$     &    $\bm{114.1}_{\pm0.2}$    &    $101.2_{\pm0.2}$     &  \hspace{2em}*  \\ 
                                   
                                      & $1.4$  &   $\bm{142.5}_{\pm0.4}$     &    $115.5_{\pm0.3}$    &\hspace{2em}*   &   \hspace{2em}*     \\ \bottomrule
\end{tabular}
\end{table}
\begin{wrapfigure}{r}{0.45\textwidth}
\vspace{-1.em}
\begin{center}
  \includegraphics[width=0.45\textwidth]{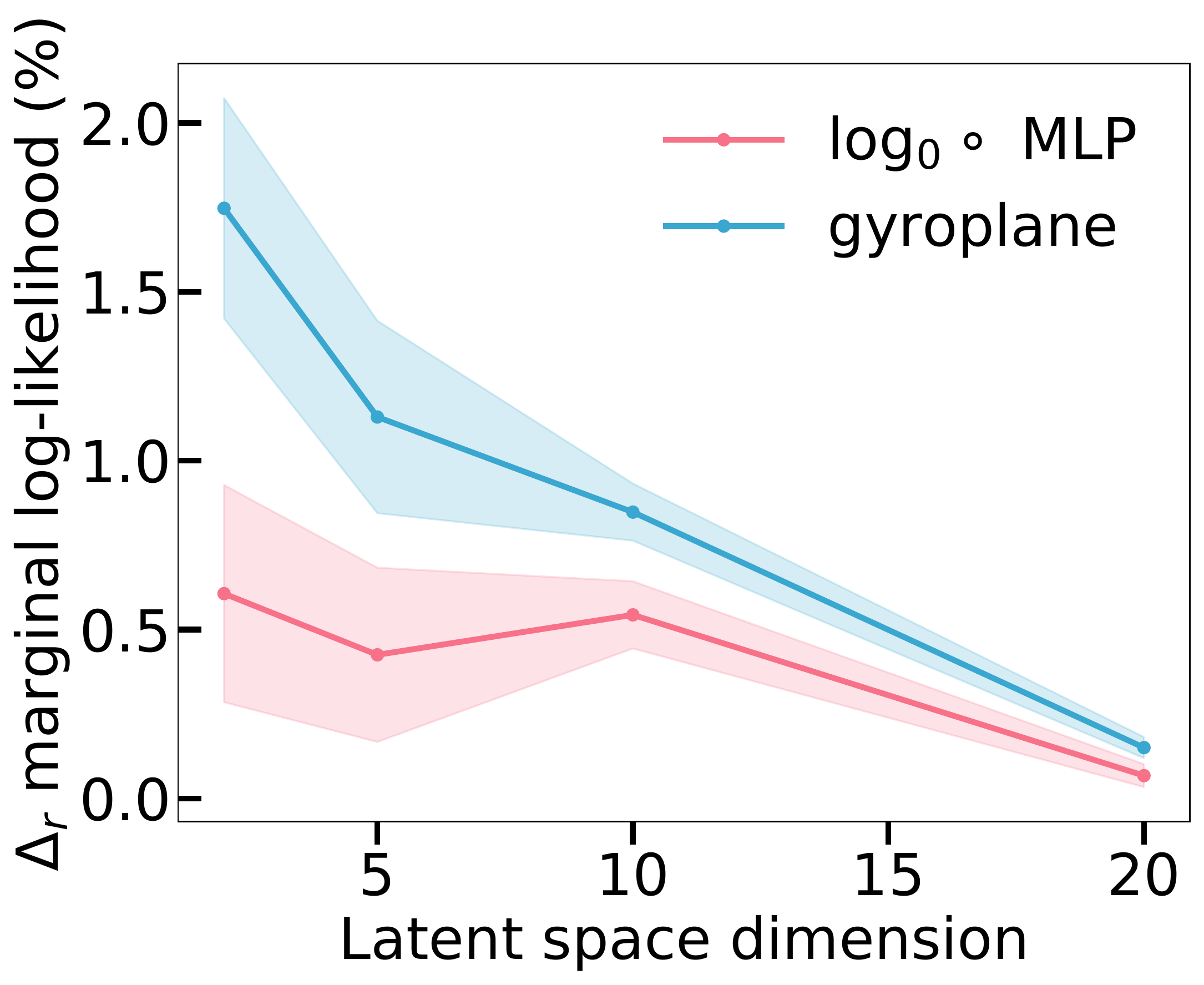}
  \caption{
  Decoder ablation study on MNIST with \emph{wrapped} normal \pvae{1}.
  Baseline decoder is a MLP.
  }
  \label{fig:decoder_ablation}
\end{center}
\vspace{-1.5em}
\end{wrapfigure}
We conduct an ablation study to assess the usefulness of the \emph{gyroplane} layer introduced in Section \ref{sec:arch}.
To do so we estimate the test marginal log-likelihood for different choices of decoder.
We select a \gls{MLP} to be the baseline decoder.
We additionally compare to a \gls{MLP} pre-composed by $\log_{\bm{0}}$, which can be seen as a linearisation of the space around the centre of the ball.
Figure \ref{fig:decoder_ablation} shows the relative performance improvement of decoders over the \gls{MLP} baseline w.r.t. the latent space dimension.
We observe that \emph{linearising} the input of a \gls{MLP} through the logarithm map slightly improves generalisation, and that using a \emph{gyroplane} layer as the first layer of the decoder additionally improves generalisation.
Yet, these performance gains appear to decrease as the latent dimensionality increases.

\begin{figure}[h]
\vspace{-1em}
\centering
\begin{subfigure}{0.28\textwidth}
  \includegraphics[width=\linewidth]{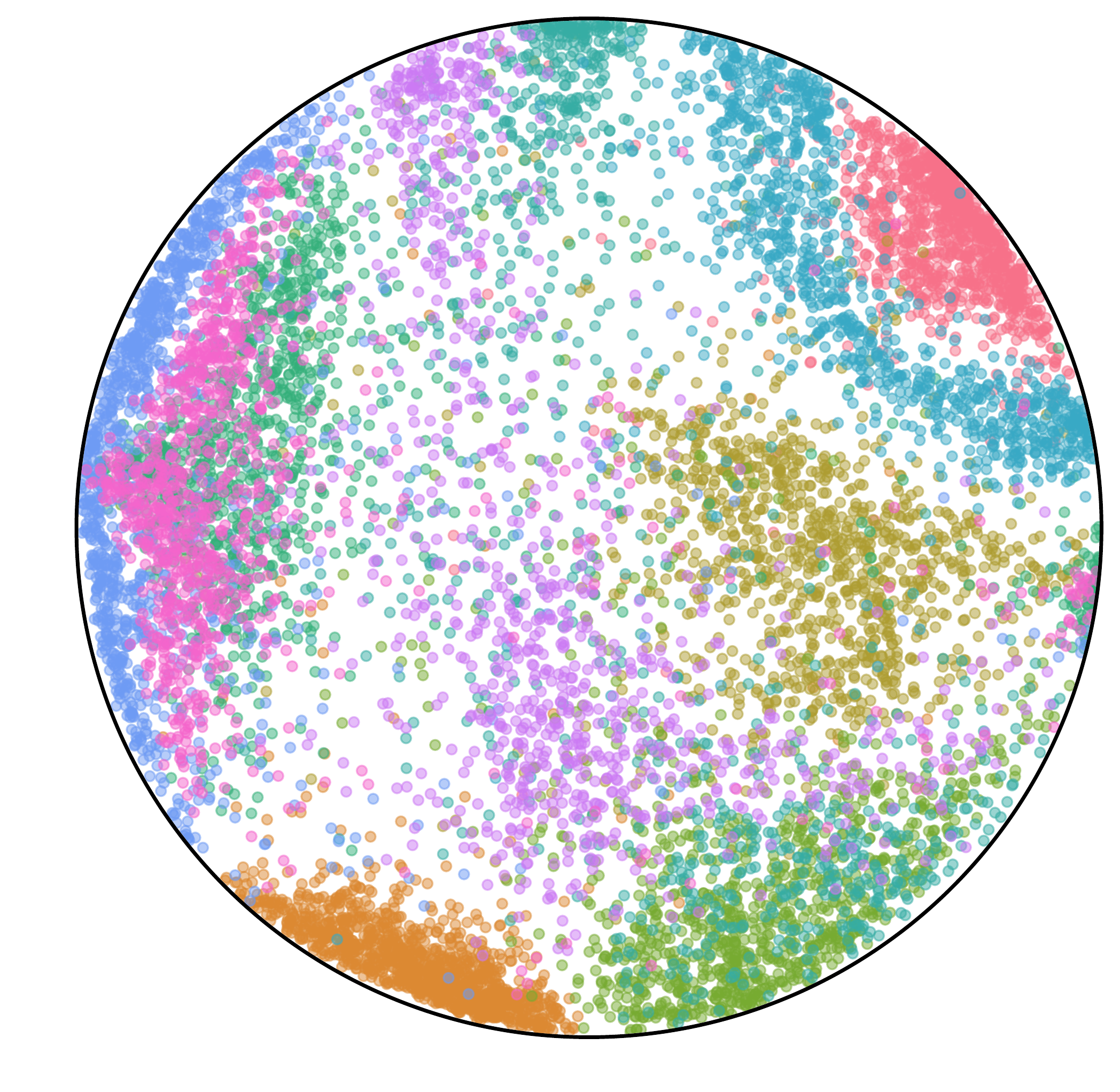}
\end{subfigure}\hfil
\begin{subfigure}{0.28\textwidth}
  \includegraphics[width=\linewidth]{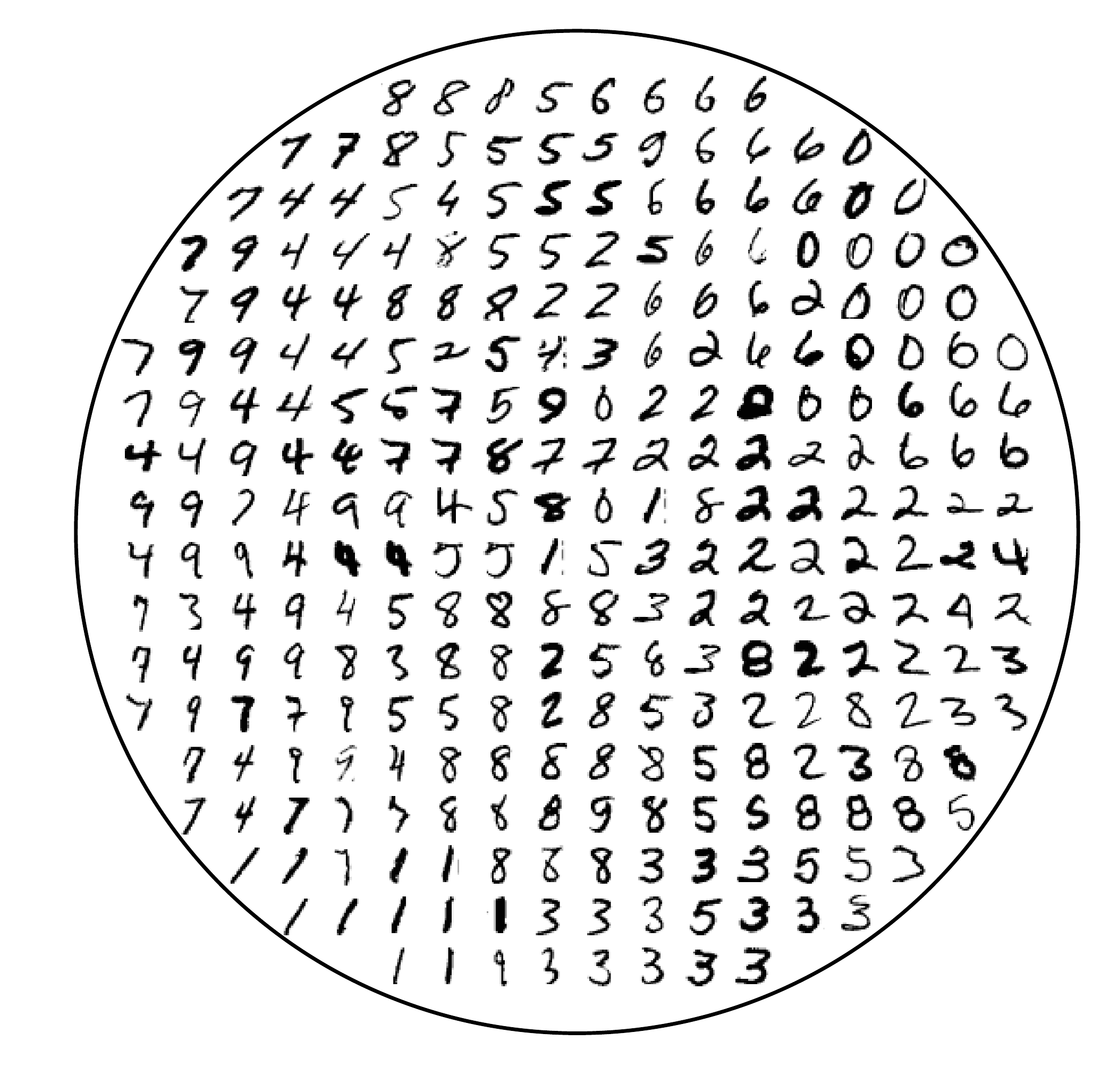}
\end{subfigure}\hfil
\begin{subfigure}{0.28\textwidth}
  \includegraphics[width=\linewidth]{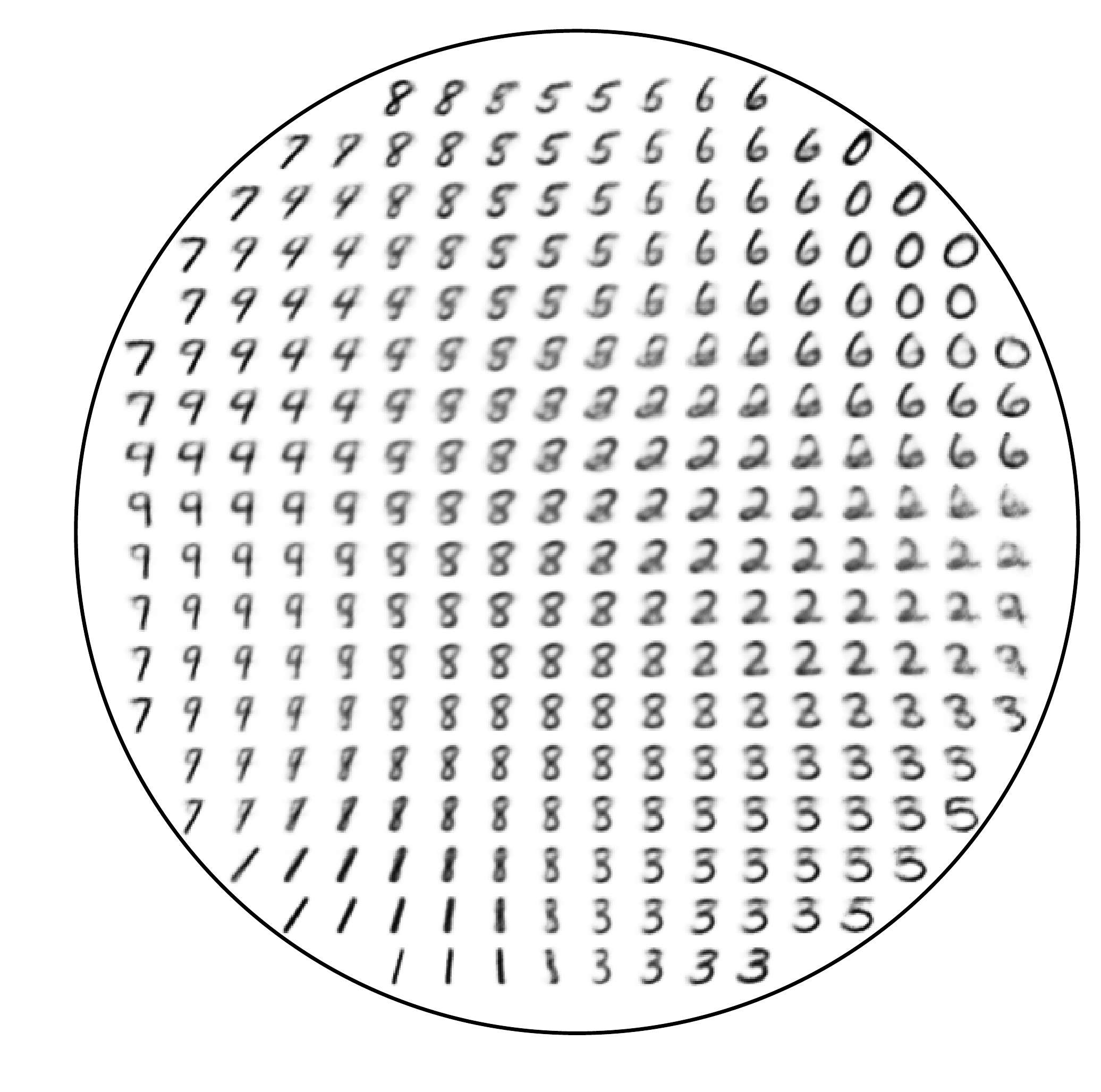}
\end{subfigure}

\medskip
\begin{subfigure}{0.28\textwidth}
\includegraphics[width=\linewidth]{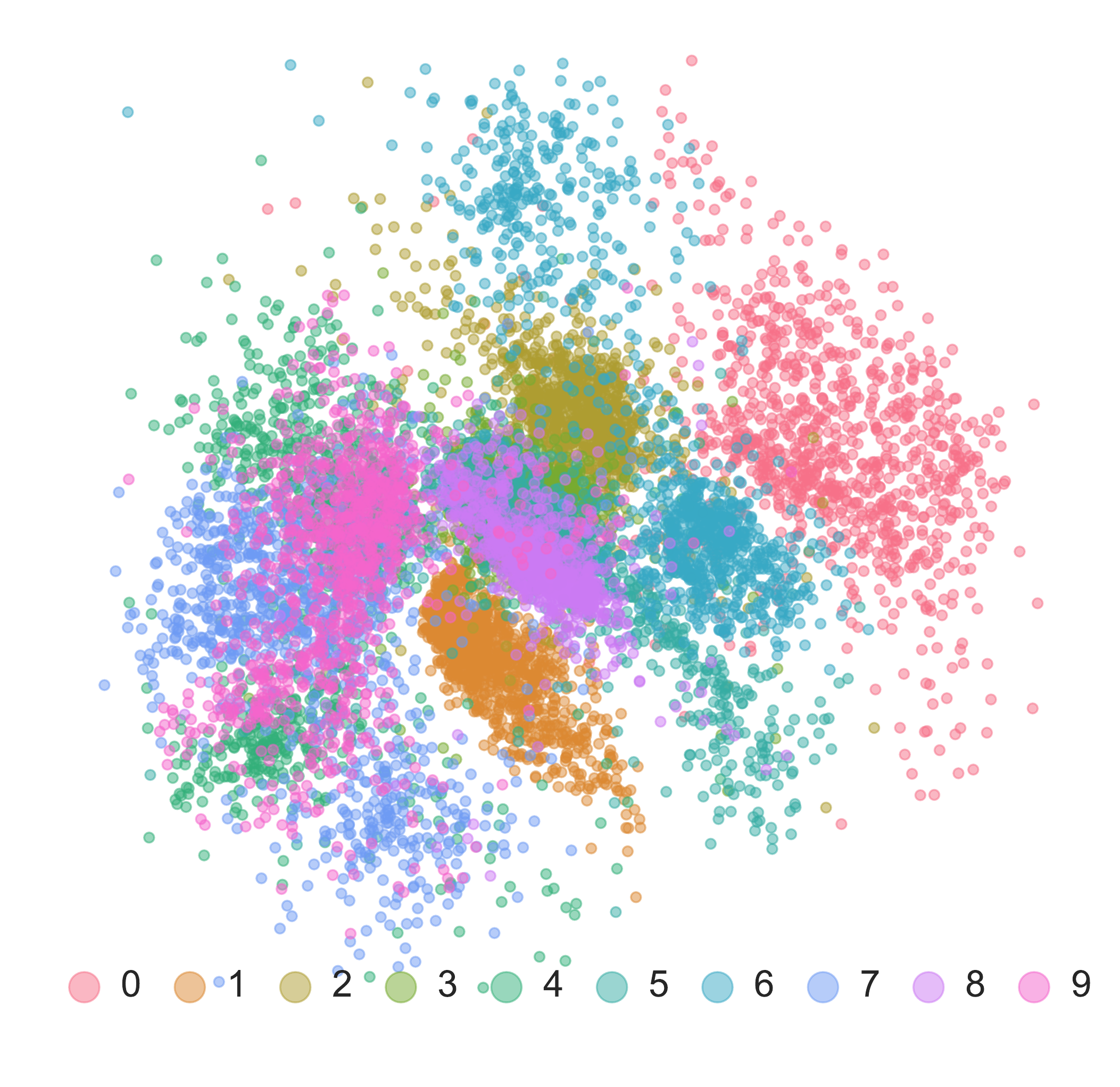}
\end{subfigure}\hfil
\begin{subfigure}{0.28\textwidth}
 \includegraphics[width=\linewidth]{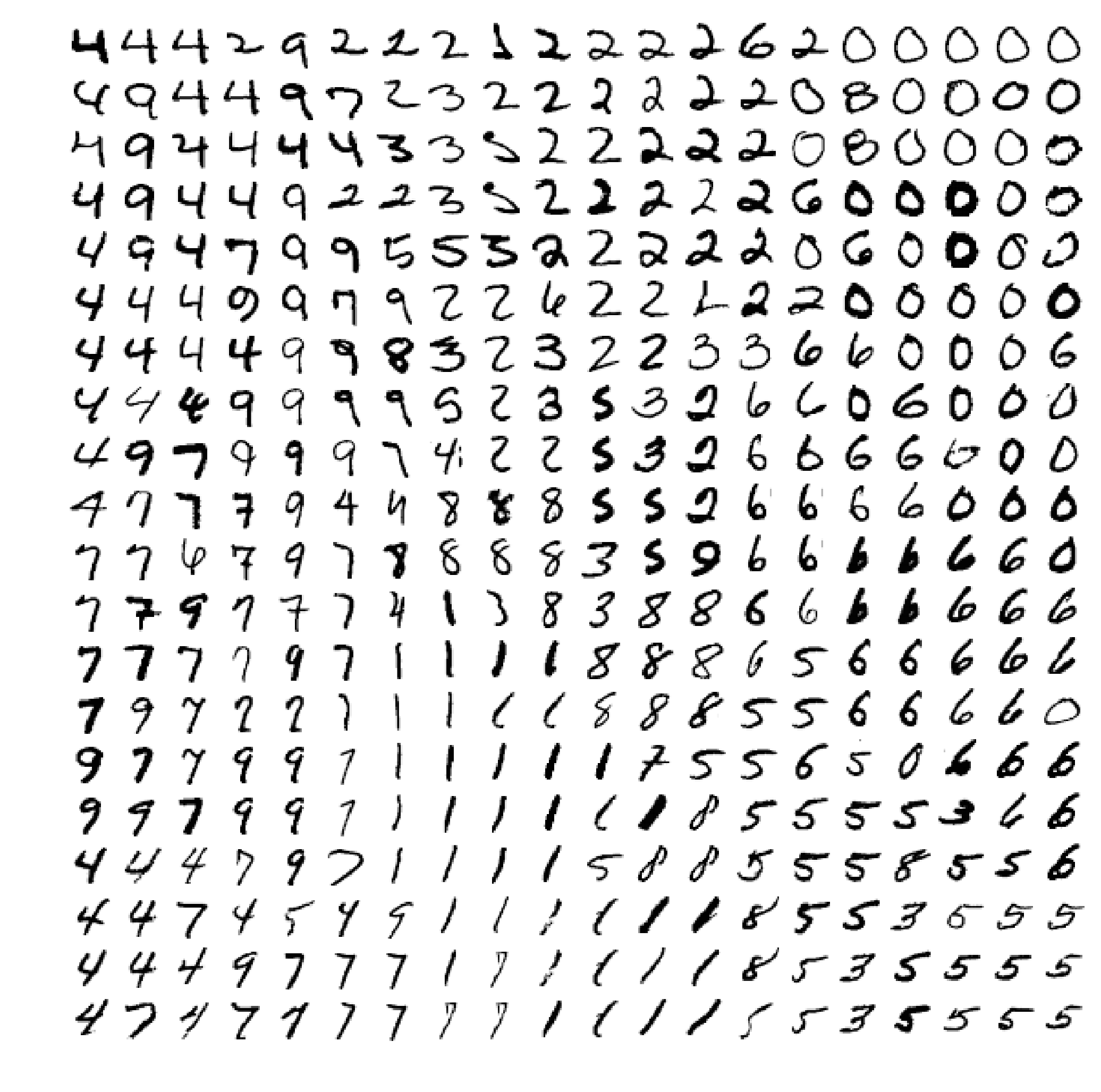}
 \end{subfigure}\hfil
 \begin{subfigure}{0.28\textwidth}
  \includegraphics[width=\linewidth]{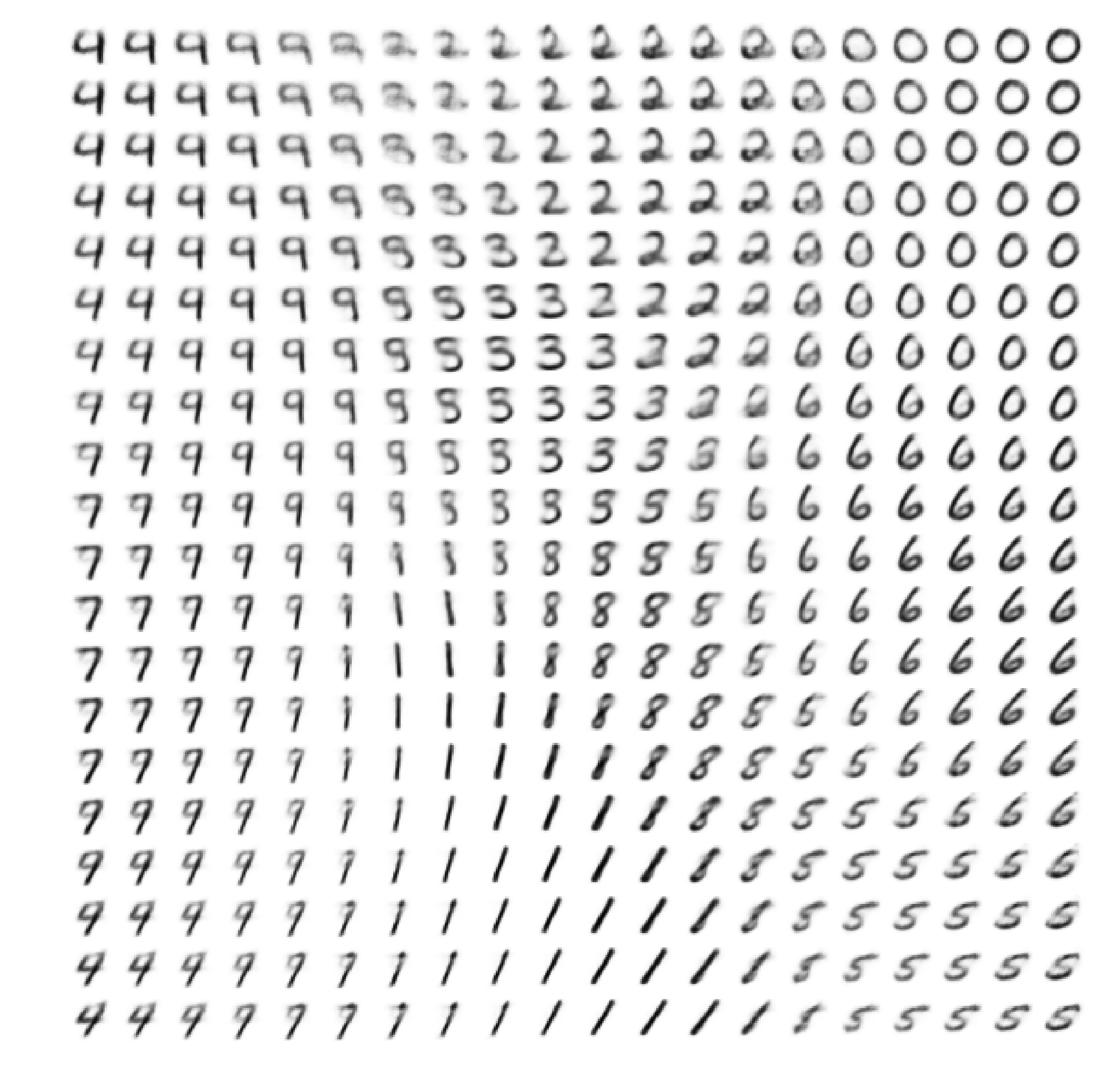}
  \end{subfigure}
  \caption{
  MNIST Posteriors mean (Left) sub-sample of digit images associated with posteriors mean (Middle) Model samples (Right) -- for \pvae{1.4} (Top) and \nvae{} (Bottom).
  }
  \label{fig:mnist}
\end{figure}

Second, we explore the learned latent representations of the trained \pvae{} and \nvae{} models shown in Figure \ref{fig:mnist}.
Qualitatively our \pvae{} produces a clearer partitioning of the digits, in groupings of $\{4, 7, 9\}$, $\{0, 6\}$, $\{2, 3, 5, 8\}$ and $\{1\}$, with right-slanting $\{5,8\}$ being placed separately from the non-slanting ones.
Recall that distances increase towards the edge of the Poincar\'e ball.
We quantitatively assess the quality of the embeddings by training a classifier predicting labels.
Table \ref{table:accuracy} shows that the embeddings learned by our \pvae{} model yield on average an $2\%$ increase in accuracy over the digits.
The full confusion matrices are shown in Figure \ref{fig:confusion} in Appendix.
\begin{table}[!h]
\vspace{-.5em}
\caption{
Per digit accuracy of a classifier trained on the learned latent $2$-d embeddings. 
Results are averaged over 10 sets of embeddings and 5 classifier trainings.
}
\label{table:accuracy}
\setlength\tabcolsep{2.pt}
\centering
\begin{tabular}{llllllllllll}
\toprule
{Digits}     & 0  & 1  & 2  & 3  & 4  & 5  & 6  & 7  & 8  & {9}  & Avg \\ \hline
{\nvae{}}     & 89     & 97 & 81      & 75     & 59       & 43      & 89 & $\bm{78}$ & 68 & {$\bm{57}$} & 73.6    \\ \hline
{\pvae{1.4}} & $\bm{94}$ & 97 & $\bm{82}$ & $\bm{79}$ & $\bm{69}$ & $\bm{47}$ & $\bm{90}$ & 77 & 68 & {53} & $\bm{75.6}$    \\
\bottomrule
\end{tabular}
\vspace{-.5em}
\end{table}
\subsection{Graph embeddings}
We evaluate the performance of a \gls{VGAE} \citep{Kipf:2016ul} with Poincar\'e ball latent space for link prediction in networks.
Edges in complex networks can typically be explained by a latent hierarchy over the nodes \citep{Clauset.Moore.ea2008Hierarchicalstructureand}.
We believe the Poincar\'e ball latent space should help in terms of generalisation.
We demonstrate these capabilities on three network datasets: 
a graph of Ph.D. advisor-advisee relationships \citep{Nooy:2011:ESN:2181139}, 
a phylogenetic tree expressing genetic heritage \citep{Hofbauer,TreeBASE}
and a biological set representing disease relationships \citep{Goh8685,Rossi:2015:NDR:2888116.2888372}.

We follow the \gls{VGAE} model, which maps the adjacency matrix $\bm{A}$ to node embeddings $\bm{Z}$ through a \gls{GCN}, and reconstructs $\bm{A}$ by predicting edge probabilities from the node embeddings.
In order to take into account the latent space geometry, we parametrise the probability of an edge by 
$p(\bm{A}_{ij} = 1|\z_i, \z_j) = 1 - \tanh(d_{\M}(\z_i, \z_j)) \in (0, 1]$ with $d_{\M}$ the latent geodsic metric.

We set the latent dimension to $5$.
We follow the training and evaluation procedures introduced in \cite{Kipf:2016ul}.
Models are trained on an incomplete adjacency matrix where some of the edges have randomly been removed.
A test set is formed from previously removed edges and an equal number of randomly sampled pairs of unconnected nodes.
We report in Table \ref{tab:network} the \emph{area under the ROC curve} (AUC) and \emph{average precision} (AP) evaluated on the test set.
It can be observed that the \pvae{} performs better than its Euclidean counterpart in terms of generalisation to unseen edges.

\begin{table}[h!]
  \vspace{-1em}
  \centering
  \caption{Results on network link prediction. 95\% confidence intervals are computed over 40 runs. 
    \label{tab:network}}
  \begin{tabular}{lccccccccc}
    \toprule
    & \multicolumn{2}{c}{\texttt{Phylogenetic}} & \multicolumn{2}{c}{\texttt{CS PhDs}} & \multicolumn{2}{c}{\texttt{Diseases}}\\
    \cmidrule(lr){2-3} \cmidrule(l){4-5} \cmidrule(l){6-7} 
     &  \textbf{AUC} & \textbf{AP} & \textbf{AUC} & \textbf{AP} & \textbf{AUC} & \textbf{AP} \\
    \midrule
\bf{\nvae{}}  & $54.2_{\pm2.2}$ & $54.0_{\pm2.1} $
       & $56.5_{\pm1.1}$ & $56.4_{\pm1.1}$ 
      &  $89.8 _{\pm0.7}$ & $91.8_{\pm0.7}$ \\

     \bf{\pvae{}} & $\bm{59.0_{\pm1.9}}$ & $ 55.5_{\pm1.6}$
       & $\bm{59.8}_{\pm1.2}$ & $ 56.7_{\pm1.2}$
       & $\bm{92.3}_{\pm0.7}$ & $\bm{93.6}_{\pm0.5}$  \\
    \bottomrule
  \end{tabular}
    \vspace{-1em}
\end{table}


\section{Conclusion}
\label{sec:future}
In this paper we have explored \gls{VAE}s with a Poincar\'e ball latent space. 
We gave a thorough treatment of two canonical -- \emph{wrapped} and \emph{maximum entropy} -- normal generalisations on that space, and a rigorous analysis of the difference between the two.
We derived the necessary ingredients for training such \gls{VAE}s, namely efficient and reparametrisable sampling schemes, along with probability density functions for these two distributions.
We introduced a decoder architecture explicitly taking into account the hyperbolic geometry, and empirically showed that it is crucial for the hyperbolic latent space to be useful.
We empirically demonstrated that endowing a \gls{VAE} with a Poincar\'e ball latent space can be beneficial in terms of model generalisation and can yield more interpretable representations if the data has hierarchical structure.

There are a number of interesting future directions.
There are many models of hyperbolic geometry, and several have been considered in a gradient-based setting.
Yet, it is still unclear which models should be preferred and which of their properties matter.
Also, it would be useful to consider principled ways of assessing whether a given dataset has an underlying hierarchical structure, 
in the same way that topological data analysis \citep{Pascucci:2011:TMD:1983580} attempts to discover the topologies that underlie datasets.

\newpage
\paragraph{Acknowledgments}
We are extremely grateful to Adam Foster, Phillipe Gagnon and Emmanuel Chevallier for their help.
EM, YWT’s research leading to these results received funding from the European Research Council under the European Union’s Seventh Framework Programme (FP7/2007- 2013) ERC grant agreement no. 617071 and they acknowledge Microsoft Research and EPSRC for funding EM’s studentship, and EPSRC grant agreement no.
EP/N509711/1 for funding CL’s studentship.

\bibliographystyle{apalike}
\bibliography{bib}

\newpage
\begin{appendices}


\section{Evidence Lower Bound}
\label{sec:elbo}
The \gls{ELBO} can readily be extended for Riemannian latent spaces by applying Jensen's inequality w.r.t. the metric induced measure $d\M$ which yield
\begin{align*}
\ln p(\x) &= \ln \int_{\mathcal{Z=\M}} p_\theta(\x, \z) d\M(\z)
 = \ln \int_{\M} p_\theta(\x|\z) p(\z) d\M(\z)  \nonumber\\
 &= \ln \int_{\M} \frac{p_\theta(\x|\z) p(\z)}{q_\phi(\z|\x)} q_\phi(\z|\x) d\M(\z)  \nonumber\\
 &\ge \int_{\M} \ln \cfrac{p_\theta(\x|\z) p(\z)}{q_\phi(\z|\x)} \ q_\phi(\z|\x) d\M(\z)  \nonumber\\
&= \int_{\M} [ \ln p_\theta(\x|\z) - \ln p(\z) - \ln q_\phi(\z|\x) ] \ q_\phi(\z|\x) \ d\M(\z)  \nonumber\\
 &= \mathbb{E}_{\z \sim q_\phi(\cdot|\x)\M(\cdot)} \left[ \ln p_\theta(\x|\z) + \ln p(\z) - \ln q_\phi(\z|\x)  \right]  \\
 &\triangleq \mathcal{L}_{\M}(\x; \theta, \phi) \nonumber \\
 &\approx \sum_k{\ln p_\theta(\x|\z^k) + \ln p(\z^k) - \ln q_\phi(\z^k|\x)}, \quad \z^k \sim q_\phi(\cdot|\x)\sqrt{|G(\cdot)|}
\end{align*}

\section{Hyperbolic normal distributions}
In this section, we first review some canonical generalisation of the normal distributions to Riemannian manifolds, and then introduce in more details the \emph{Riemannian} and \emph{wrapped} normal distributions on the Poincar\'e ball.
Finally, we give architecture and training details about the conducted experiments.

\subsection{Probability measures on Riemannian manifolds}
\label{sec:measures_riem}
Probability measures and random vectors can intrinsically be defined on Riemannian manifolds so as to model uncertainty on non-flat spaces \citep{Pennec2006}.
The Riemannian metric $G(\z)$ induces an infinitesimal volume element on each tangent space $\T_{\z}\M$, and thus a measure on the manifold,
\begin{align}
d\M(\z) = \sqrt{|G(\z)|}d\z,
\end{align}
with $d\z$ being the Lebesgue measure.
Random variables $\z \in \M$ would naturally be characterised by the Radon-Nikodym derivative of a measure $\nu$ w.r.t. the Riemannian measure $d\M(\cdot)$ (assuming absolute continuity)
\begin{align*}
f(\z) = \frac{d\nu(\z)}{d\M(\z)}.
\end{align*}
Since the normal distribution plays such a canonical role in statistics, generalising it to manifold is of interest.
Given a Fr\'echet expectation $\bm{\mu} \in \M$ -- 
defined as minimisers of $\int_{\M} d_{\M}(\bm{\mu}, \z)^2 p(\z) d\M(\z)$ -- and a dispersion parameter $\sigma > 0$ (generally not equal to the standard deviation), several properties ought to be verified by such generalised normal distributions.
Such a distribution should tend towards a delta function at $\bm{\mu}$ when $\sigma \rightarrow 0$ and to an (improper for non-compact) uniform distribution when $\sigma \rightarrow \infty$.
Also, as the curvature tends to $0$, one should recover the vanilla normal distribution.
Hereby, we review canonical generalisations of the normal distribution, which have different theoretical and computational advantages.

\paragraph{Maximum entropy normal}
The property that \cite{Pennec2006} takes for granted is the maximization of the entropy given a mean and a covariance matrix, yielding in the isotropic setting
\begin{align}
\frac{d\nu^{\text{R}}(\z|\bm{\mu}, \sigma^2)}{d\M(\z)} &= \mathcal{N}^{\text{R}}_{\M}(\z|\bm{\mu}, \sigma^2)
 = \frac{1}{Z^{\text{R}}} \exp\left(- \frac{d_{\M}(\bm{\mu}, \z)^2} {2 \sigma^2}\right),
\label{eq:maxent_normal}
\end{align}
with $d_{\M}$ being the Riemannian distance on the manifold induced by the tensor metric.
Such a formulation -- sometimes referred as \emph{Riemannian Normal} distribution --  is used by \cite{DBLP:journals/entropy/SaidBB14} in the Poincar\'e half-plane, or by \cite{Hauberg} in the hypersphere $\mathbb{S}^d$.
Sampling from such distributions and computing the normalising constant -- especially in the anisotropic setting -- is usually challenging.

\paragraph{Wrapped normal}
Another generalisation is defined by taking the image by the exponential map of a Gaussian distribution on the tangent space centered at the mean value.
Such a distribution has been referred in literature as \emph{wrapped}, \emph{push-forward}, \emph{exp-map} or \emph{tangential} normal distribution.
Sampling is therefore straightforward.
The pdf is then readily available through the change of variable formula if one can compute the Jacobian of the exponential map (or its inverse).
Hence such a distribution is attractive from a computational perspective.
\cite{Grattarola:2018ue} and \cite{Nagano} rely on such a distribution defined on the hyperboloid model.
Wrapped distributions are often encountered in the \emph{directional} statistics \citep{directional_stats,Hauberg}.

\paragraph{Restricted normal}
What is more, for sub-manifolds of $\R^n$, one can consider the restriction of a normal distribution pdf to the manifold.
This yields the Von Mises distribution on $\mathbb{S}^1$ and the Von Mises-Fisher distribution on $\mathbb{S}^d$ \citep{Hauberg} and the Stiefel manifold.
It is the maximum entropy distribution but with respect to the ambient euclidean metric \citep{doi:10.1111/j.2517-6161.1975.tb01550.x}.

\paragraph{Diffusion normal}
Yet another generalisation arises by defining the normal pdf through the \emph{heat kernel}, or fundamental solution of the heat equation, $K: \R^+ \times \M \times \M \rightarrow \M$,
\begin{align}
\mathcal{N}^{\Delta}_{\M}(\z|\bm{\mu}, \sigma^2) = K(\sigma^2 / 2, \bm{\mu}, \z).
\end{align}
See for instance \cite{Hsu2008} for an introduction of Brownian motion on Riemannian manifolds and \cite{PAENG2011940} for conditions on existence and uniqueness of the kernel.
Sampling amounts to simulating a Brownian motion, which may be challenging for non sub-manifolds of $\R^n$.
Closed form solutions of the heat kernel is available for some manifolds such as spheres or flat tori, otherwise numerical approximations can be used.
Such a distribution has been used in a \gls{VAE} setting \citep{Rey:2019ty,Li:2019wa}.

\paragraph{Other than normal distributions}
Of course one needs not to restrict itself to generalisations of the normal distribution.
For instance, one could consider a wrapped spherical Student-t as 
$\z \sim \exp_{\bm{\mu}\#} S_t(0, \nu)$
or a Riemannian Student-t with density proportional to $\left(1 + d_{\M}(\z, \bm{\mu})^2 / \nu \right)^{(-\nu + 1) / 2}$ (by making sure that this density is $d\M$-integrable).

\subsection{Hyperbolic polar coordinates}
\label{sec:changeofvar}
In this subsection, we review the hyperbolic polar change of coordinates allowing us to reparametrise hyperbolic normal distributions in a similar fashion than the Box–Muller transform \citep{box1958}.

\paragraph{Polar coordinates}
Euclidean polar coordinates, express points $\z \in \R^d$ through a radius $r\ge0$ and a direction $\bm{\alpha} \in \mathbb{S}^{d-1}$ such that $\z = r \bm{\alpha}$. Yet, one could choose another \emph{pole} (or \emph{reference point}) $\bm{\mu}\neq\bm{0}$ such that $\z= \bm{\mu} + r \bm{\alpha}$. Consequently, $r = d_E(\bm{\mu},\z)$.
An analogous change of variables can also be constructed in Riemannian manifolds relying on the exponential map instead of the addition operator.
Given a \emph{pole} $\bm{\mu}\in \B_c^d$, the point of hyperbolic polar coordinates $\z=(r, \bm{\alpha})$ is defined as $\z = \gamma(r)$, with $r=d^c_p(\bm{\mu}, \z)$ and $\gamma: \R^+ \rightarrow \B_c^d$ a curve such that $\gamma'(0) = \bm{\alpha} \in \mathbb{S}^{d-1}$.
Hence $\z = \exp^c_{\bm{\mu}} \left(\frac{r}{\lambda^c_{\bm{\mu}}} \bm{\alpha} \right)$ 
since $d^c_p(\bm{\mu}, \z)=\|\ln^c_{\bm{\mu}}(x)\|_{\bm{\mu}}=\|\frac{r}{\lambda^c_{\bm{\mu}}} \bm{\alpha}\|_{\bm{\mu}}=r$.

\paragraph{Tensor metric}
We derive below the expression of the Poincar\'e ball metric in such hyperbolic polar coordinate, for the specific setting where $\bm{\mu}=\bm{0}$: $\z = \exp^c_{\bm{0}}( \frac{r}{2} \bm{\alpha})$.
Switching to Euclidean polar coordinate we get
\begin{align}
ds^2_{\B_c^d} &= (\lambda^c_{\z})^2 (dz_1^2 + \dots + dz_d^2)
 = \frac{4}{\left(1 - c \|x\|^2\right)^2}  d\z^2 \nonumber\\
& = \frac{4}{(1 - c \rho^2)^2} (d\rho^2 + \rho^2 ds^2_{\mathbb{S}^{d-1}}).
\label{eq:polarcoord_proof}
\end{align}
Let's define $r=d^c_p(\bm{0}, \z)=L(\gamma)$, with $\gamma$ being the geodesic joining $\bm{0}$ and $\z$. Since such a geodesic is the segment $[\bm{0}, \z]$, we have
\begin{align*}
r = \int_{0}^{\rho} \lambda_t^c dt = \int_{0}^{\rho} \frac{2}{1 - c t^2} dt 
= 
\int_{0}^{\sqrt{c}\rho} \frac{2}{1 - t^2} \frac{dt}{\sqrt{c}}
= 
\frac{2}{\sqrt{c}} \tanh^{-1}(\sqrt{c}\rho).
\end{align*}
Plugging $\rho = \frac{1}{\sqrt{c}}\tanh(\sqrt{c}\frac{r}{2})$ (and $d\rho = (1 - c\rho^2)/2 dr$) into Eq \ref{eq:polarcoord_proof} yields
\begin{align}
ds^2_{\B_c^d} &= \frac{4}{(1 - c \rho^2)^2} \frac{1}{4} (1 - c\rho^2)^2 dr^2
+ \left(2\frac{\rho}{1 - c \rho^2}\right)^2 ds^2_{\mathbb{S}^{d-1}} \nonumber \\
&= dr^2 + \left(2\frac{\frac{1}{\sqrt{c}}\tanh(\sqrt{c}\frac{r}{2})}{1 - c \left(\frac{1}{\sqrt{c}}\tanh(\sqrt{c}\frac{r}{2}\right)^2}\right)^2 ds^2_{\mathbb{S}^{d-1}} \nonumber \\
&= dr^2 + \left(\frac{1}{\sqrt{c}}\sinh(\sqrt{c}r)\right)^2 ds^2_{\mathbb{S}^{d-1}}.
\label{eq:polarcoord_proof2}
\end{align}
The Euclidean line element is recovered when $c \rightarrow 0$
\begin{align}
ds^2_{\R^d}= dr^2 + r^2 ds^2_{\mathbb{S}^{d-1}}.
\label{eq:polarcoord_proof2}
\end{align}
In an appropriate orthonormal basis of $\T_{\bm{\mu}}\B_c^d$, the hyperbolic polar coordinate leads to the following expression of the matrix of the metric
\begin{align}
G(\z) = \begin{pmatrix} 1 & 0 \\ 0 & \left(\frac{\sinh(\sqrt{c}r)}{\sqrt{c}r}\right)^2 \bm{I}_{d-1} \end{pmatrix}.
\end{align}
Hence, the density of the Riemannian measure with respect to the image of the Lebesgue measure of $\T_{\bm{\mu}}\B_c^d$ by $\exp^c_{\bm{\mu}}$ is given by
\begin{align}
\sqrt{|G(\z)|} = \left(\frac{\sinh(\sqrt{c}r)}{\sqrt{c}r}\right)^{d-1}.
\end{align}
This result holds for any \emph{reference point} $\bm{\mu} \in \B_c^d$, with $r=d^c_p(\bm{\mu}, \z)$, since the metric induced measure is invariant under the isometries of the manifold (i.e. M\"{o}bius transformations).
This result can also be found in \cite{10.1007/978-3-319-25040-3_80,DBLP:journals/entropy/SaidBB14}.
Also, the fact that the line element $ds^2_{\B_c^d}$ and equivalently the metric $G$ only depends on the radius in hyperbolic polar coordinate, is a consequence of the hyperbolic space's isotropy.
\paragraph{Integration}
\label{sec:pr}
We now make use of the aforementioned hyperbolic polar coordinates to integrate functions following \cite{DBLP:journals/entropy/SaidBB14}.
The integral of a function $f: \B_c^d \rightarrow \R$ can be computed by using polar coordinates,
\begin{align}
\int_{\B_c^d} f(\z) d\M(\z) 
&= \int_{\B_c^d} f(\z) \sqrt{|G(\z)|} \ d\z \nonumber \\
& = \int_{\T_{\bm{\mu}}\B_c^d \cong \R^d} f(\v) \sqrt{|G(\v)|} \ d\v \label{eq:polar_coord_integral_bis} \\
& = \int_{\R_+} \int_{\mathbb{S}^{d-1}} f(r) \sqrt{|G(r)|} dr ~r^{d-1} ~ds_{\mathbb{S}^{d-1}} \nonumber \\
& = \int_{\R_+} \int_{\mathbb{S}^{d-1}} f(r) \left(\frac{\sinh(\sqrt{c}r)}{\sqrt{c}r}\right)^{d-1} dr ~r^{d-1} ~ds_{\mathbb{S}^{d-1}} \nonumber \\
& = \int_{\R_+} \int_{\mathbb{S}^{d-1}} f(r) \left(\frac{\sinh(\sqrt{c}r)}{\sqrt{c}} \right)^{d-1} dr ~ds_{\mathbb{S}^{d-1}}. \label{eq:polar_coord_integral}
\end{align}
\subsection{Wrapped hyperbolic normal distribution on $\B_c^d$}
\label{sec:wrapped_normal}
\paragraph{Anisotropic}
The \emph{wrapped normal} distribution considers a normal distribution in the tangent space $\T_{\bm{\mu}}\B_c^d$ being pushed-forward along the exponential map.
One can obtain sampled as follow
\begin{align}
\z= \exp^c_{\bm{\mu}} \left( G(\bm{\mu})^{-\frac{1}{2}} ~\v \right)=\exp^c_{\bm{\mu}} \left( \frac{\v}{\lambda^c_{\bm{\mu}}} \right)
 ,\ \text{with} \ \bm{\v} \sim \mathcal{N}(\cdot|\bm{0}, \Sigma).
\end{align}

Then, its density is given by
\begin{align} 
\mathcal{N}^{\text{W}}_{\B_c^d}(\z|\bm{\mu}, \Sigma)
&= \mathcal{N} \left(G(\bm{\mu})^{1/2}\log_{\bm{\mu}}(\z) ~\middle|~\bm{0}, \Sigma \right)
\bigg( \frac{\sqrt{c} ~d_p^c(\bm{\mu}, \z)}{\sinh(\sqrt{c} ~d_p^c(\bm{\mu}, \z))}  \bigg)^{d-1} \nonumber  \\
&= \mathcal{N} \left(\lambda^c_{\bm{\mu}}~ \log_{\bm{\mu}}(\z) ~\middle|~\bm{0}, \Sigma \right)
\bigg( \frac{\sqrt{c} ~d_p^c(\bm{\mu}, \z)}{\sinh(\sqrt{c} ~d_p^c(\bm{\mu}, \z))}  \bigg)^{d-1} \label{eq:wrapped_normal_hyp}
\end{align}
with $G(\bm{\mu})^{1/2}$ the unique square-root matrix of $G(\bm{\mu})$ (thanks to the positive definiteness of the metric tensor).
This can be shown by plugging this density as $f$ in Equation (\ref{eq:polar_coord_integral_bis}) with 
$\v= r \bm{\alpha}=\lambda^c_{\bm{\mu}} \log_{\bm{\mu}}(\z)$, we get 
\begin{align}
\int_{\B_c^d} \mathcal{N}^{\text{W}}_{\B_c^d}(\z|\bm{\mu}, \Sigma) ~d\M(\z) 
&=\int_{\T_{\bm{\mu}}\B_c^d \cong \R^d} \mathcal{N} \left(\v ~\middle|~\bm{0}, \Sigma \right)
\bigg( \frac{\sqrt{c} ~\left\|\v\right\|_{2}}{\sinh(\sqrt{c} ~\left\|\v\right\|_{2})}  \bigg)^{d-1} \sqrt{|G(\v)|} ~d\v \nonumber \\
&=\int_{\R^d} \mathcal{N} \left(\v ~\middle|~\bm{0}, \Sigma \right)
\bigg( \frac{\sqrt{c} ~\left\|\v\right\|_{2}}{\sinh(\sqrt{c} ~\left\|\v\right\|_{2})}  \bigg)^{d-1} \left(\frac{\sinh(\sqrt{c}\left\|\v\right\|_{2})}{\sqrt{c}\left\|\v\right\|_{2}}\right)^{d-1} ~d\v \nonumber \\
&=\int_{\R^d} \mathcal{N} \left(\v ~\middle|~\bm{0}, \Sigma \right) ~d\v. \nonumber
\end{align}

\begin{figure}[h!]
    \vspace{-0em}
    \centering
    \begin{subfigure}{0.32\textwidth}
    \includegraphics[width=\linewidth]{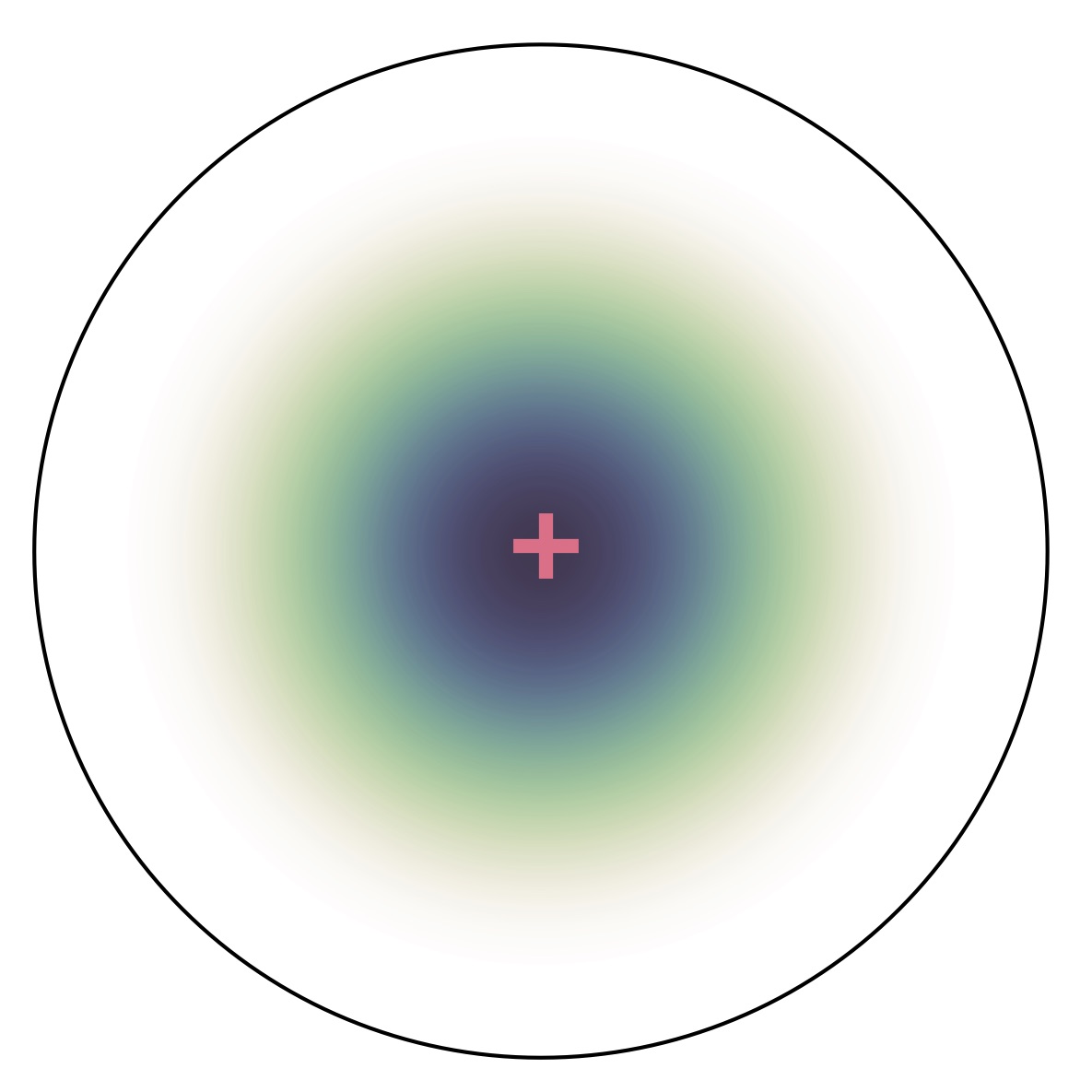}
    \put(-140,40){\rotatebox{90}{$\|\bm{\mu}\|_2=0.0$}}
    \end{subfigure}
    \begin{subfigure}{0.32\textwidth}
    \includegraphics[width=\linewidth]{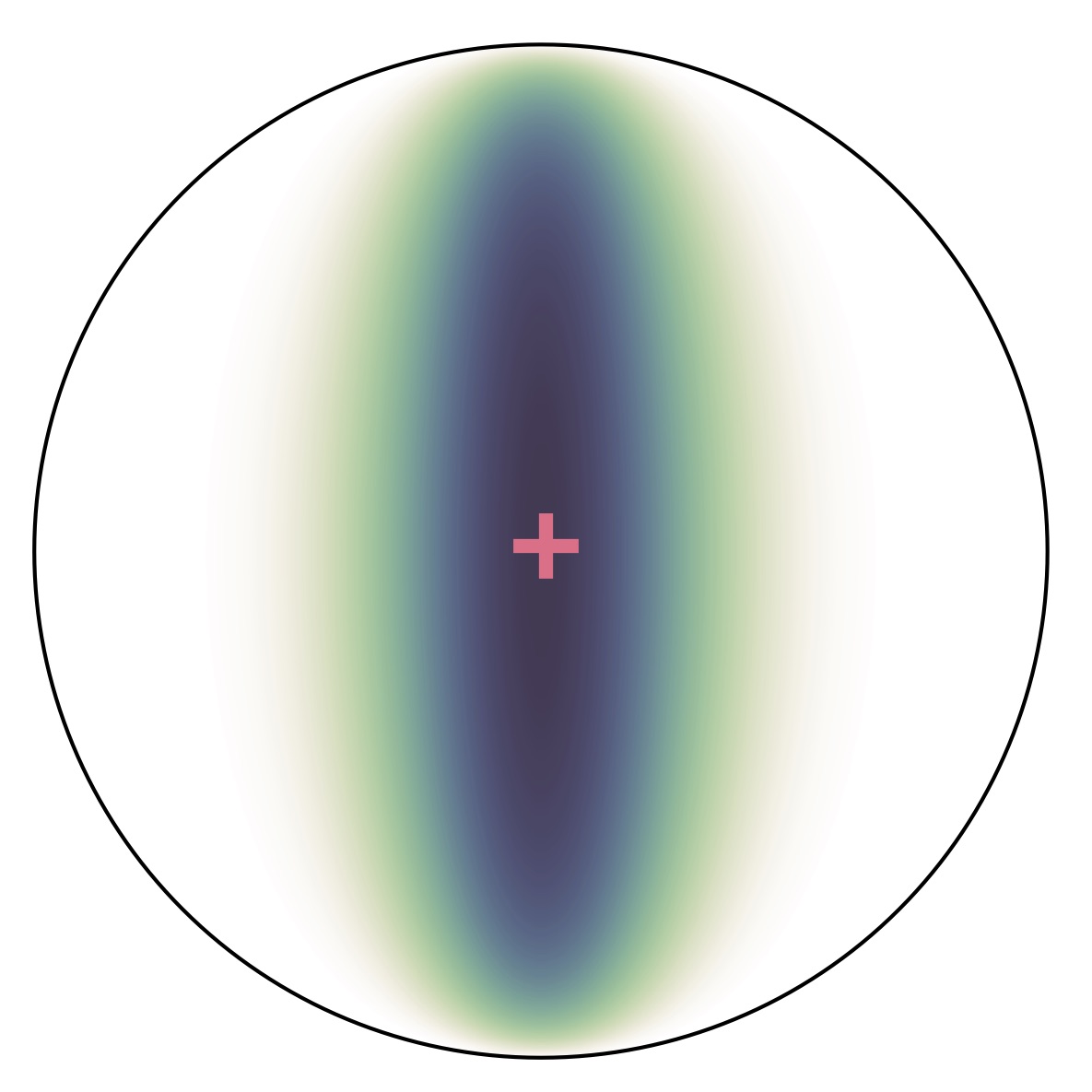}
    \end{subfigure}
    \begin{subfigure}{0.32\textwidth}
    \includegraphics[width=\linewidth]{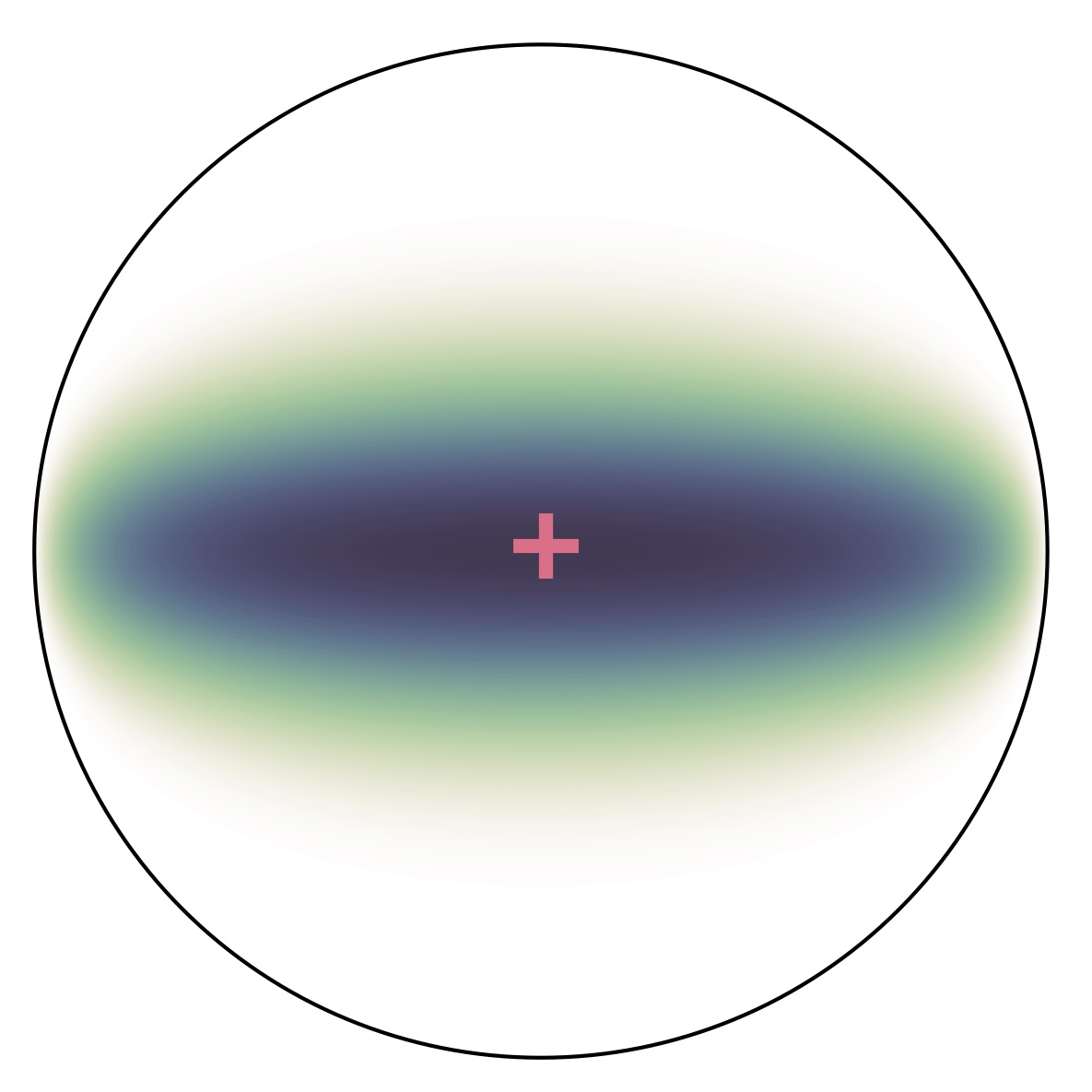}
    \end{subfigure}
    \begin{subfigure}{0.32\textwidth}
    \includegraphics[width=\linewidth]{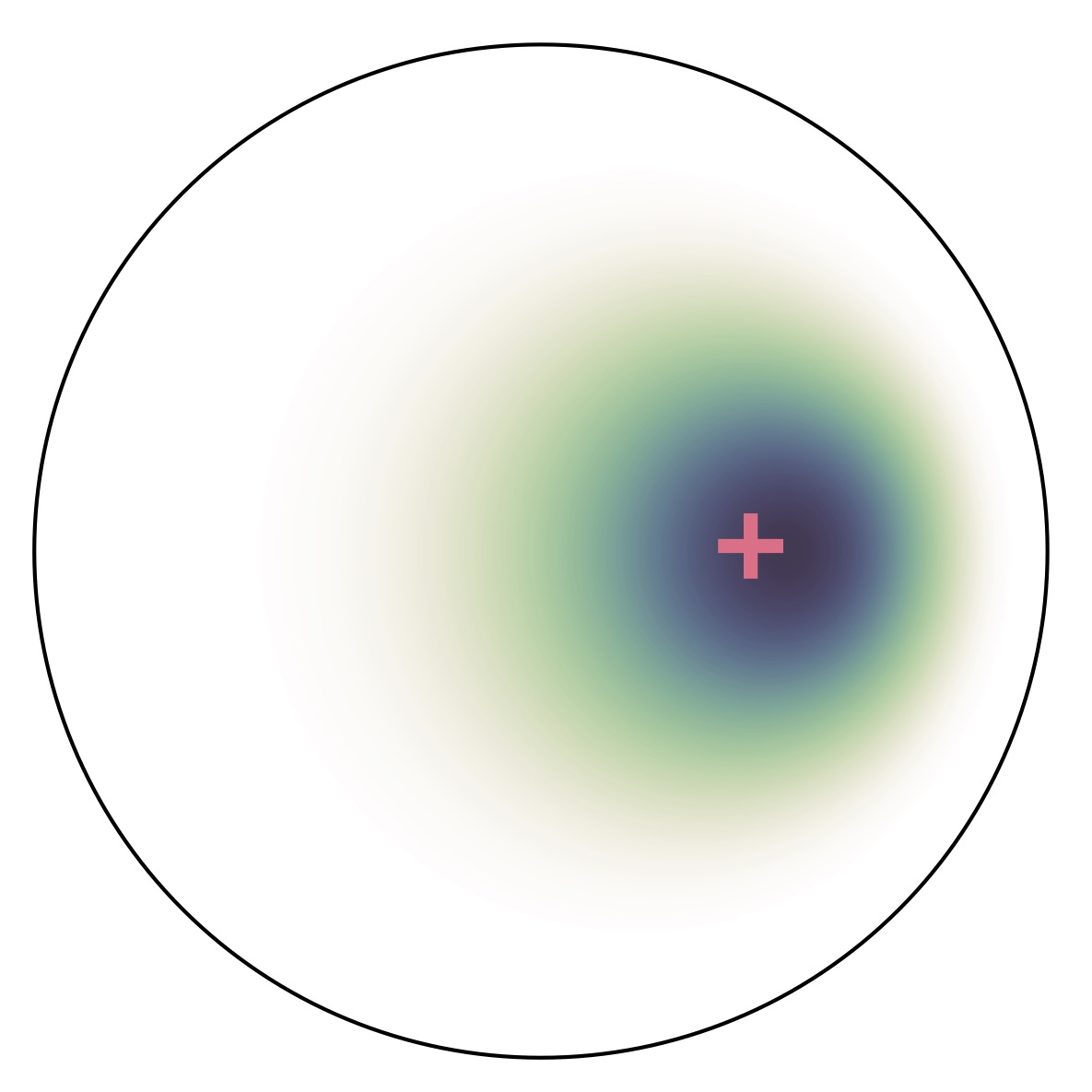}
    \put(-140,40){\rotatebox{90}{$\|\bm{\mu}\|_2=0.4$}}
    \put(-95,-10){$\bm{\sigma}=(1.0, 1.0)$}
    \end{subfigure}
    \begin{subfigure}{0.32\textwidth}
    \includegraphics[width=\linewidth]{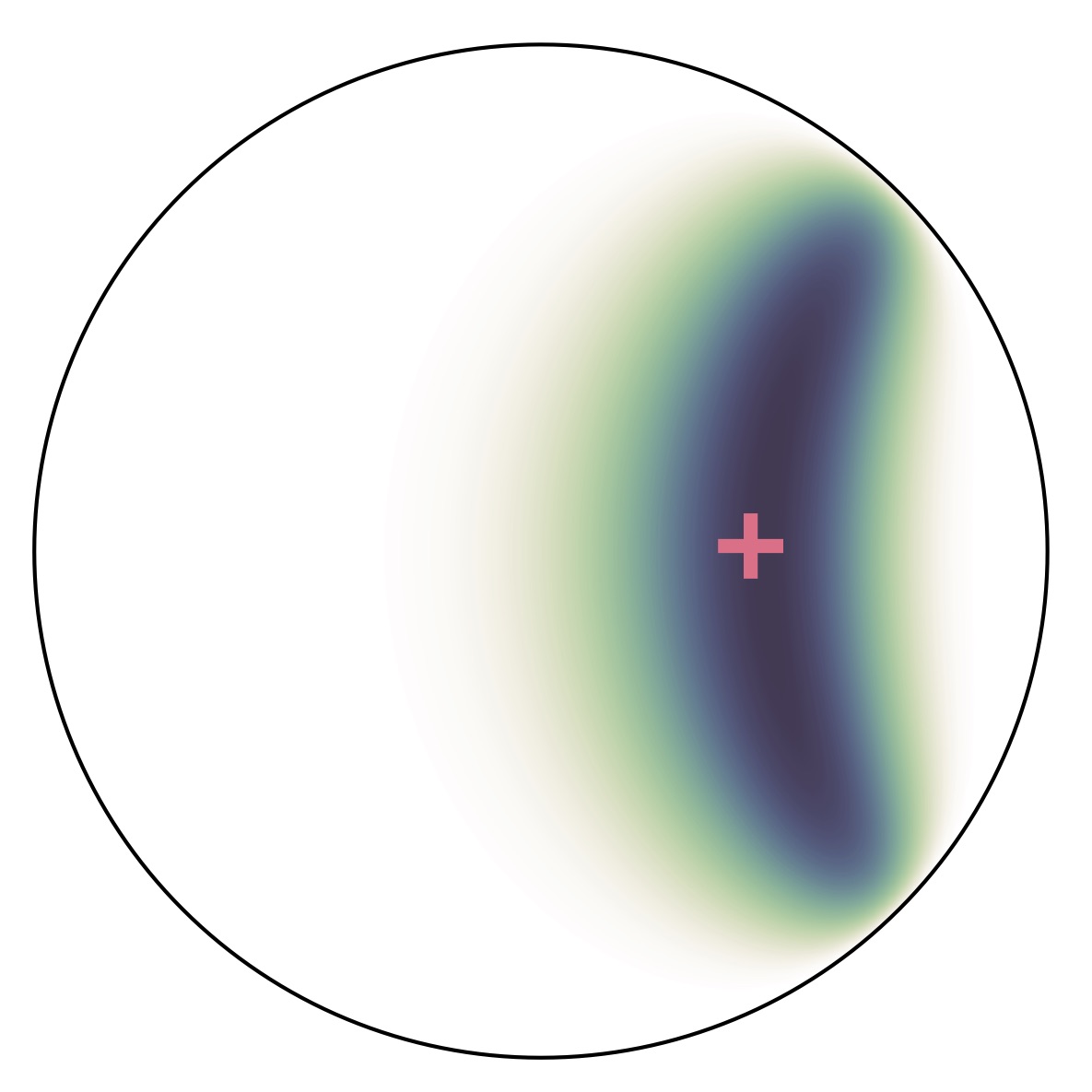}
    \put(-95,-10){$\bm{\sigma}=(2.0, 0.5)$}
    \end{subfigure}
    \begin{subfigure}{0.32\textwidth}
    \includegraphics[width=\linewidth]{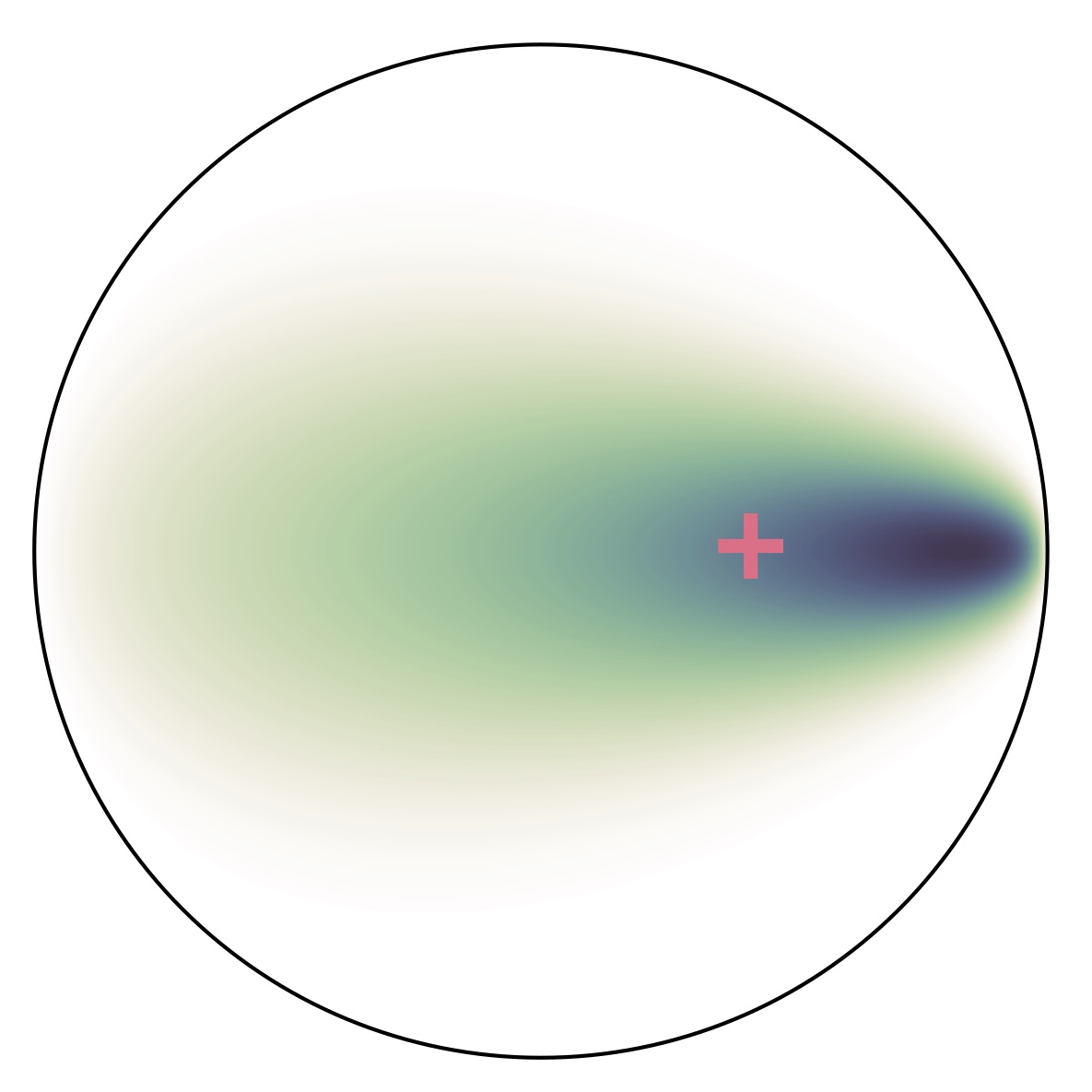}
    \put(-95,-10){$\bm{\sigma}=(0.5, 2.0)$}
    \end{subfigure}
    
      \caption{
      Anisotropic wrapped normal probability measures for Fr\'echet means $\bm{\mu}$ (red +), concentrations $\Sigma=\text{diag}(\bm{\sigma})$ and $c=1$.
      }
      \label{fig:hypp_norm_samples}
    \end{figure}

\paragraph{Isotropic}
In the isotropic setting, we therefore get
\begin{align}
\int_{\B_c^d} \mathcal{N}^{\text{W}}_{\B_c^d}(\z|\bm{\mu}, \sigma^2) ~d\M(\z) 
&= \int_{\R_+} \int_{\mathbb{S}^{d-1}} \frac{1}{Z^{\text{R}}} e^{- \frac{r^2}{2\sigma^2}} r^{d-1} dr ~ds_{\mathbb{S}^{d-1}}.
\end{align}

The hyperbolic radius $r = d_p^c(\bm{\mu}, \z)$ consequently follows the usual $\chi$ distribution with density
\begin{align}
\rho^{\text{W}}(r) \propto \1_{\R_+}(r) ~e^{- \frac{r^2}{2\sigma^2}} r^{d-1},
\end{align}
and the density of the \emph{wrapped} normal given by
\begin{align*}
\mathcal{N}^{\text{W}}_{\B_c^d}(\z|\bm{\mu}, \sigma^2)
 = \frac{d\nu^{\text{W}}(\z|\bm{\mu}, \sigma^2)}{d\M(\z)}
&= (2 \pi \sigma^2)^{-d/2} \exp\left(- \frac{d_p^c(\bm{\mu}, \z)^2} {2 \sigma^2}\right)
\bigg( \frac{\sqrt{c} ~d_p^c(\bm{\mu}, \z)}{\sinh(\sqrt{c} ~d_p^c(\bm{\mu}, \z))}  \bigg)^{d-1}.
\end{align*}

\subsection{Maximum entropy hyperbolic normal distribution on $\B_c^d$}
\label{sec:maxent_normal}
Alternatively, by considering the maximum entropy generalisation of the normal distribution one gets \citep{Pennec2006}
\begin{align}
\mathcal{N}^{\text{R}}_{\B_c^d}(\z|\bm{\mu}, \sigma^2)
 = \frac{d\nu^{\text{R}}(\z|\bm{\mu}, \sigma^2)}{d\M(\z)} 
= \frac{1}{Z^{\text{R}}} \exp\left(- \frac{d_p^c(\bm{\mu}, \z)^2} {2 \sigma^2}\right).
\label{eq:maxent_normal_hyp}
\end{align}
Such a pdf can be computed pointwise once $Z^{\text{R}}$ is known, which we derive in Appendix \ref{sec:normconst}.
Also, we observe that as $c$ and $\sigma$ get smaller (resp. bigger), the \emph{Riemannian} normal pdf gets closer (resp. further) to the \emph{wrapped} normal pdf.

\subsubsection{Reparametrisation} \label{sec:riemannian_reparametrisation}
Plugging the \emph{Riemannian} normal density as $f$ in Equation (\ref{eq:polar_coord_integral}), with $r=d^c_p(\bm{\mu}, \z)$, we have
\begin{align}
\int_{\B_c^d} \mathcal{N}^{\text{R}}_{\B_c^d}(\z|\bm{\mu}, \sigma^2) ~d\M(\z)  &= \int_{\R_+} \int_{\mathbb{S}^{d-1}} \frac{1}{Z^{\text{R}}} e^{- \frac{r^2}{2\sigma^2}} \left(\frac{\sinh(\sqrt{c}r)}{\sqrt{c}} \right)^{d-1} dr ~ds_{\mathbb{S}^{d-1}} \nonumber \\
& = \frac{1}{Z^{\text{R}}} \left( \int_{\R_+} e^{- \frac{r^2}{2\sigma^2}} \left(\frac{\sinh(\sqrt{c}r)}{\sqrt{c}} \right)^{d-1} dr \right)  \left( \int_{\mathbb{S}^{d-1}} ds_{\mathbb{S}^{d-1}} \right)  \label{eq:volume}
\end{align}
Hence, samples $\z \sim \mathcal{N}^{\text{R}}_{\M}(\z|\bm{\mu}, \sigma^2) d\M(\z)$ can be reparametrised as
\begin{align} \label{eq:reparameterisaition}
\z = \exp^c_{\bm{\mu}} \left(\frac{r}{\lambda^c_{\bm{\mu}}} \bm{\alpha} \right)
\end{align}
with the direction $\bm{\alpha}$ being uniformly distributed on the hypersphere $\mathbb{S}^{d-1}$, i.e.
\begin{align*}
\bm{\alpha} \sim \mathcal{U}(\mathbb{S}^{d-1})
\end{align*}
and the hyperbolic radius $r=d^c_p(\bm{\mu}, \z)$ distributed according to the following density (w.r.t the Lebesgue measure)
\begin{align}
\rho^{\text{R}}(r) = \frac{\1_{\R_+}(r)}{Z^{\text{R}}_r} e^{- \frac{r^2}{2\sigma^2}} \left(\frac{\sinh(\sqrt{c}r)}{\sqrt{c}} \right)^{d-1}.
\end{align}
\paragraph{Developed expression}
By expanding the $\sinh$ term using the binomial formula, we get
\begin{align}
\rho^{\text{R}}(r) &= \frac{\1_{\R_+}(r)}{Z^{\text{R}}_r} e^{- \frac{r^2}{2 \sigma^2}} \left(\frac{\sinh(\sqrt{c}r)}{\sqrt{c}} \right)^{d-1} \nonumber \\
&= \frac{\1_{\R_+}(r)}{Z^{\text{R}}_r} e^{- \frac{r^2}{2 \sigma^2}} \left( \frac{e^{\sqrt{c}r}-e^{-\sqrt{c}r}}{2\sqrt{c}} \right)^{d-1} \nonumber \\
&= \frac{\1_{\R_+}(r)}{Z^{\text{R}}_r} e^{- \frac{r^2}{2 \sigma^2}} \frac{1}{(2\sqrt{c})^{d-1}} 
 \sum_{k=0}^{d-1} {d-1 \choose k} \left(e^{\sqrt{c}r}\right)^{d-1-k} \left(-e^{-\sqrt{c}r}\right)^{k} \nonumber \\
&= \frac{\1_{\R_+}(r)}{Z^{\text{R}}_r}  \frac{1}{(2\sqrt{c})^{d-1}}  e^{- \frac{r^2}{2 \sigma^2}}
 \sum_{k=0}^{d-1} \left(-1 \right)^{k} {d-1 \choose k} e^{(d-1-2k)\sqrt{c}r}  \nonumber \\
&= \frac{\1_{\R_+}(r)}{Z^{\text{R}}_r} \frac{1}{(2\sqrt{c})^{d-1}} 
 \sum_{k=0}^{d-1} \left(-1 \right)^{k} {d-1 \choose k} e^{- \frac{r^2}{2 \sigma^2} + (d-1-2k)\sqrt{c}r} \nonumber \\
&= \frac{\1_{\R_+}(r)}{Z^{\text{R}}_r} \frac{1}{(2\sqrt{c})^{d-1}} 
 \sum_{k=0}^{d-1} \left(-1 \right)^{k} {d-1 \choose k}
 e^{\frac{(d-1-2k)^2}{2}c\sigma^2 } e^{- \frac{1}{2 \sigma^2} \left[r - (d-1-2k)\sqrt{c} \sigma^2 \right]^2 }  \label{eq:developedpr}.
\end{align}

\subsubsection{Sampling}
\label{sec:sampling}
In this section we detail the sampling scheme that we use for the Riemannian normal distribution $\mathcal{N}^{\text{R}}_{\B_c^d}(\cdot|\bm{\mu}, \sigma^2)$, along with a reparametrisation which allows to compute gradients with respect to the parameters $\bm{\mu}$ and $\sigma$.
\paragraph{Sampling challenges due to the hyperbolic geometry}
Several properties of the Euclidean space do not generalise to the hyperbolic setting, unfortunately hardening the task of obtaining samples from \emph{Riemannian} normal distributions.
First, one can factorise a normal density through the space's dimensions -- thanks to to the Pythagorean theorem -- hence allowing to divide the task on several subspaces and then concatenate the samples. Such a property does not extend to the hyperbolic geometry, thus seemingly preventing us from focusing on $2$-dimensional samples.
Second, in Euclidean geometry, the polar radius $r$ is distributed according to $\rho^{\text{W}}(r) = \frac{\1_{\R_+}(r)}{Z_r} e^{- \frac{r^2}{2\sigma^2}} r^{d-1}$, making it easy by a linear change of variable to take into account different scaling values. The non-linearity of $\sinh$  prevent us from using such a simple change of variable.
\paragraph{Computing gradients with respect to parameters}
So as to compute gradients of samples $\z$ with respect to the parameters $\bm{\mu}$ and $\sigma$ of samples of a hyperbolic distributions, we respectively rely on the reparametrisation given by Eq \ref{eq:reparameterisaition} for $\nabla_{\bm{\mu}} \z$, and on an implicit reparametrisation \citep{Figurnov:2018vr} of $r$ for $\nabla_{\sigma} \z$.
We have 
$\z = \exp^c_{\bm{\mu}} \left( \frac{r}{\lambda^c_{\bm{\mu}}} \bm{\alpha} \right)$ with 
$\bm{\alpha} \sim \mathcal{U}(\mathbb{S}^{d-1})$
and $r \sim \rho^{\text{R}}(\cdot)$.
Hence, 
\begin{align}
\nabla_{\bm{\mu}} \z= \nabla_{\bm{\mu}}\exp^c_{\bm{\mu}}(\u),
\end{align}
with $\u=\frac{r}{\lambda^c_{\bm{\mu}}} \bm{\alpha}$ (actually) independent of $\bm{\mu}$, and
\begin{align}
\nabla_{\sigma} \z = \nabla_{\sigma}\exp^c_{\bm{\mu}}(\u) = 
\nabla_{\u}\exp^c_{\bm{\mu}}(\u) \frac{\bm{\alpha}}{\lambda^c_{\bm{\mu}}} \nabla_{\sigma} r,
\end{align}
with $\nabla_{\sigma}(r)$ computed via the implicit reparametrisation given by
\begin{align}
\nabla_{\sigma}(r) &= - \left( \nabla_r F^{\text{R}}(r, \sigma) \right)^{-1} \nabla_{\sigma} F^{\text{R}}(r, \sigma) \nonumber \\
&= - \left( \rho^{\text{R}}(r;\sigma) \right)^{-1} \nabla_{\sigma} F^{\text{R}}(r, \sigma).
\end{align}
\paragraph{Sampling hyperbolic radii}
Unfortunately the density of the hyperbolic radius $\rho^{\text{R}}(r)$ is not a well-known distribution and its cumulative density function does not seem analytically invertible.
We therefore rely on rejection sampling methods.
\paragraph{Adaptive Rejection Sampling}
By making use of the log-concavity of $\rho^{\text{R}}$, we can rely on a piecewise exponential distribution proposal from \gls{ARS} \citep{ars}.
Such a proposal automatically adapt itself with respect to the parameters $\sigma$, $c$ and $d$.
Even though $\mathcal{N}_{\B_c^d}$ is defined on a d-dimensional manifold, $\rho^{\text{R}}$ is a univariate distribution hence the sampling scheme is not directly affected by dimensionality.
The difficulty in \gls{ARS} is to choose the initial set of points to construct the piecewise exponential proposal.
To do so, we first compute the mean $m = \mathbb{E}_{r \sim \rho^{\text{R}}}[r]$ and standard deviation $s = \mathbb{V}_{r \sim \rho^{\text{R}}}[r]^{1/2}$ of the targeted distribution.
Then we choose a grid $\eta = (\eta_1, \dots, \eta_K) = \left(\text{linspace}(\eta_{max}, \eta_{min}, K/2),\ \text{linspace}(\eta_{min}, \eta_{max}, K/2) \right)$.
Eventually, we set the initial points $(x_1, \dots, x_K)$ to $x_k = m + \eta_k * \min(s, 0.95 * m / \eta_{max})$.
For our experiments we chose $\eta_{min}=.1, \eta_{max}=3, K=20$.
We do not adapt the proposal within the rejection sampling since we empirically found it unnecessary. \\ \\
Alternatively, we derived bellow two non-adaptive proposal distributions along with their rejection rate constants.
Yet, we observe that these rates do not scale well the dimensionality $d$ and distortion $\sigma$, making them ill-suited for practical purposes.
\paragraph{Rejection Sampling with truncated Normal proposal}
The developed expression of $\rho(r)^{\text{R}}$ from Eq (\ref{eq:developedpr}) highlights the fact that the density can immediately be upper bounded by a truncated normal density:
\begin{align*}
\rho(r)^{\text{R}} &= \frac{\1_{\R_+}(r)}{Z^{\text{R}}_r} \frac{1}{(2\sqrt{c})^{d-1}}  \sum_{k=0}^{d-1} \left(-1 \right)^{k} {d-1 \choose k} e^{\frac{(d-1-2k)^2}{2}c\sigma^2 } e^{- \frac{1}{2 \sigma^2} \left[r - (d-1-2k)\sqrt{c} \sigma^2 \right]^2 }  \\
 &\le \frac{\1_{\R_+}(r)}{Z^{\text{R}}_r} \frac{1}{(2\sqrt{c})^{d-1}} \sum_{2k=0}^{d-1} {d-1 \choose 2k} e^{\frac{(d-1-4k)^2}{2}c\sigma^2 } e^{- \frac{1}{2 \sigma^2} \left[r - (d-1-4k)\sqrt{c} \sigma^2 \right]^2 }.
\end{align*}
Then we choose our proposal $g$ to be the truncated normal distribution associated with $k=0$, i.e. with mean $(d-1)\sqrt{c}\sigma^2$ and variance $\sigma^2$
\begin{align}
g(r) &= \frac{\1_{r>0}}{\sigma \left(1 - \Phi \left(- \frac{(d-1)\sqrt{c}\sigma^2}{\sigma} \right) \right)} \frac{1}{\sqrt{2\pi}} e^{- \frac{1}{2 \sigma^2} \left(r -  (d-1)\sqrt{c}\sigma^2 \right)^2} \nonumber \\
&= \frac{1}{\sqrt{2\pi}}\frac{\1_{r>0}}{\sigma \left(1 - \frac{1}{2} - \frac{1}{2}\erf\left(- (d-1)\sqrt{c}\frac{\sigma}{\sqrt{2}} \right) \right)} e^{- \frac{1}{2 \sigma^2} \left(r -  (d-1)\sqrt{c}\sigma^2 \right)^2} \nonumber \\
&= \sqrt{\frac{2}{\pi}}\frac{\1_{r>0}}{\sigma \left(1 + \erf\left(\frac{(d-1)\sqrt{c}\sigma}{\sqrt{2}} \right) \right)} e^{- \frac{1}{2 \sigma^2} \left(r -  (d-1)\sqrt{c}\sigma^2 \right)^2} \nonumber \\
&= \frac{\1_{r>0}}{Z_g(\sigma)} e^{- \frac{1}{2 \sigma^2} \left(r -  (d-1)\sqrt{c}\sigma^2 \right)^2}
\end{align}
with
\begin{align}
Z_g = \sqrt{\frac{\pi}{2}} \sigma \left(1 + \erf\left(\frac{(d-1)\sqrt{c}\sigma}{\sqrt{2}} \right) \right).
\end{align}
Computing the ratio of the densities yield
\begin{align*}
\frac{\rho(r)^{\text{R}}}{g(r)} &= \frac{Z_g(\sigma)}{Z^{\text{R}}_r} \frac{1}{(2\sqrt{c})^{d-1}} \sum_{k=0}^{d-1} \left(-1 \right)^{k} {d-1 \choose k} e^{\frac{(d-1-2k)^2}{2}c\sigma^2 } e^{- \frac{1}{2 \sigma^2} \left[r - (d-1-2k) \sqrt{c} \sigma^2 \right]^2 } e^{+ \frac{1}{2 \sigma^2} \left[r - (d-1)\sqrt{c} \sigma^2 \right]^2} \\ 
&= \frac{Z_g(\sigma)}{Z^{\text{R}}_r} \frac{1}{(2\sqrt{c})^{d-1}} \sum_{k=0}^{d-1} \left(-1 \right)^{k} {d-1 \choose k} e^{\frac{(d-1-2k)^2}{2}c\sigma^2 } e^{2k \sqrt{c} \left((d-1-k) \sqrt{c}\sigma^2 - r \right) }.
\end{align*}
Hence
\begin{align}
\rho(r)^{\text{R}}/g(r) \le M \triangleq \frac{Z_g(\sigma)}{Z^{\text{R}}_r} \frac{1}{(2\sqrt{c})^{d-1}} e^{\frac{(d-1)^2c\sigma^2}{2}}.
\end{align}

\paragraph{Rejection Sampling with Gamma proposal}
Now let's consider the following $\text{Gamma}(2, \sigma)$ density:
\begin{align}
g(r) &= \frac{\1_{r>0}}{Z_g(\sigma)} r e^{-\frac{r}{\sigma}} \nonumber
\end{align}
with 
\begin{align}
Z_g(\sigma) &= \Gamma(2) \sigma^2 \nonumber.
\end{align}

Then log ratio of the densities can be upper bounded as following:
\begin{align*}
\ln \left(\frac{\rho(r)^{\text{R}}}{g(r)} \right) &= \ln \frac{Z_g(\sigma)}{Z^{\text{R}}_r}
-\frac{r^2}{2\sigma^2} + (d-1)\ln(e^{\sqrt{c}r}- e^{-\sqrt{c}r}) - (d-1)\ln 2 -  \ln r + \frac{r}{\sigma} \nonumber \\
&= \ln \frac{Z_g(\sigma)}{Z^{\text{R}}_r} - (d-1)\ln 2 \underbrace{-\frac{r^2}{2\sigma^2} + \left((d-1)\sqrt{c}+\frac{1}{\sigma}\right)r}_{\le \frac{((d-1)\sqrt{c}\sigma+1)^2}{2}} + \underbrace{(d-1)\ln\left(\frac{1 - e^{-2\sqrt{c}r}}{r} \right)}_{\le (d-1)\ln(2\sqrt{c})}  \nonumber \\
&\le \ln \frac{Z_g(\sigma)}{Z^{\text{R}}_r} + \frac{((d-1)\sqrt{c}\sigma+1)^2}{2} + (d-1)\ln{\sqrt{c}}.  \\
\end{align*}
Hence
\begin{align}
\rho(r)^{\text{R}}/g(r) \le M \triangleq \frac{Z_g(\sigma)}{Z^{\text{R}}_r} c^{\frac{d-1}{2}} e^{\frac{((d-1)\sqrt{c}\sigma+1)^2}{2}}.
\end{align}

\subsubsection{Normalisation constant}
\label{sec:normconst}
In order to evaluate the density of the \emph{Riemannian} normal distribution, we need to compute the normalisation constant, which we derive in this subsection.
\paragraph{Cumulative density function}
First let's derive the cumulative density function of the hyperbolic radius.
Integrating the expended density of Eq (\ref{eq:developedpr}) yields
\begin{align}
F^{\text{R}}_r(r) &= \int_{-\infty}^{r} \rho^{\text{R}}(r) dr \nonumber \\
&= \frac{1}{Z^{\text{R}}_r}  \frac{1}{(2\sqrt{c})^{d-1}} \sum_{k=0}^{d-1} \left(-1 \right)^{k} {d-1 \choose k} e^{\frac{(d-1-2k)^2}{2}c\sigma^2 }  \nonumber \\
&\hspace{2em} \times \int_{0}^{r} e^{- \frac{1}{2 \sigma^2} \left[r - (d-1-2k)\sqrt{c} \sigma^2 \right]^2 } dr
 \sum_{k=0}^{d-1} \left(-1 \right)^{k} {d-1 \choose k} e^{\frac{(d-1-2k)^2}{2}c\sigma^2 } \nonumber \\
& \hspace{2em} \times \left[\erf\left( \frac{r-(d-1-2k)\sqrt{c}\sigma^2}{\sqrt{2}\sigma} \right) \right.  \left. \erf\left( \frac{(d-1-2k)\sqrt{c}\sigma}{\sqrt{2}} \right) \right] \label{eq:cdf}
\end{align}
with $\Phi: x \mapsto \frac{1}{2}\left(1 + \erf\left( \frac{x}{\sqrt{2}} \right) \right)$, the cumulative distribution function of a standard normal distribution.
\paragraph{Taking the limit}
$F^{\text{R}}_r(r) \xrightarrow[r \to \infty]{} 1$ in Eq (\ref{eq:cdf}) yield
\begin{align}
&Z_r^{\text{R}} = \sqrt{\frac{\pi}{2}} \sigma \frac{1}{(2\sqrt{c})^{d-1}} \sum_{k=0}^{d-1} \left(-1 \right)^{k} {d-1 \choose k}
 e^{\frac{(d-1-2k)^2}{2}c\sigma^2 } \left[1 +\erf\left( \frac{(d-1-2k)\sqrt{c}\sigma}{\sqrt{2}} \right) \right].
\label{eq:Zr2}
\end{align}
Note that by the antisymmetry of $\erf$, one can simplify Eq (\ref{eq:Zr2}) with a sum over $\ceil{d / 2}$ terms (as done in \cite{Hauberg}).
Also, computing such a sum is much more stable by relying on the \emph{log sum exp} trick.
Integrating Equation (\ref{eq:volume}) of Appendix \ref{sec:pr} gives
\begin{align}
Z^{\text{R}} = Z_r^{\text{R}} Z_{\bm{\alpha}}
\end{align}
As a reminder, the surface area of the $d-1$-dimensional hypersphere with radius $1$ is given by
\begin{align*}
Z_\alpha = A_{\mathbb{S}^{d-1}} = \frac{2 \pi^{d/2}}{\Gamma(d/2)}.
\end{align*}
For the special case of $c=1$ and $d=2$ we recover the formula given in \cite{DBLP:journals/entropy/SaidBB14}
\begin{align*}
Z_r^{\text{R}} = \sqrt{\frac{\pi}{2}} \sigma e^{\frac{\sigma^2}{2}} \erf\left(\frac{\sigma}{\sqrt{2}}\right).
\end{align*}



%
\subsubsection{Expectation of hyperbolic radii}
\label{sec:expectation}

%
%

Computing the expectation of the hyperbolic radius $r \sim \rho^{\text{R}}$ is of use to choose the initial set of points to construct the piecewise exponential proposal.
By integrating the expended density of Eq (\ref{eq:developedpr}), we get
\begin{align}
\mathbb{E}[r] &= \int_{-\infty}^{\infty} r \rho^\text{R}(r) dr \nonumber \\
&= \frac{1}{Z^\text{R}_r}  \frac{1}{(2\sqrt{c})^{d-1}} \sum_{k=0}^{d-1} \left(-1 \right)^{k} {d-1 \choose k} e^{\frac{(d-1-2k)^2}{2}c\sigma^2 } \int_{0}^{\infty} re^{- \frac{1}{2 \sigma^2} \left[r - (d-1-2k)\sqrt{c} \sigma^2 \right]^2 } dr \nonumber \\
&= \frac{1}{Z^\text{R}_r}  \frac{1}{(2\sqrt{c})^{d-1}} \sum_{k=0}^{d-1} \left(-1 \right)^{k} {d-1 \choose k} e^{\frac{(d-1-2k)^2}{2}c\sigma^2 } \nonumber \\
& \hspace{2em} \times  \left[\sqrt{\frac{\pi}{2}} (d-1-2k)\sqrt{c} \sigma^2 \sigma \left(1 + \erf \left(\frac{(d-1-2k)\sqrt{c} \sigma}{\sqrt{2}} \right) \right) + \sigma^2 e^{-\frac{(d-1-2k)^2c \sigma^2}{2}} \right] \nonumber  \\
&= \frac{1}{Z^\text{R}_r} \sqrt{\frac{\pi}{2}} \sigma  \frac{1}{(2\sqrt{c})^{d-1}} \sum_{k=0}^{d-1} \left(-1 \right)^{k} {d-1 \choose k}  \nonumber \\
& \hspace{2em} \times  \left[e^{\frac{(d-1-2k)^2}{2}c\sigma^2 } (d-1-2k)\sqrt{c} \sigma^2  \left(1 + \erf \left(\frac{(d-1-2k)\sqrt{c} \sigma}{\sqrt{2}} \right) \right) + \sigma \sqrt{\frac{2}{\pi}}  \right] \nonumber  \\
\end{align}
\section{Experimental details}
\label{sec:exp_details}
In this section we give more details on the datasets, architecture designs and optimisation schemes used for the experimental results given in Section \ref{sec:experiments}.
\subsection{Synthetic Branching Diffusion Process}
\paragraph{Generation}
Nodes $(\y_1, \dots, \y_N) \in \R^n$ of the branching diffusion process are sampled as follow
\begin{align*}
\y_i \sim \mathcal{N} \left(\cdot ~| \y_{\pi(i)}, \sigma_0^2 \right) \quad \forall i \in 1,\dots,N
\end{align*}
with $\pi(i)$ being the index of the $i$th node's ancestor and $d(i)$ its depth.
Then, noisy observations are sampled for each node $\x_i$,
\begin{align*}
\x_{i,j} = \y_i + \bm{\epsilon}_{i,j}, \quad \bm{\epsilon}_{i,j} \sim \mathcal{N} \left(\cdot ~| \bm{0}, \sigma_j^2 \right) \quad & \forall i,j.
\end{align*}
The root $x_0$ is set to $\bm{0}$ for simplicity.
The observation dimension is set to $n=50$.
The dataset $\left(\x_{i,j}\right)_{i,j}$ is centered and normalised to have unit variance.
Thus, the choice of variance $\sigma_0^2$ does not matter and it is set to $\sigma_0=1$.
The number of noisy observations is set to $J=5$, and its variance to $\sigma_j^2 = \sigma_0^2/5 = 1/5$.
The depth is set to $6$ and the branching factor to $2$.
\paragraph{Architectures}
Both \nvae{} and \pvae{c} decoders parametrise the mean of the unit variance Gaussian likelihood $\mathcal{N}(\cdot|f_{\bm{\theta}}(\z), 1)$.
Their encoders parametrise the mean and the log-variance of respectively an isotropic normal distribution $\mathcal{N}(\cdot|g_{\bm{\phi}}(\z))$ and an isotropic hyperbolic normal distribution $\mathcal{N}_{\B^d_c}(\cdot|g_{\bm{\phi}}(\z))$.
The \nvae{}'s encoder and decoder are composed of $2$ Fully-Connected layers with a ReLU activation  in between, as summed up in Tables \ref{table:euc_enc_arc} and \ref{table:euc_dec_arc}.
The \pvae{c}'s design is similar, the differences being that the decoder's output is mapped to manifold via the exponential map $\exp^c_{\bm{0}}$, and the decoder's first layer is made of \emph{gyroplane units} presented in Section \ref{sec:arch}, as summarised in Tables \ref{table:poin_enc_arc} and \ref{table:poin_dec_arc}.
Observations live in $\X=\R^{50}$ and the latent space dimensionality $d$ is set to $d=2$.
\begin{table}[h!]
\parbox{.45\linewidth}{
\centering
\caption{Encoder network for \nvae{}}
\label{table:euc_enc_arc}
\begin{tabular}{@{}lll@{}}
                \textbf{Layer}    & \textbf{Output dim}  &  \textbf{Activation} \\ \midrule
                Input & $50$ & Identity \\ \midrule
                 FC & $200$ & ReLU \\  \midrule
                 FC & $2, 1$ & Identity \\  \bottomrule
\end{tabular}
}
\hfill
\parbox{.45\linewidth}{
\centering
\caption{Decoder network for \nvae{}}
\label{table:euc_dec_arc}
\begin{tabular}{@{}lll@{}}
                \textbf{Layer}    & \textbf{Output dim}  &  \textbf{Activation} \\ \midrule
                 Input & $2$ & Identity \\ \midrule
                 FC & $200$ & ReLU \\ \midrule
                 FC & $50$ & Identity \\ \bottomrule
\end{tabular}
}
\end{table}

\begin{table}[h!]
\parbox{.45\linewidth}{
\centering
\caption{Encoder network for \pvae{c}}
\label{table:poin_enc_arc}
\begin{tabular}{@{}lll@{}}
                \textbf{Layer}    & \textbf{Output dim}  &  \textbf{Activation} \\ \midrule
                Input & $50$ & Identity \\ \midrule
                 FC & $200$ & ReLU \\ \midrule
                 FC & $2, 1$ & $\exp^c_{\bm{0}}$, Identity \\ \bottomrule
\end{tabular}
}
\hfill
\parbox{.45\linewidth}{
\centering
\caption{Decoder network for \pvae{c}}
\label{table:poin_dec_arc}
\begin{tabular}{@{}lll@{}}
                \textbf{Layer}    & \textbf{Output dim}  &  \textbf{Activation} \\ \midrule
                Input & $2$ & Identity \\ \midrule
                Gyroplane &  200 & ReLU   \\ \midrule
                FC & $50$ & Identity \\ \bottomrule
\end{tabular}
}
\end{table}
The synthetic datasets are generated as described in Section \ref{sec:experiments}, then centred and normalised to unit variance.
There are then randomly split into training and testing datasets with a proportion $0.7$.
\paragraph{Optimisation}
Gyroplane offset $\p \in \B^d_c$ are only \emph{implicitly} parametrised to live in the manifold, by projecting a real vector $\p = \exp^c_{\bm{0}}(\p')$.
Hence, all parameters $\{\bm{\theta}, \bm{\phi}\}$ of the model explicitly live in Euclidean spaces which means that usual optimisation schemes can be applied.
We therefore rely on Adam optimiser \citep{Kingma:2014us} with parameters $\beta_1 = 0.9$, $\beta_2 = 0.999$ and a constant learning rate set to $1e-3$. Models are trained with mini-batches of size $64$ for $1000$ epochs.
The \gls{ELBO} is approximated with a \gls{MC} estimate with $K=1$.
\paragraph{Baselines}
The \acrfull{PCA} embeddings are obtained via a \gls{SVD} by projecting the dataset on the basis associated with the two highest singular values.
The \acrfull{GPLVM} embeddings are obtained by maximising the marginal likelihood of a (non-Bayesian) \gls{GPLVM} with RBF kernel, and whose latent variables are initialised with \gls{PCA}.

\subsection{MNIST digits}
The MNIST dataset \citep{lecun-mnisthandwrittendigit-2010} contains 60,000 training and 10,000 test images of ten handwritten digits (zero to nine), with 28x28 pixels.
\paragraph{Architectures}
The architectures used for the encoder and the decoder for Mnist are similar to the ones used for the Synthetic Branching Diffusion Process.
They differ by the dimensions of the observation space ($\X=\R^{28 \times 28}$) and hidden space.
The output of the first fully connected layer is here equal to $600$.
The latent space dimensionality $d$ is set to $2$, $5$, $10$ and $20$ respectively.
The bias of the decoder's last layer is set to the average value of digits (for each pixel).
The architectures used for the classifier are similar than the decoder architectures, the only difference being the output dimensionality (10 labels).
We initialise the classifier's first layer with decoder's first layer weights.
Then the classifier is trained to minimise the cross entropy for $5$ epochs, with mini-batches of size $64$ and a constant learning rate of $1e{-3}$.
\paragraph{Optimisation}
We use $[0,1]$ normalised data as targets for the mean of a Bernoulli distribution, using negative cross-entropy for log $p(\x|\z)$.
We set the prior distribution’s distortion to $\sigma = 1$.
We rely on Adam optimiser with parameters $\beta_1 = 0.9$, $\beta_2 = 0.999$ and a constant learning rate of $5e^{-4}$.
Models are trained with mini-batches of size $128$ for $80$ epochs.

\subsection{Graph embeddings}
The PhD advisor-advisee relationships graph \citep{Nooy:2011:ESN:2181139} contains $344$ nodes and $343$ edges.
The phylogenetic tree expressing genetic heritage  \citep{Hofbauer,TreeBASE} contains $1025$ nodes and $1043$ edges.
The biological set representing disease relationships \citep{Goh8685,Rossi:2015:NDR:2888116.2888372} contains $516$ nodes and $1188$ edges.
We follow the training and evaluation procedure introduced in \cite{Kipf:2016ul}.

\paragraph{Architectures}
We also follow the featureless architecture introduced in \cite{Kipf:2016ul}, namely a two-layer \gls{GCN} with $32$ hidden dimensions to parametrise the variational posteriors, and a likelihood which factorises along edges 
$p(\bm{A}|\bm{Z}) = \prod_{i=1}^N \prod_{j=1}^N p({A}_{ij}|\z_i,\z_j)$, with $\bm{A}$ being the adjacency matrix.
The probability of an edge is defined through the latent metric by $p(A_{ij} = 1|\z_i, \z_j) = 1 - \tanh(d_{\M}(\z_i, \z_j))$.
For the Poincar\'e ball latent space, the encoder output is projected on the manifold: $\bm{\mu} = \exp_{\bm{0}}(\text{GCN}_{\bm{\mu}}({\bm{A}}))$.
The latent dimension is set to $5$ for the experiments.
We use a \emph{Wrapped} Gaussian prior and variational posterior.

\paragraph{Optimisation}
We use the adjacency matrix $\bm{A}$ as target for the mean of a Bernoulli distribution, using negative cross-entropy for log $p(\bm{A}|\bm{Z})$.
We rely on Adam optimiser with parameters $\beta_1 = 0.9$, $\beta_2 = 0.999$ and a constant learning rate of $1e^{-2}$.
We perform full-batch gradient descent for $800$ epochs and make use of the reparametrisation trick for training.

\section{More experimental qualitative results}
\label{sec:qual_results}
\begin{figure}[h]
\begin{center}
  \includegraphics[width=0.32\textwidth]{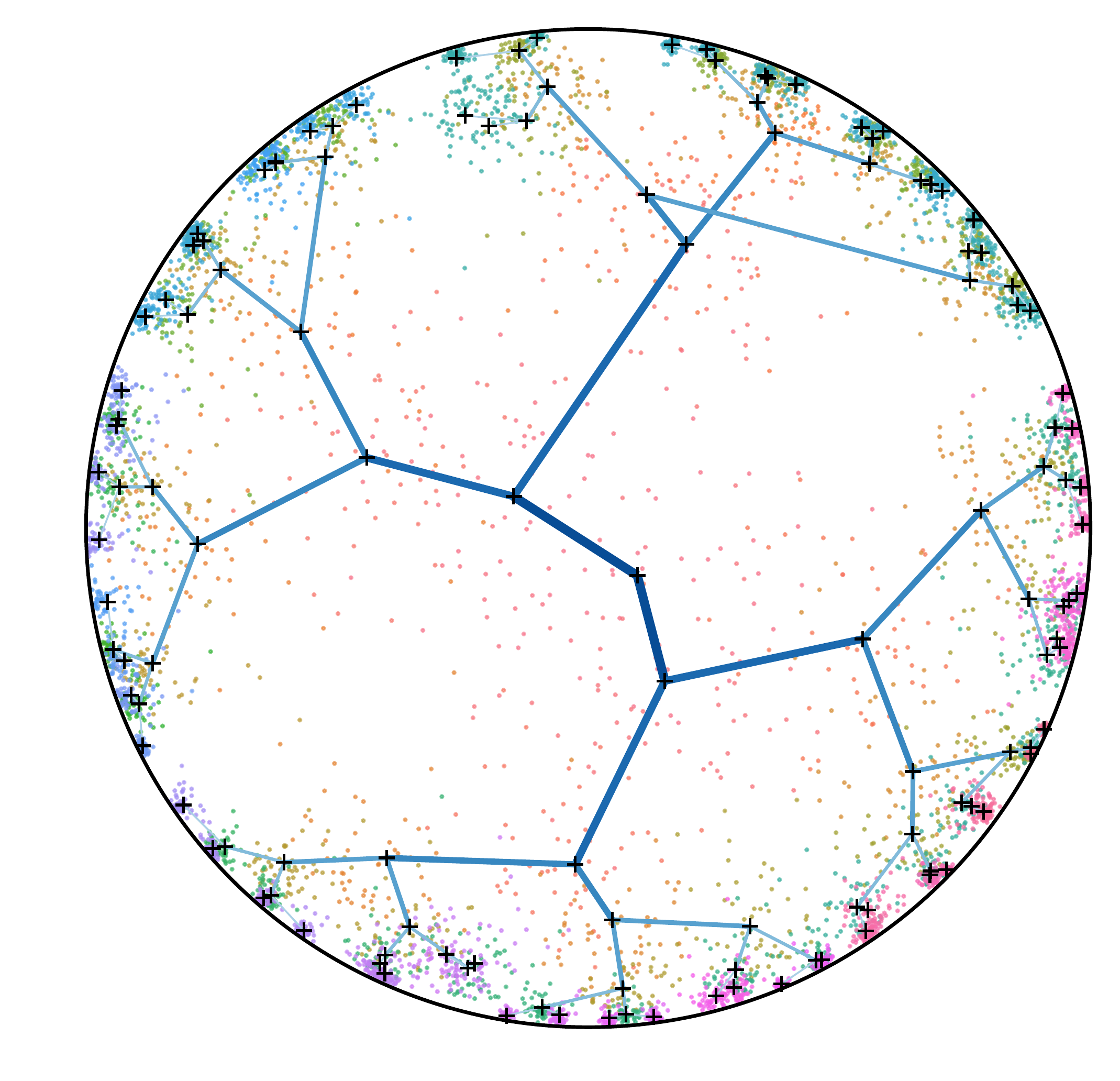}
  \includegraphics[width=0.32\textwidth]{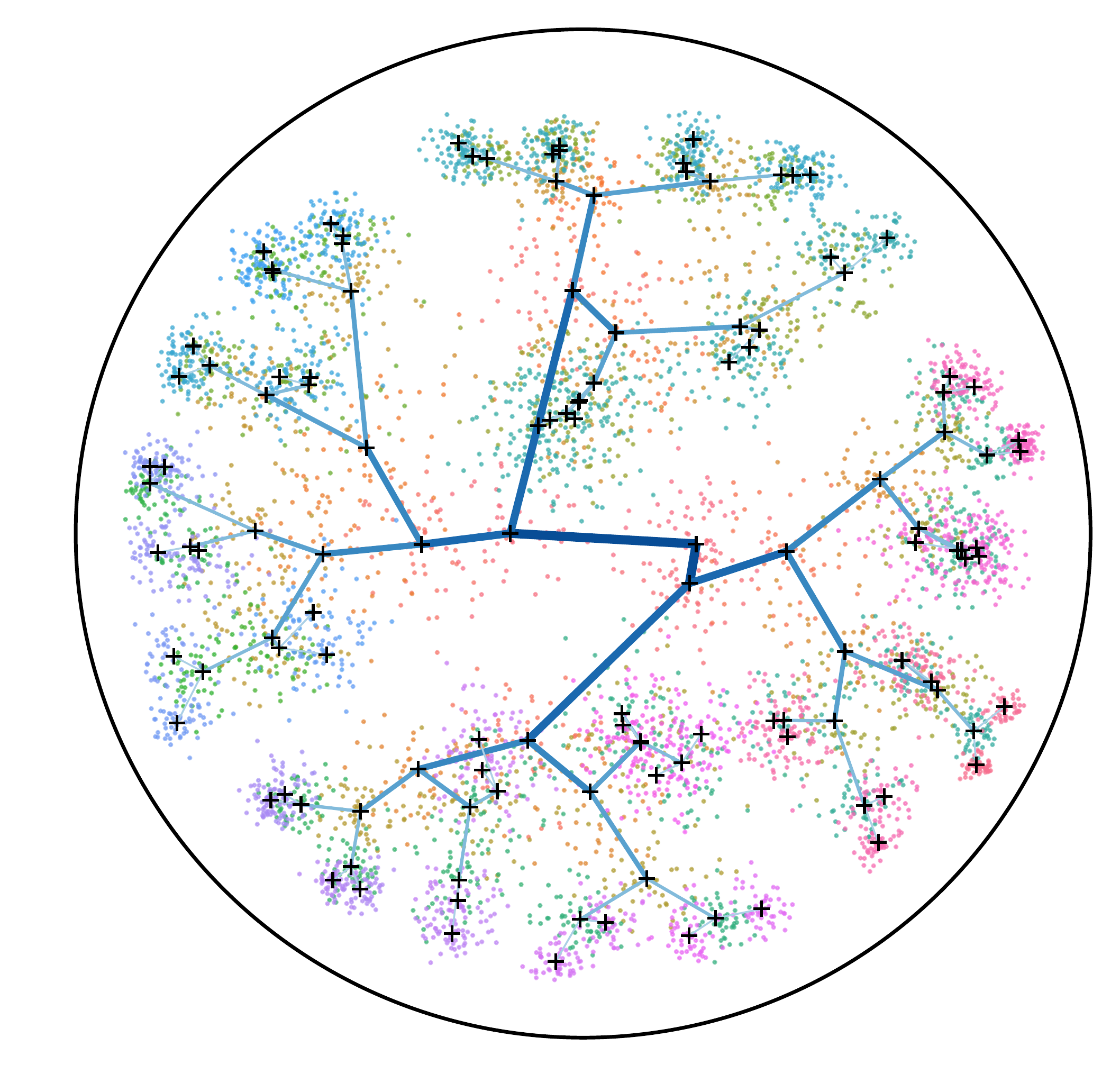}
  \includegraphics[width=0.32\textwidth]{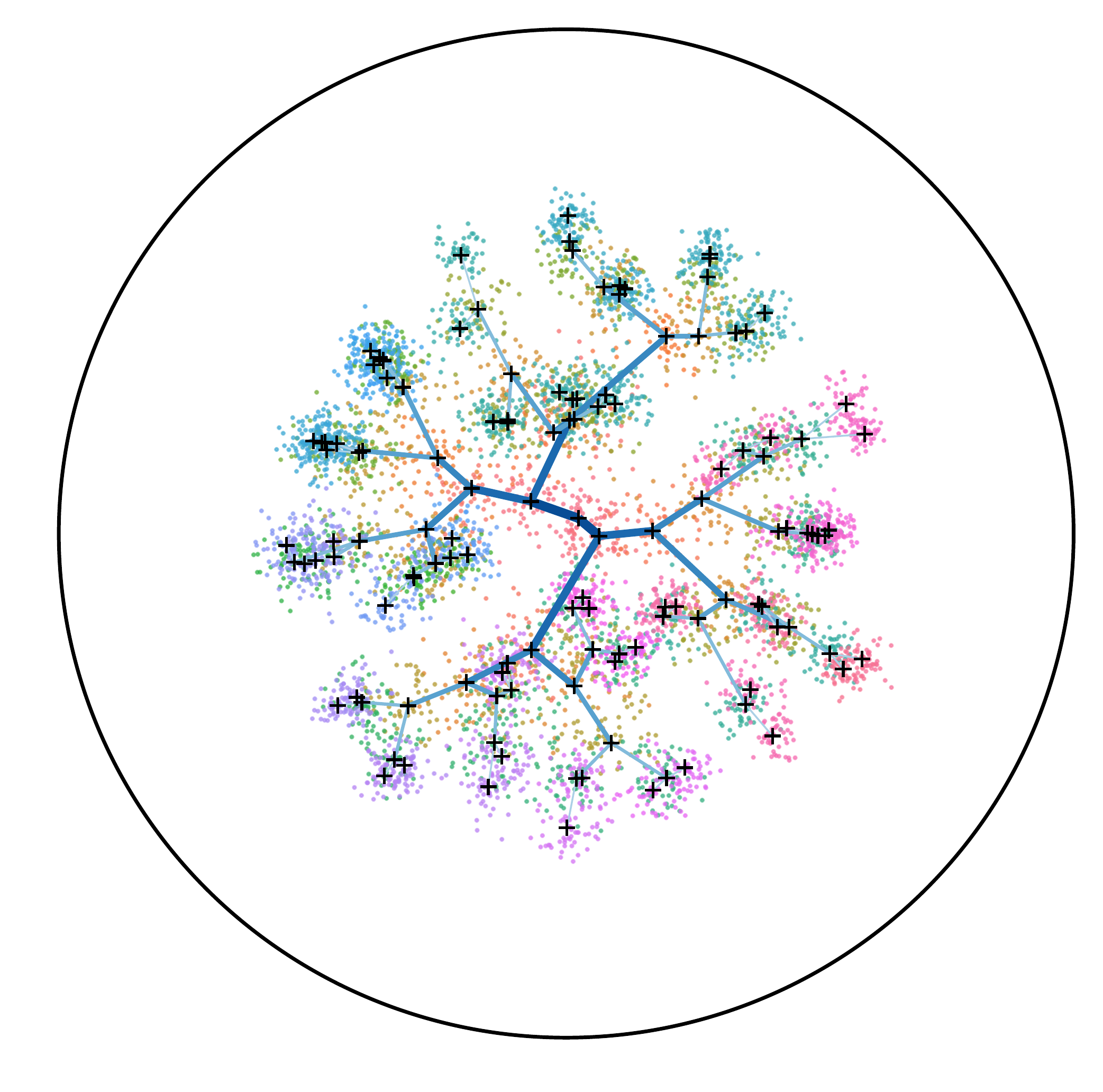}
  \caption{Branching diffusion process latent representations of \pvae{c} with decreasing curvatures $c=1.2, 0.3, 0.1$ (Left to Right).}
  \label{fig:synthetic_posterior_curvatures}
\end{center}
\end{figure}

\begin{figure}[h]
\begin{center}
  \includegraphics[width=0.32\textwidth]{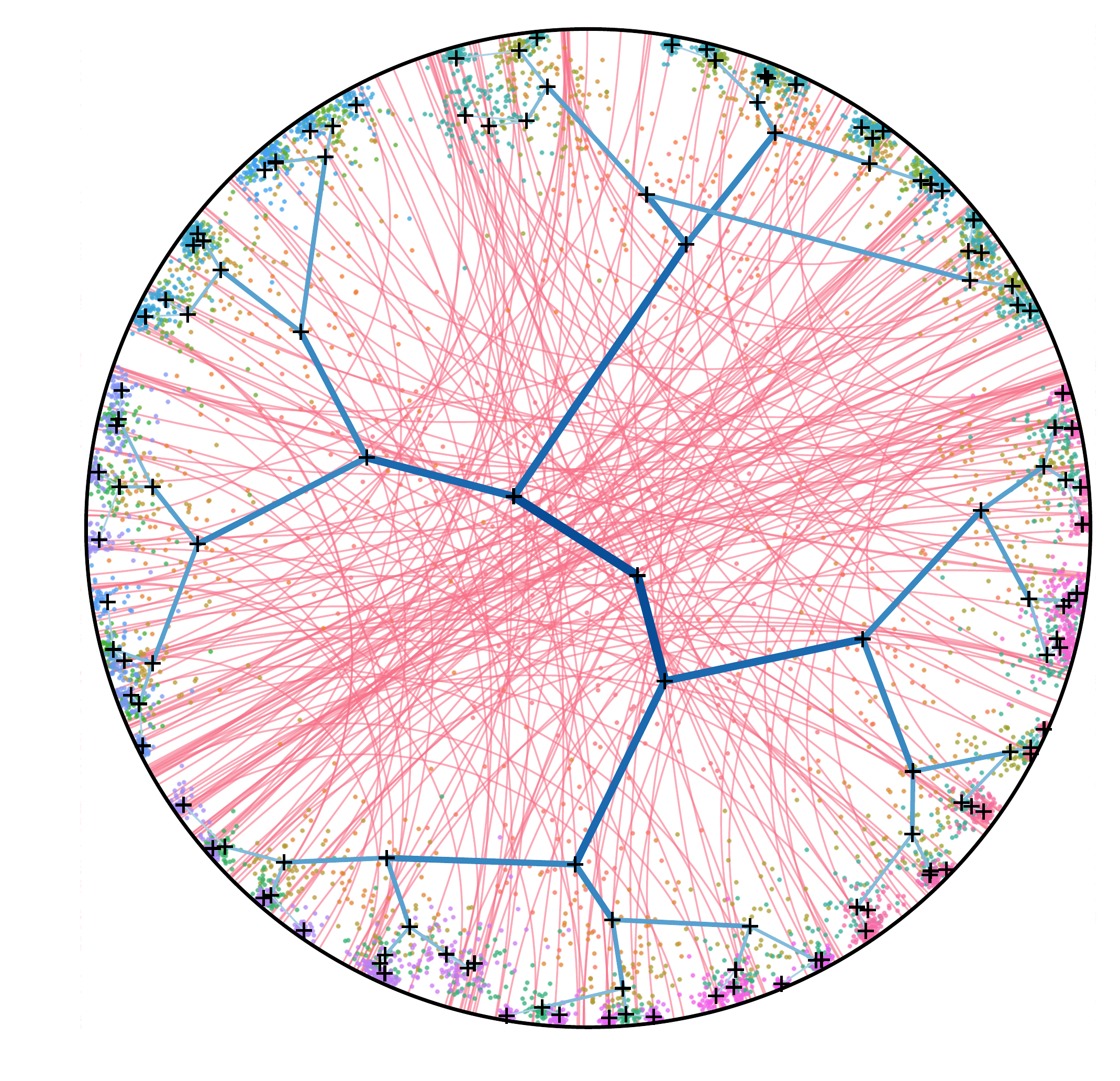}
  \includegraphics[width=0.32\textwidth]{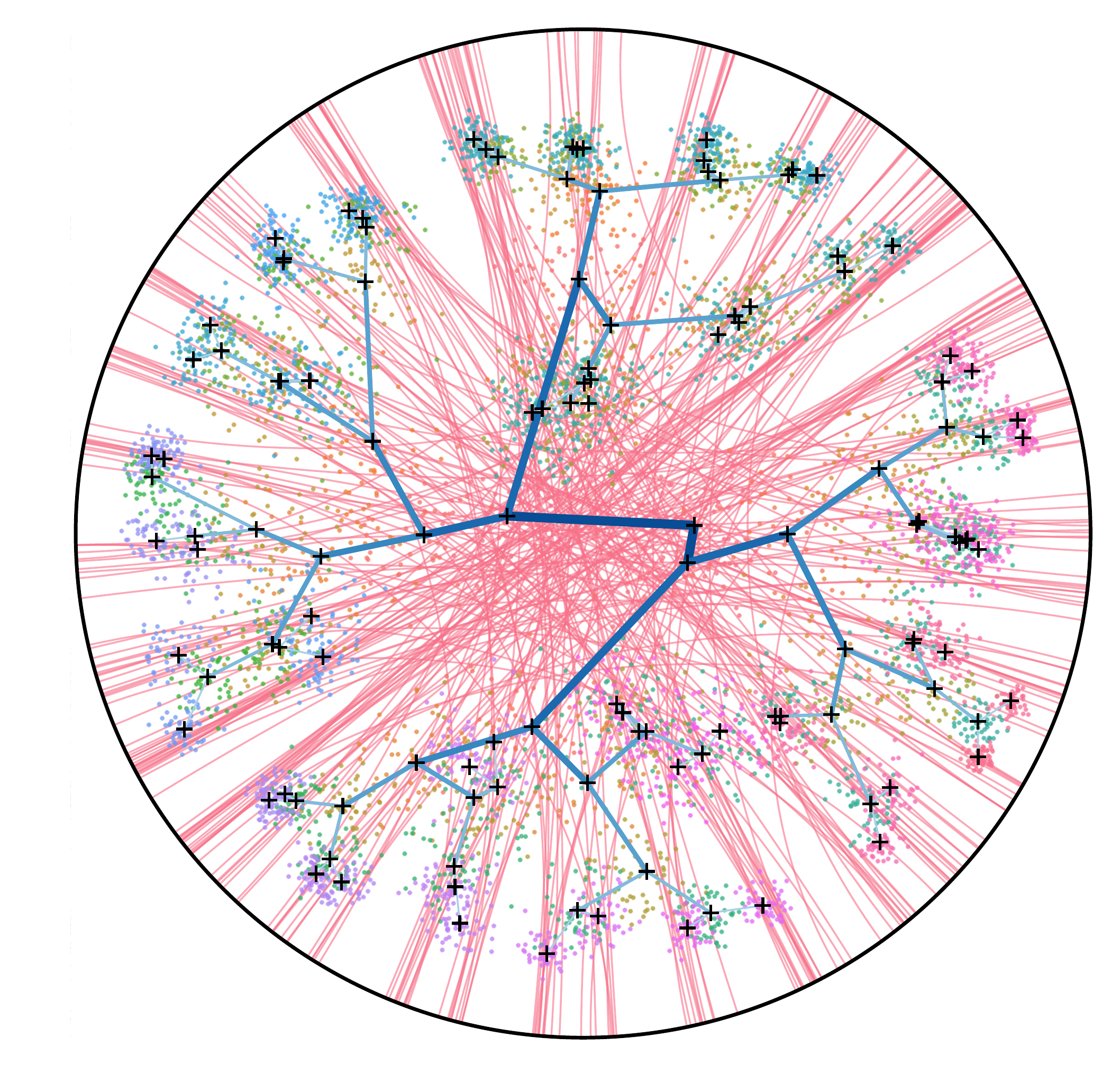}
  \includegraphics[width=0.32\textwidth]{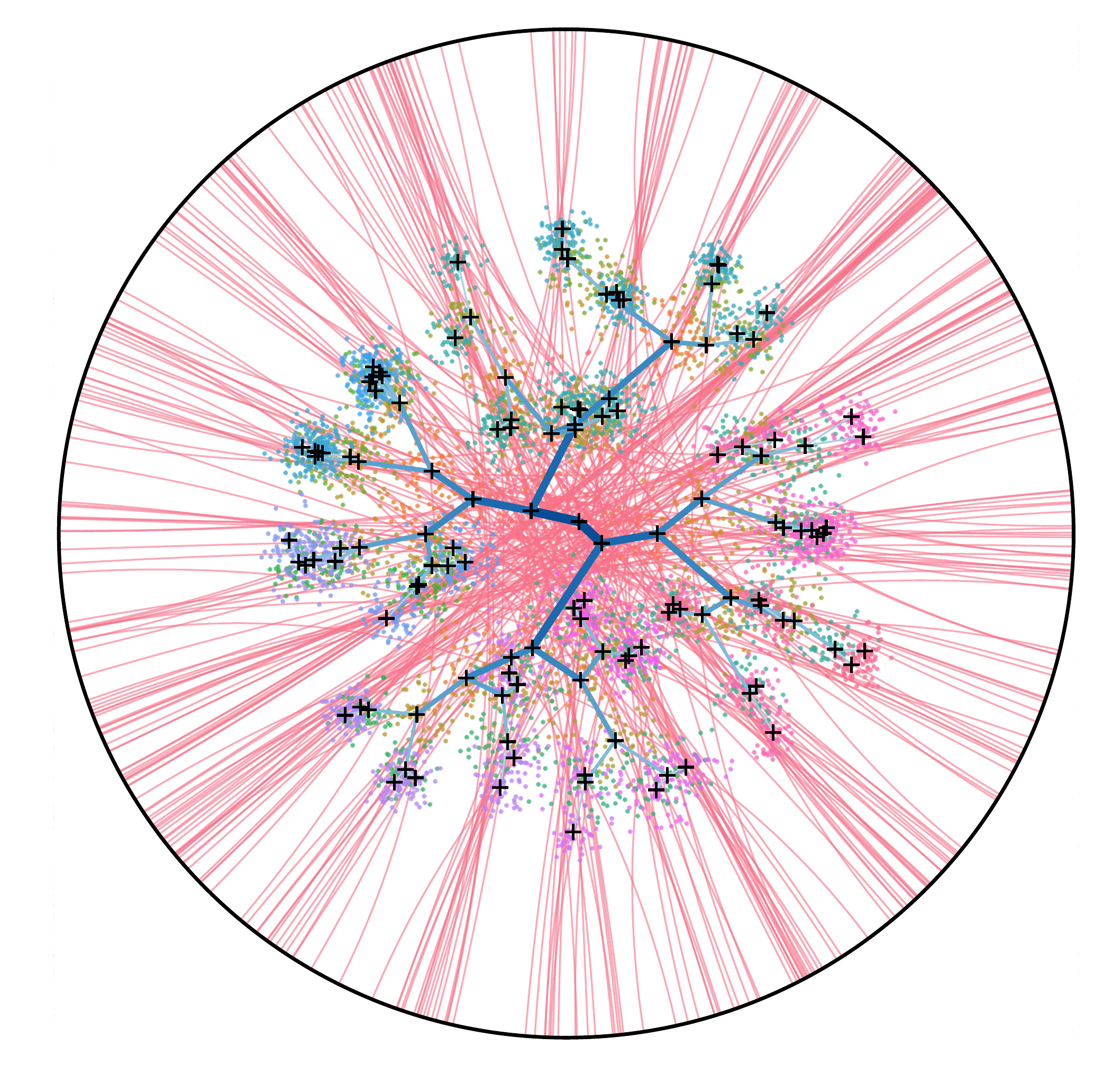}
  \caption{Branching diffusion process latent representations of \pvae{c} with decreasing curvatures $c=1.2, 0.3, 0.1$ (Left to Right).}
  \label{fig:synthetic_posterior_curvatures_hyperplanes}
\end{center}
\end{figure}

Figure \ref{fig:synthetic_posterior_curvatures} shows latent representations of \pvae{c}s with different curvatures.
With "small" curvatures, we observe that embeddings lie close the center of the ball, where the geometry is close to be Euclidean.
Similarly as Figure \ref{fig:synthetic_posterior_curvatures}, Figure \ref{fig:synthetic_posterior_curvatures_hyperplanes} illustrates the learned latent representations of \pvae{c} with decreasing curvatures $c$, by highlighting the leaned \emph{gyroplanes} of the decoder.
\begin{figure}[h]
\begin{center}
  \includegraphics[width=0.40\textwidth, trim={1.64cm 1.00cm 0.60cm 0.54cm}, clip]{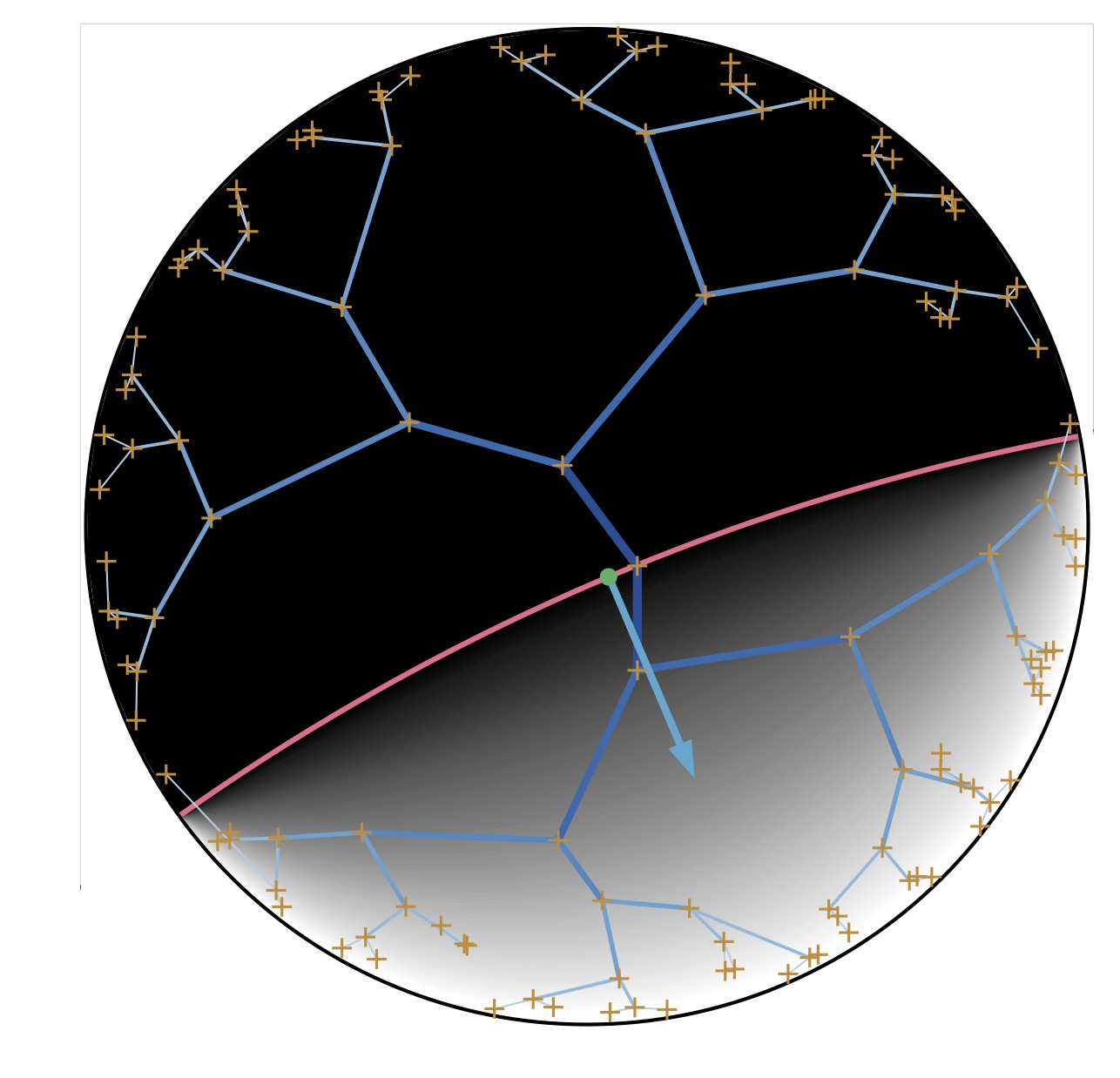}
  \includegraphics[width=0.40\textwidth, trim={2cm 2cm 2cm 2cm}, clip]{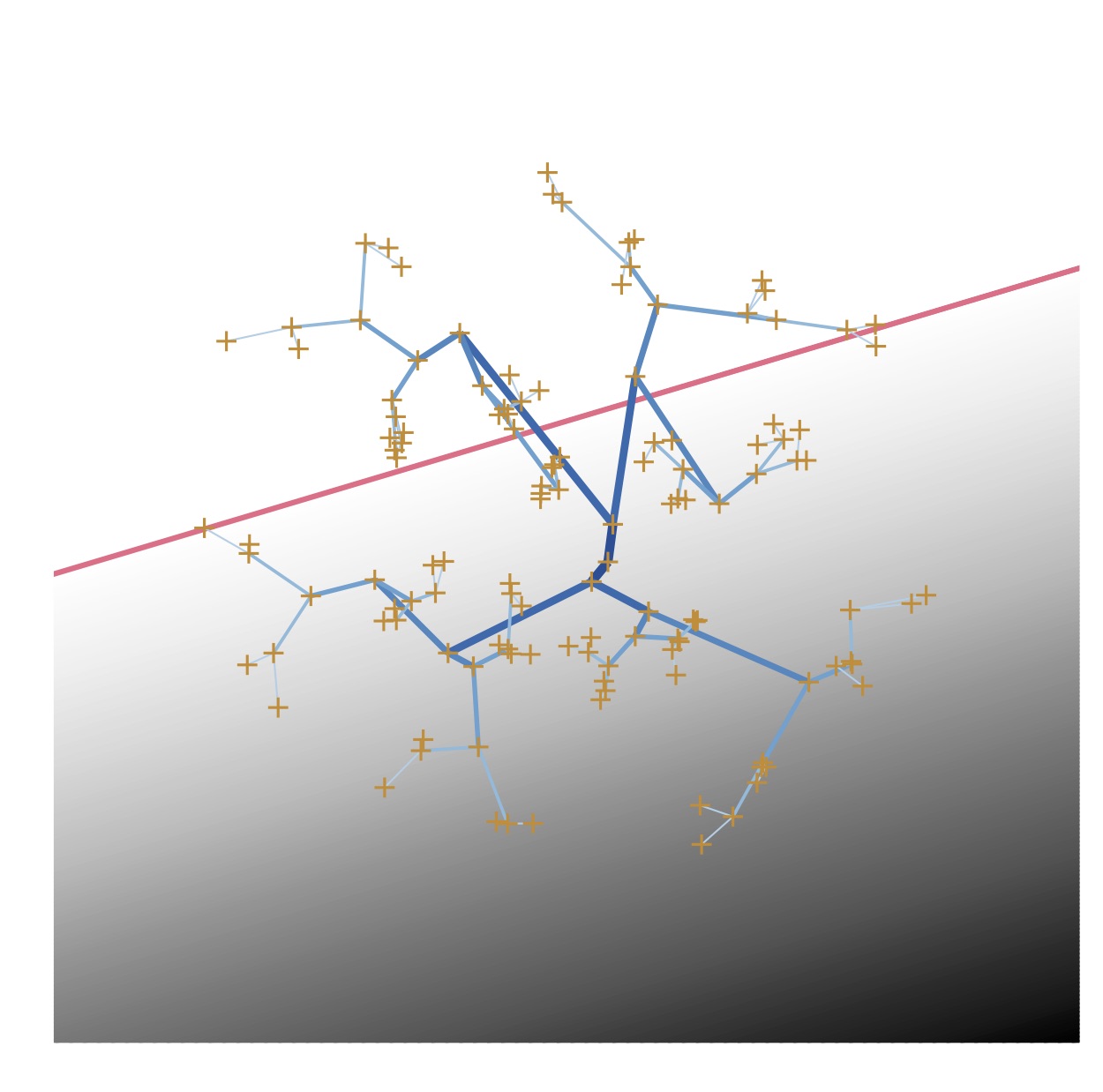}
  \caption{Branching diffusion process latent representation of \pvae{1} (Left) and \nvae{} (Right) with heatmap of the log distance to the hyperplane (in pink).}
  \label{fig:distance_gyroplane}
\end{center}
\end{figure}
\begin{figure}[h]
\begin{center}
  \includegraphics[width=0.40\textwidth]{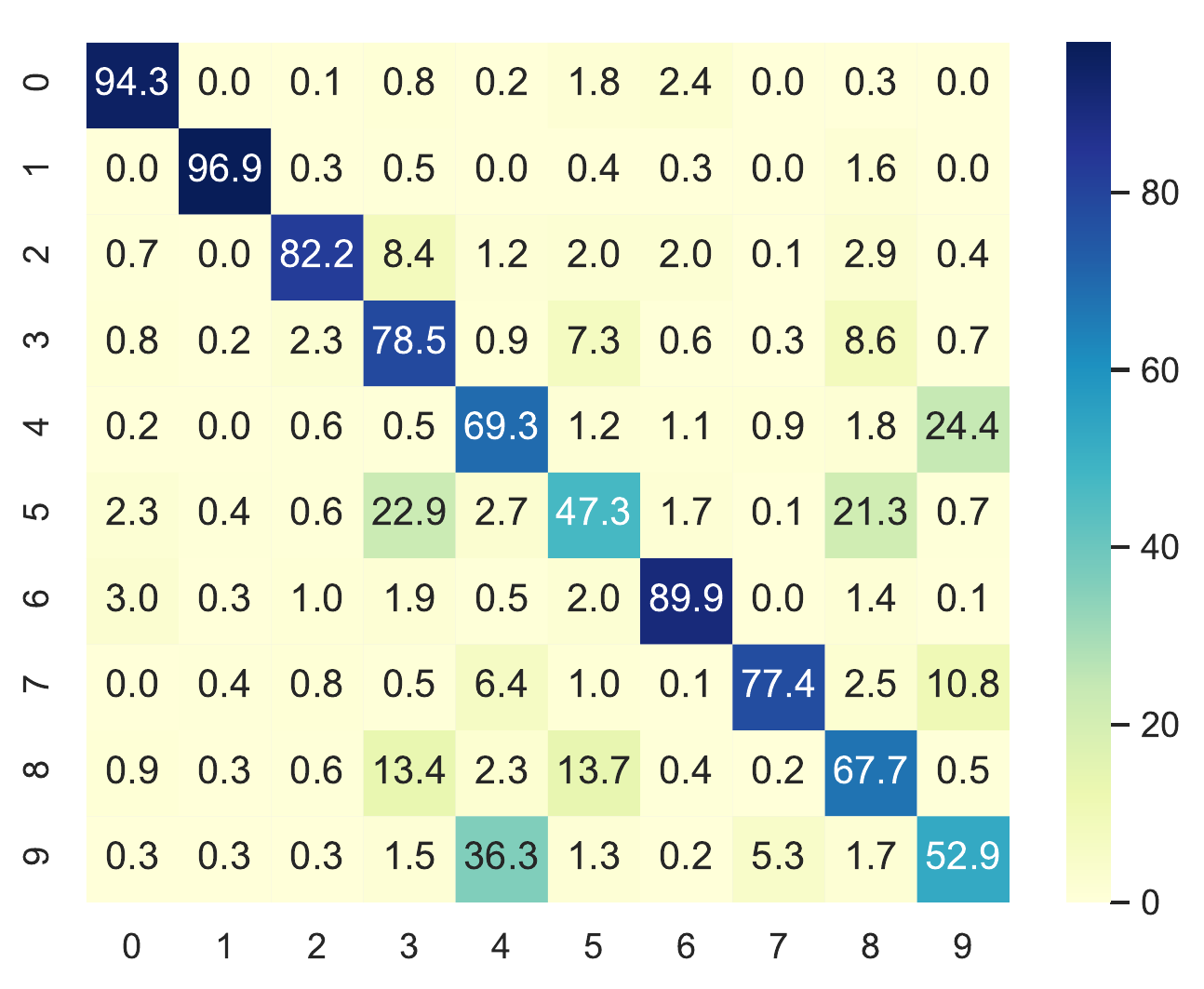}
  \includegraphics[width=0.40\textwidth]{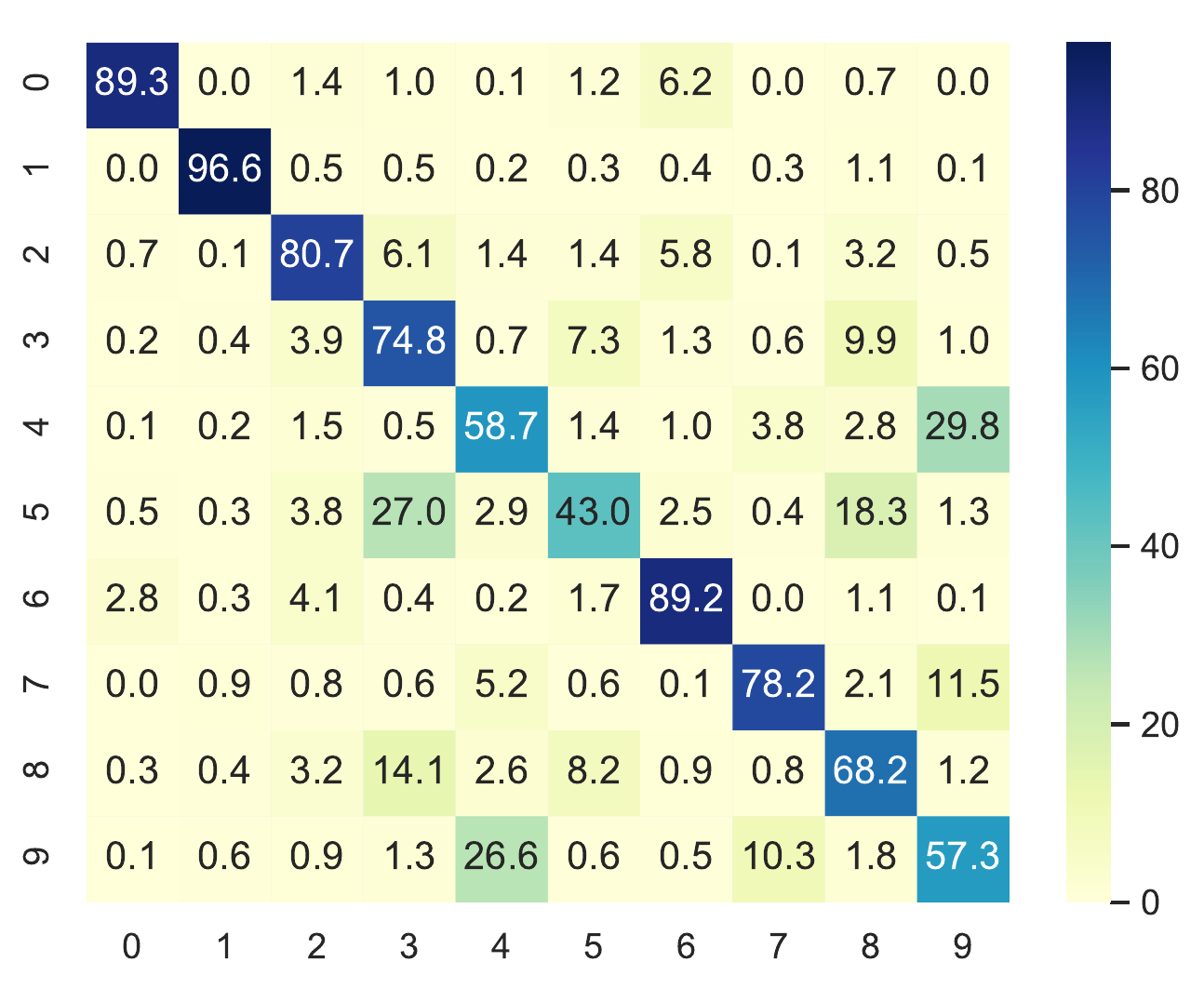}
  \caption{MNIST average confusion matrices of the classifiers trained on embeddings from the \pvae{1.4} (Left) and \nvae{} (Right) models.}
  \label{fig:confusion}
\end{center}
\end{figure}

\end{appendices}

\end{document}